\documentclass[12pt]{article}

\usepackage{amsmath}
\usepackage{graphicx}
\usepackage{amssymb, bbm}
\usepackage{caption}
\usepackage{times}
\usepackage{natbib}
\usepackage{xcolor}
\usepackage{enumitem}

\usepackage{booktabs, threeparttable, siunitx}       
\usepackage{ amsthm, mathrsfs, mathtools}
\usepackage{nicefrac}       
\usepackage{multirow}  
\usepackage{setspace}
\usepackage{xr}

\usepackage[plain,noend]{algorithm2e}

\makeatletter
\renewcommand{\algocf@captiontext}[2]{#1\algocf@typo. \AlCapFnt{}#2} 
\def\@algocf@capt@plain{top}
\renewcommand{\algocf@makecaption}[2]{%
	\addtolength{\hsize}{\algomargin}%
	\sbox\@tempboxa{\algocf@captiontext{#1}{#2}}%
	\ifdim\wd\@tempboxa >\hsize
	\hskip .5\algomargin%
	\parbox[t]{\hsize}{\algocf@captiontext{#1}{#2}}
	\else%
	\global\@minipagefalse%
	\hbox to\hsize{\box\@tempboxa}
	\fi%
	\addtolength{\hsize}{-\algomargin}%
}
\makeatother

\makeatletter
\newcommand{\distas}[1]{\mathbin{\overset{#1}{\kern\z@\sim}}}%
\makeatother

\newtheorem{assumption}{Assumption}
\newtheorem{theorem}{Theorem}
\newtheorem{corollary}{Corollary}[theorem]
\newtheorem{lemma}{Lemma}

\newcommand{\E}{\mathbb{E}}
\newcommand{\prob}{\mathbb{P}}
\newcommand{\ind}{\mathbbm{1}}
\newcommand{\var}{\textnormal{var}}

\DeclareMathOperator*{\argmax}{arg\,max}
\newcommand\independent{\protect\mathpalette{\protect\independenT}{\perp}}
\def\independenT#1#2{\mathrel{\rlap{$#1#2$}\mkern2mu{#1#2}}}
\definecolor{amber}{rgb}{1.0, 0.49, 0.0}

\usepackage[plain,noend]{algorithm2e}

\usepackage[left=0.9in,right=0.9in,top=1in,bottom=1in]{geometry}

\begin{document}

	\def\spacingset#1{\renewcommand{\baselinestretch}%
		{#1}\small\normalsize} \spacingset{1}

		\title{\bf Robust Sample Weighting to Facilitate Individualized Treatment Rule Learning for a Target Population}
		\author{Rui Chen
			\\
			Department of Statistics, University of Wisconsin\\
			Jared D. Huling \\
			Division of Biostatistics, University of Minnesota\\
			Guanhua Chen\thanks{Corresponding authors: gchen25@wisc.edu} \\ 
			Department of Biostatistics and Medical Informatics, University of Wisconsin\\
			Menggang Yu\thanks{Corresponding authors: meyu@biostat.wisc.edu} \\
			Department of Biostatistics and Medical Informatics, University of Wisconsin\\
		}
		\maketitle

\bigskip
\begin{abstract}
Learning individualized treatment rules (ITRs) is an important topic in precision medicine. Current literature mainly focuses on deriving ITRs from a single source population. We consider the observational data setting when the source population differs from a target population of interest. Compared with causal generalization for the average treatment effect which is a scalar quantity, ITR generalization poses new challenges due to the need to model and generalize the rules based on a prespecified class of functions which may not contain the unrestricted true optimal ITR. The aim of this paper is to develop a weighting framework to mitigate the impact of such misspecification and thus facilitate the generalizability of optimal ITRs from a source population to a target population. Our method seeks covariate balance over a non-parametric function class characterized by a reproducing kernel Hilbert space and can improve many ITR learning methods that rely on weights. We show that the proposed method encompasses importance weights and overlap weights as two extreme cases, allowing for a better bias-variance trade-off in between. Numerical examples demonstrate that the use of our weighting method can greatly improve ITR estimation for the target population compared with other weighting methods.
\end{abstract}

\noindent%
\vfill


\newpage
\spacingset{1.75} 
\setlength{\abovedisplayskip}{7pt}%
\setlength{\belowdisplayskip}{7pt}%
\setlength{\abovedisplayshortskip}{5pt}%
\setlength{\belowdisplayshortskip}{5pt}%

\section{Introduction}
Individualized decision making based on a subject's characteristics has received increasing attention in the past decades. It has broad application in many fields, such as establishing personalized treatment regimes based on electronic health record data \citep{wang2016learning}, personalizing ad placement based on user profiles \citep{bottou2013counterfactual}, and optimally allocating public resources based on individual characteristics \citep{kube2019allocating}. The common goal of these problems is to find an \textit{individualized treatment rule} (ITR), a mapping from individual characteristics to a treatment assignment such that the expected outcome is optimized under the rule \citep{murphy2001marginal,murphy2003optimal,robins2004optimal}. In order for the estimated ITR to be practically useful, it is usually desirable to limit the search of the optimal ITR within a prespecified class of restricted rules. Such restriction may arise from logistic, legal, ethical or political considerations \citep{kitagawa2018should,athey2021policy}, or from the need for an interpretable ITR to inform better clinical practice; see, for example, \citet{zhang2018interpretable,murdoch2019definitions}.

A wide array of statistical methods have been developed to estimate the optimal ITR using experimental or observational data \citep{kosorok2019precision} since the pioneer work of \citet{murphy2001marginal}, \citet{murphy2003optimal}, and \citet{robins2004optimal}. One can model observed outcomes as functions of the covariates and then obtains the optimal ITR by inverting the regression estimates \citep{qian2011performance, foster2011subgroup}.  Of course, the success of these methods depends on accuracy of the outcome models. Alternatively, weighting-based ITR learning approaches obviate or reduce the need for fitting outcome models by directly optimizing the expected ITR outcome or value function \citep{zhao2012estimating,zhou2017residual,chen2017general}. A critical component of this class of methods is a weighting function that reweights the source data based on the covariates and treatment information such that the treated and control groups are comparable. Nearly all such methods use inverse probability (of treatment) weighting for this task. Inverse probability weighting makes the treated and control groups look as though they followed the same covariate distribution as the whole source sample. 

Despite the vast literature in estimating the optimal ITR, much less attention has been paid to its generalizability to a target population with the exception of the work by \citet{zhao2019robustifying}, and \citet{mo2020learning}. In contrast, there is a surge of publications for average treatment effect generalization (e.g., \cite{Cole2010,tipton2013improving, buchanan2018generalizing,dahabreh2020extending}). Compared with causal generalization for the average treatment effect which is a scalar quantity, ITR generalization poses new challenges due to the need to model and generalize the rules based on a prespecified class of functions which may not contain the unrestricted true optimal ITR. The aim of this paper is to develop a weighting framework to mitigate the impact of such misspecification thus facilitate the generalizability of optimal ITR from a source population to a target population. Therefore our framework needs to be coupled with existing ITR learning methods for estimation. For this setting, our work is different from the goal of prior works \citep{zhao2019robustifying,mo2020learning} which focused on developing robust (thus generalizable) ITRs. 

The distributional difference in covariates between the two populations is termed as \textit{covariate shift} in the literature \citep{sugiyama2007covariate}. The unrestricted optimal ITR can be determined from the conditional average treatment effect $\tau(x)$ which is the average outcome difference under two treatment for a given patient characteristics $x$ (defined more precisely in Section \ref{sec:bal_trt_ctrl}). On the other hand, the best-in-class ITR, unlike the optimal ITR over all possible function classes, depends on the distribution of covariates. Thus, the weights in this setting ensure that the best-in-class ITR applies to the target population rather than the source population.

\begin{figure}[!ht]
	\centering
		\includegraphics[width=.34\linewidth]{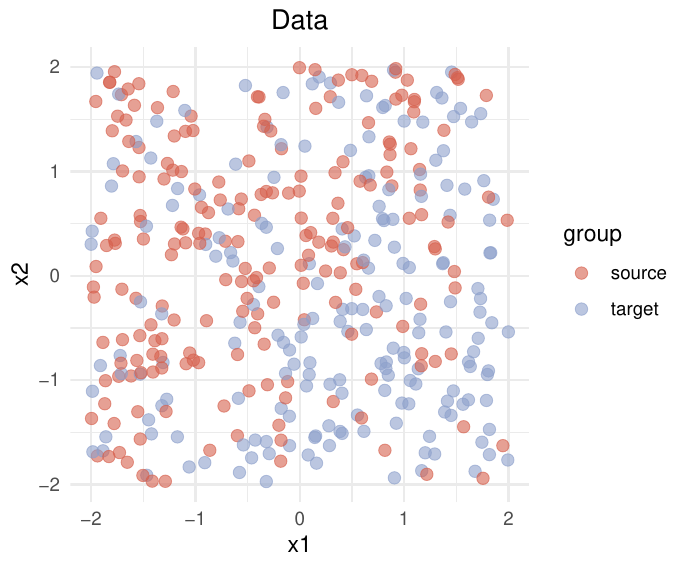}
	\includegraphics[width=.65\linewidth]{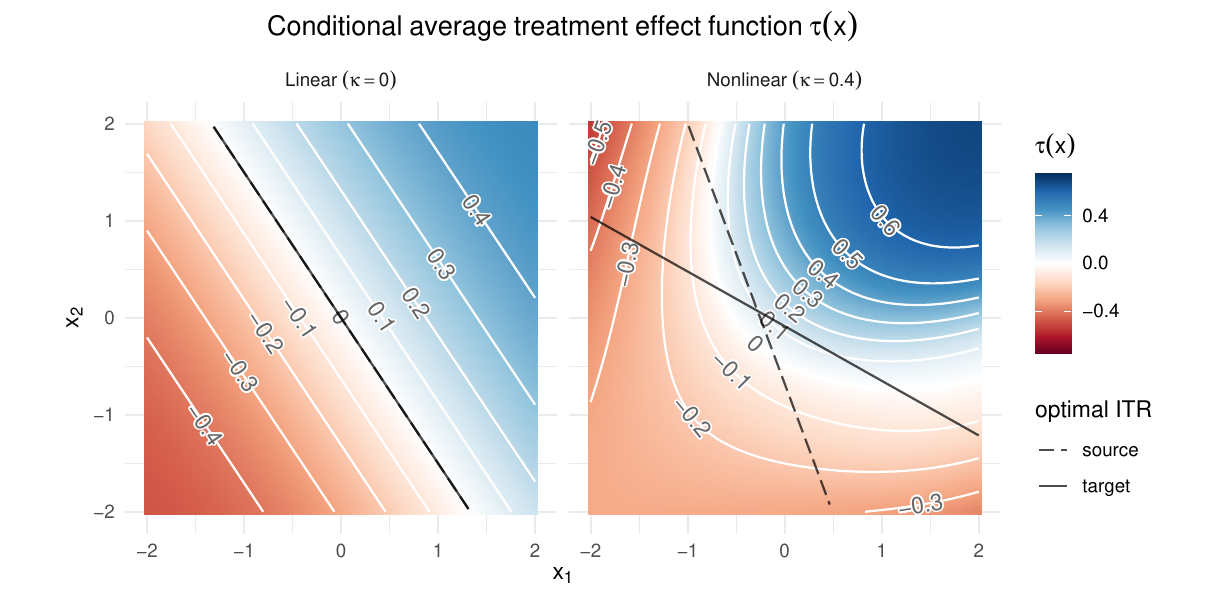}
	\captionsetup{width=1\linewidth}
	\caption{Left: covariate distribution of source and target populations with covariate shift; Right: contour plot of the conditional average treatment effect function $\tau(x)$ and the lines are the optimal linear decision boundaries, with the solid line indicating the boundary for the optimal linear ITR for the target population and the dashed line indicating the boundary for the optimal linear ITR on the source population.}
	\label{fig:simu_setting_illus}
\end{figure}

To illustrate the interplay between ITR class misspecification and covariate shift, we consider  the following example for  learning a linear ITR under two scenarios of $\tau(x)$ when covariate shift is present in Figure \ref{fig:simu_setting_illus}. Here, two covariates, $X_1$ and $X_2$, determine the value of $\tau(x)$ (see Section \ref{sec:simu} for the explicit formula). In the first scenario, the unrestricted optimal ITR is a linear function of the covariates, hence the optimal within-class ITRs (optimal linear ITR) are the same for source and target populations. In other words, the issue of covariate shift does not need to be addressed since the prespecified rule class is correctly specified. In the second scenario, the unrestricted optimal ITR is a nonlinear function of the covariates, hence the optimal within-class ITR (optimal linear ITR) is different for the source and target populations. Hence, using the target population's covariate information is critical to learn an optimal within-class ITR for the target population.

Weighting adjustment by the \textit{importance weights} has been a commonly used approach for dealing with covariate shift \citep{sugiyama2007covariate}. In the context of ITR learning, the source data are reweighted by the density ratio between the target and source populations on top of the inverse probability weighting, as suggested by \citet[Remark 2.2]{kitagawa2018should}. However the importance weights can contain extreme values especially when there is insufficient overlap between the treated and control groups or between the source and target populations, leading to unstable estimation.
\citet{kallus2020more} proposed to alleviate the instability issue in this context by retargeting to a population whose density ratio against the source population is proportional to $\pi(x)\{1-\pi(x)\}$, where $\pi(x)$ denotes the propensity score. This retargeted ``hypothetical'' population allocates more emphasis on the covariate region where the treated and control groups have good overlap. When combined with inverse probability weighting, the resulting weights are essentially the \textit{overlap weights} of \citet{li2018balancing}. \citet{kallus2020more} showed that when the prespecified rule class can well approximate the unrestricted true optimal ITR, their retargeting approach incurs no bias to the learning process and can lead to better performing ITRs, and thus can work well if one expects certain amount of covariate shift in the target population. However, when the unrestricted true optimal ITR is \textit{not} contained in the prespecified rule class, this claim no longer holds because the retargeted population can be substantially different from the target population of interest, and thus their restricted optimal ITRs within the prespecified rule class also differ. Therefore, practitioners interested in generalizing an ITR to a new population are presented with two extreme weighting methods: the importance weights that strive for eliminating bias from confounding and covariate shift but may lead to large variance, and the overlap weights that emphasize robust and stable estimation but are prone to bias as they only mitigate confounding. In this paper, we aim to develop a unifying weighting framework that not only encompasses the importance and the overlap weights, but also allows one to search a better bias-variance trade-off between them in a data-driven manner.

Another increasingly recognized issue of these existing weighting methods is that they are constructed in an indirect way based on the propensity score and density ratio. In practice these functions need to be estimated and can be extremely sensitive to model misspecification \citep{kang2007demystifying}. Our proposed method addresses this issue by constructing the weights directly,  bypassing the need for direct modeling of the propensity score or density ratio. Our approach to doing so is inspired by the recent growing literature of covariate balancing weights for estimation of the aggregate causal estimands such as the average treatment effect \citep{hainmueller2012entropy,imai2014covariate,wong2018kernel,hirshberg2019minimax,huling2020energy, chattopadhyay2020balancing, kallus2020generalized,kallus2022optimal}. Instead of explicitly specifying a functional form for the underlying propensity score model, covariate balancing approaches directly construct weights so that they improve the balance between treatment groups over a given set of covariate functions. 
The variation of the weights is typically controlled by minimizing the dispersion of the weights subject to the specified balance constraints for these covariate functions.

As we will show, proper weights for ITR learning with possibly misspecified rule classes should be able to achieve balance on a possibly very complex set of covariate functions to cover candidate assignment rule classes. We start by establishing a covariate balancing weight strategy that minimizes the imbalance between the treated group, the control group, and the target population over a non-parametric function class characterized by a reproducing kernel Hilbert space (RKHS, see \citet{wahba1990spline}). By carefully inspecting the target population value function, we observe that different from adjusting for confounding, adjusting for covariate shift is not essential when the prespecified ITR class contains the unrestricted true optimal ITR. As we cannot know such containment in practice, we propose a three-way balancing strategy based on a convex combination of objectives that seek to induce covariate shift balance and confounding balance. As our weighting approach emphasizes full distributional balance of covariates, it is robust to the functional form of confounding and works well empirically. 
Despite our weights being non-parametrically constructed through balance-seeking objectives, we show that the resulting weights converge to either the importance weights or the overlap weights under different hyperparameters when sample size tends to infinity. To the best of our knowledge, this is the first result that characterizing the asymptotic behavior of weights that optimize a balancing criterion under the causal generalization setting. In contrast, previous RKHS based balancing weights methods \citep{wong2018kernel,kallus2020generalized,kallus2022optimal} focused on estimating the average treatment effect, and they only provide asymptotic results for the corresponding weighted average treatment effect estimator but not for the weights themselves. Moreover, the proposed method has potential real-world applications, such as generalizing ventilation rules for COVID-19 patients across diverse geographic locations that exhibit substantial covariate shifts in factors like age and pre-existing conditions  \citep{kyono2023selecting}.

\section{Preliminaries}
\label{sec:framework}

Suppose we have collected data on $n$ subjects: $(X_i, S_i, S_i A_i, S_i Y_i)$, $i = 1, ..., n$. $X_i \in \mathcal{X} \subset \mathbb{R}^p$ denotes a vector of pre-treatment covariates which contain confounding factors and treatment effect modifiers. $S_i$ is a population indicator such that $S_i = 1$ indicates subject $i$ is from the source population $\mathcal{S}$ and $S_i = 0$ indicates $i$ is from the target population $\mathcal{T}$. $A_i \in\{0, 1\}$ is a binary treatment indicator and $Y_i$ is the outcome of interest for which we assume larger outcomes are preferred without loss of generality. In our setting, $A_i$ and $Y_i$ are only observed for observations from the source population ($S_i = 1$). We will refer to the observations from $\mathcal{S}$ as the source sample and the rest as the target sample. Let $n_s$, $n_t$ be the sample sizes of the source and target samples, respectively, and $n = n_s + n_t$ be the total sample size.

For causal interpretation, we use the potential outcome framework \citep{rubin1974estimating,rosenbaum1983central}, wherein each subject $i$ is assumed to have two potential outcomes  $Y_i(1)$, the outcome if subject $i$ were to receive the treatment, and $Y_i(0)$, the outcome if $i$ were not to receive the treatment. Under the assumption known as the Stable Unit Treatment Value Assumption, the observed outcome in the source sample can be expressed as $Y_i = Y_i(A_i)$. Under this assumption, we can associate each observation with a ``full'' random variate $(X_i, S_i, A_i, Y_i(0), Y_i(1))$, which across $i$ are assumed to be i.i.d. draws from a super-population.

Consider a prespecified ITR class $\mathcal{D} \subset \{0, 1\}^\mathcal{X}$. For an ITR $d \in \mathcal{D}$, the potential outcome following the treatment assignment prescribed by $d$ is $Y(d(X))$. The utility of the ITR $d$ can be measured by the \textit{value function}, which is defined as the population mean outcome were all subjects to receive treatment according to $d$. Specifically, the value function under the source distribution and the target distribution are
\begin{equation*}
    \mathcal{V}_\mathcal{S}(d) = \E[Y(d(X)) \mid S = 1]
    \quad\text{and}\quad
    \mathcal{V}_\mathcal{T}(d) = \E[Y(d(X)) \mid S = 0]
,\end{equation*}
respectively. The objective is then to find the optimal ITR in $\mathcal{D}$ for the target population, i.e. the one that maximizes $\mathcal{V}_\mathcal{T}(d)$ among all rules in the prespecified rule class:
\begin{equation}
    \label{eq:objective}
    \argmax_{d \in \mathcal{D}} \; \mathcal{V}_\mathcal{T}(d)
.\end{equation}

We assume that the treatment assignment mechanism in the source sample is determined by a \textit{propensity score} $\pi(x) = \prob (A = 1 \mid X = x, S=1)$ \citep{rosenbaum1983central}.  If the source sample is from a randomized trial, then $\pi(x)$ is known.
 We further denote $\rho(x) = \prob(S=1 \mid X=x)$ and refer to this as \textit{participation probability} \citep{dahabreh2020extending}. To identify $\mathcal{V}_\mathcal{T}(d)$ from the observed data, we impose the following assumptions throughout the paper.

\begin{assumption}
\label{assp:unconf}
    (Unconfoundedness of treatment assignment)
    In the source population, $(Y(0), Y(1))$ are conditionally independent of $A$ given $X$:
    $(Y(0), Y(1)) \independent A \mid X, S=1$.
\end{assumption}
\begin{assumption}
\label{assp:ovlptrt}
    (Positivity of propensity score)
    The propensity score of the source population is bounded away from 0 and 1: for some $c > 0$, $c \le \pi(X) \le 1 - c$ almost surely. 
\end{assumption}

\begin{assumption}
\label{assp:exchange}
    (Mean exchangability across populations)
    The conditional mean of the potential outcomes given the covariates are equal between the two populations: $\E [Y(a) \mid X, S=1] = \E [Y(a) \mid X, S=0]$ almost surely for $a \in \{0, 1\}$.
\end{assumption}
\begin{assumption}
\label{assp:ovlppop}
    (Positivity of participation probability)
    The participation probability is bounded away from 0: $\rho(X) > c$ almost surely for some $c > 0$. 
\end{assumption}

The first two are common assumptions in causal inference; and together with the SUTVA enable identification of causal quantities with respect to the source population, such as $\mathcal{V}_\mathcal{S}(d)$, from the observed sample. The last two assumptions are adopted from \citet{dahabreh2020extending} and \citet{Rudolph2017} and allow us to generalize causal estimates to the target population \citep{colnet2020causal}. In particular, Assumption $3$ is a strong but necessary assumption. Without this assumption, it is impossible to learn meaningful ITRs for a target population using the data from the source population as the optimal ITRs for these two populations can be completely unrelated.  

Under these assumptions, the value function on the target population can be identified with the observed source sample quantities through
\begin{align}
    \mathcal{V}_\mathcal{T}(d) 
    &=
    \E\left[ 
        w^*(A, X) \ind \{d(X) = A\} Y \mid S = 1
     \right]
    \nonumber
    \\&=
    \E\left[ A w^*(1, X) d(X) Y \mid S = 1 \right] + 
    \E\left[ (1-A) w^*(0, X) \{ 1-d(X) \} Y \mid S = 1 \right]
    \label{eq:weighted_estimator}
\end{align}
where $\ind(\cdot)$ is the indicator function and
\begin{equation}
    \label{def:w_star_a_x}
    w^*(a, x) = 
    \left\{\frac{a}{\pi(x)} + \frac{1-a}{1-\pi(x)} \right\}
    \frac{\E(S)\{ 1 - \rho(x)\}}{\{ 1-\E(S) \} \rho(x)} 
.\end{equation}
For $a = 1$ (or 0), $w^*(1, x)$ (or $w^*(0, x)$) is proportional to the density ratio between the target population and the treated group (or control group). In other words, $w^*(a, x)$ acts as the importance weight that reweights the underlying distribution of the source data to the target distribution. It can be estimated from data because $\E(S)$ can be estimated as the empirical average of $S_i$ and $\pi(x), \rho(x)$ can be estimated by probability models for binary outcomes such as logistic regression. However, since the probability estimates appear in the denominator of $w^*(a, x)$, estimation error and model misspecification of the probability models can lead to significant error in estimating $\mathcal{V}_\mathcal{T}(d)$ and thus jeopardize the ITR learning process. In the next section, we propose approaches to directly estimate weights that are suitable for ITR learning specifically.


\section{Method} \label{sec:method}
Let $w: \{0, 1\} \times \mathcal{X} \rightarrow [0, \infty)$ denote a general weighting function. Let $\mathcal{S}_a = \{i: S_i = 1, A_i = a\}$ be the index set of treatment group $a$ among the source sample, $a \in \{0, 1\}$. By abuse of notation we let $\mathcal{S}$ also denote $\{i: S_i = 1\}$ and similarly $\mathcal{T} = \{i: S_i = 0\}$. Following \eqref{eq:weighted_estimator} we consider estimators for $\mathcal{V}_\mathcal{T}(d)$ of the form:
\begin{equation}
    \label{eq:hat_V_n}
    \widehat{\mathcal{V}}_n(d; w) = 
    \frac{1}{n_s} \sum_{i\in \mathcal{S}_1}  w_i d(X_i) Y_i +
    \frac{1}{n_s} \sum_{i\in \mathcal{S}_0}  w_i \{ 1-d(X_i) \} Y_i
,\end{equation}
where the weights $w_i$ are normalized as
\begin{equation*}
    w_i = \begin{cases}
        \dfrac{w(1, X_i)}{\tfrac{1}{n_s} \sum_{i \in \mathcal{S}_1} w(1, X_i)}, &\text{if } i \in \mathcal{S}_1, \vspace{6pt} \\
        \dfrac{w(0, X_i)}{\tfrac{1}{n_s} \sum_{i \in \mathcal{S}_0} w(0, X_i)}, &\text{if } i \in \mathcal{S}_0.
    \end{cases}
\end{equation*}
The weights are normalized by treatment group, i.e., $\sum_{i \in \mathcal{S}_1} w_i = \sum_{i \in \mathcal{S}_0} w_i = n_s $. Such weighted estimators are also known as the H\'ajek estimator \citep{chattopadhyay2020balancing}. Given the weights, the optimal ITR can be estimated as 
\begin{equation}
    \label{eq:weighting_learning}
    \hat{d} = \argmax_{d \in \mathcal{D}} \widehat{\mathcal{V}}_n(d; w)
.\end{equation}
When $w(1, x) = 1 / \pi(x)$ and $w(0, x) = 1 / \{ 1 - \pi(x) \}$, \eqref{eq:hat_V_n} becomes the objective function of outcome weighted learning approach \citep{zhao2012estimating}. 

In this section, we will study the requirements on $w(\cdot)$ or $w_i$ such that maximizing $\widehat{\mathcal{V}}_n(d; w)$ produces a well-performing ITR for the target population and will then develop a balancing weight framework to meet these conditions. 
We first derive two different balancing weight methods, one that makes a covariate shift adjustment in addition to adjusting for confounding and the other that only adjusts for confounding. We will demonstrate that each of these two approaches is suitable under different situations, depending on whether or not the prespecified rule class contains the unrestricted optimal ITR. However, it is not clear \textit{a priori} which of the weighting approaches is best suited for a given data-generating mechanism. To address this, 
our final approach will be formulated as a hybrid of these two balancing weighting methods that combines them in a data-driven manner.

\subsection{General case: balancing to the target population}
\label{sec:bal_to_target}

In this section we focus on the general case where the true optimal ITR is potentially not contained in the prespecified class $\mathcal{D}$. In this case we investigate what properties a set of weights should have in order to reduce error in estimation of the optimal ITR.
From the perspective of ITR evaluation \citep{kallus2018balanced}, a natural idea is to choose $w$ for the source population so that make $\widehat{\mathcal{V}}_n(d; w)$ is as close to $\mathcal{V}_\mathcal{T}(d)$ as possible for all $d \in \mathcal{D}$. Given such weights, an ITR $d$ that is the maximizer of $\widehat{\mathcal{V}}_n(d; w)$ will also achieve a high value of $\mathcal{V}_\mathcal{T}(d)$.

Define $\mu_a(x) = \E[Y(a) \mid X = x]$ and $\varepsilon_i \equiv Y_i - \mu_{A_i}(X_i)$. We assume that $\varepsilon_i$ are mutually independent and have uniformly bounded variance, i.e., $\E[\varepsilon_i^2 \mid X_i, A_i] \le \sigma^2 < \infty$. Let $\mathcal{V}_{n,\mathcal{T}}(d) = \sum_{i \in \mathcal{T}} [d(X_i) \mu_1(X_i) + \{ 1 - d(X_i) \} \mu_0(X_i) ]/ n_t$ be the sample version of $\mathcal{V}_\mathcal{T}(d)$, where $n_t = \sum_i (1 - S_i)$. Then the estimation error $\widehat{\mathcal{V}}_n(d; w) - \mathcal{V}_\mathcal{T}(d)$ admits the following decomposition:
\begin{equation}
   \label{eq:err_decomp_s_t}
    \begin{aligned} 
        \widehat{\mathcal{V}}_n(d; w) - \mathcal{V}_\mathcal{T}(d)
        &=
        \frac{1}{n_s} \sum_{i \in \mathcal{S}_1} w_i d(X_i) \{ \mu_1(X_i) + \varepsilon_i \} +
        \frac{1}{n_s} \sum_{i \in \mathcal{S}_0} w_i \{ 1-d(X_i) \} \{ \mu_0(X_i) + \varepsilon_i \}
        \\&\quad -
        \frac{1}{n_t} \sum_{i \in \mathcal{T}} d(X_i) \mu_1(X_i) -
        \frac{1}{n_t} \sum_{i \in \mathcal{T}} \{ 1 - d(X_i) \} \mu_0(X_i) +
        \big\{ \mathcal{V}_{n,\mathcal{T}}(d) - \mathcal{V}_\mathcal{T}(d) \big\}
        \\&=
        \int_\mathcal{X} d(x) \mu_1(x) \textnormal{d} \big( \prob_{n,\mathcal{S}_1}^{(w)} - \prob_{n,\mathcal{T}}\big)(x) + 
        \int_\mathcal{X} \{ 1-d(x) \} \mu_0(x) \textnormal{d} \big( \prob_{n,\mathcal{S}_0}^{(w)} - \prob_{n,\mathcal{T}}\big)(x)
        \\&\quad +
        \frac{1}{n_s}\sum_{i \in \mathcal{S}} \ind \{ d(X_i)=A_i \} w_i \varepsilon_i +
        \big\{ \mathcal{V}_{n,\mathcal{T}}(d) - \mathcal{V}_\mathcal{T}(d) \big\}  
    .\end{aligned}
\end{equation}
Here $\prob_{n,\mathcal{S}_a}^{(w)} = \sum_{i \in \mathcal{S}_a} w_i \delta_{X_i}/ n_s$ are the weighted empirical distributions of the covariates of the treated ($a=1$) and control ($a=0$) units in the source sample, and $\prob_{n,\mathcal{T}}$ is the empirical distribution of the covariates of the target sample. 

The decomposition \eqref{eq:err_decomp_s_t} allows a transparent understanding about what properties the $w_i's$ should possess to ensure the estimation error of the value function in the target population is small. The first term in the decomposition is the source of estimation bias arising from the imbalance between $\prob_{n,\mathcal{S}_1}^{(w)}$ and $\prob_{n,\mathcal{T}}$ on the function $d(x)\mu_1(x)$. In other words, depending on the behavior of the unknown quantity $d(x)\mu_1(x)$, this term can contribute to the bias if the weighted distribution of covariates of treated group in the source sample differs from the empirical distribution of covariates in the target sample. Similarly, the second term is due to the imbalance between $\prob_{n,\mathcal{S}_0}^{(w)}$ and $\prob_{n,\mathcal{T}}$ on $\{ 1-d(x) \}\mu_0(x)$. The third term demonstrates how variance is inflated by the weighting function $w(a, x)$. These three terms can be controlled by properly selecting the weighting function $w(a, x)$. The last term on the right-hand side of \eqref{eq:err_decomp_s_t}  is completely due to sampling error and is not related to $w(a,x)$ in any manner. It converges to 0 at the rate of $\mathcal{O}\big(n^{-1/2} \big)$ as long as the target sample is representative of the target population. 

We first consider the first term on the right-hand side of \eqref{eq:err_decomp_s_t}. Ideally, we want to eliminate or mitigate the bias by balancing $\prob_{n,\mathcal{S}_1}^{(w)}$ and $\prob_{n,\mathcal{T}}$ on $d(x)\mu_1(x)$. However, $\mu_1(x)$ is typically unknown. Even if $\mu_1(x)$ can be well approximated by a single function, multiplication by $d(x)$ with $d(x)$ varying over the ITR class $\mathcal{D}$ can induce a highly complex set of functions. Therefore, we consider the maximum imbalance between $\prob_{n,\mathcal{S}_1}^{(w)}$ and $\prob_{n,\mathcal{T}}$ over a large and flexible class of functions. Specifically, we choose this function class to be the unit ball of an RKHS $\mathcal{H}$ associated with a positive definite kernel function $K(x, x')$. We will restrict our attention to universal and bounded kernel functions \citep{simon2018kernel}. This includes many commonly used kernel functions, such as the Gaussian kernel and the Mat\'ern kernel. This maximum or worst-case imbalance is also commonly referred to as the \textit{maximum mean discrepancy} (MMD) \citep{gretton2012kernel}:

\begin{equation}
    \label{eq:mmd_def}
    \textnormal{MMD}_\mathcal{H}(\prob_{n,\mathcal{S}_1}^{(w)}, \prob_{n,\mathcal{T}})
    = 
    \sup_{h \in \mathcal{H}, \| h \|_\mathcal{H} \le 1}
    \int_\mathcal{X} h(x) \textnormal{d} \big( \prob_{n,\mathcal{S}_1}^{(w)} - \prob_{n,\mathcal{T}}\big)(x) 
.\end{equation}

Although $d(x)\mu_1(x)$ is typically not itself a member of the RKHS $\mathcal{H}$ due to the discrete nature of $d(x)$, \eqref{eq:mmd_def} still provides a \textit{probabilistic} bound for the imbalance between $\prob_{n,\mathcal{S}_1}^{(w)}$ and $\prob_{n,\mathcal{T}}$ on $d(x)\mu_1(x)$. Specifically, if $\mu_1 \in \mathcal{H}$ and the kernel function takes the form of $K(x, x') = \theta^{-p} K_0(\|x-x'\| / \theta)$ then under suitable regularity conditions provided in the Supplementary Material, we can show that the following 
\begin{equation}
    \label{eq:bound_mmd_d_mu}
    \int_\mathcal{X} d(x) \mu_1(x) \textnormal{d} \big( \prob_{n,\mathcal{S}_1}^{(w)} - \prob_{n,\mathcal{T}}\big)(x)
    \le
    \| \mu_1 \|_\mathcal{H}
    \textnormal{MMD}_\mathcal{H}(\prob_{n,\mathcal{S}_1}^{(w)}, \prob_{n,\mathcal{T}}) +
    C_1 \left( \frac{1}{n \theta^p} + \theta^{\gamma/2} \right)
\end{equation}
holds with probability at least $1 - \exp\{-C_2 n \theta^{p + \gamma}\}$, where $\gamma, C_1, C_2$ are constants not related to $d(x)$. The derivation of this result is based on \citet{gretton2012kernel} and is deferred to the Supplementary Material. This observation motivates us to find a set of appropriate weights by minimizing $\textnormal{MMD}_\mathcal{H}(\prob_{n,\mathcal{S}_1}^{(w)}, \prob_{n,\mathcal{T}})$.
\citet{gretton2012kernel} showed that when $\mathcal{X}$ is a compact space, $\textnormal{MMD}_\mathcal{H}$ is indeed a metric of the probability distributions on $\mathcal{X}$. Therefore, by minimizing $\textnormal{MMD}_\mathcal{H}(\prob_{n,\mathcal{S}_1}^{(w)}, \prob_{n,\mathcal{T}})$ with respect to the weights, we are in essence searching for a set of weights that make the treated group in the source sample close to the target sample and further due to \eqref{eq:bound_mmd_d_mu}, this minimization can help mitigate one of the key sources of systematic bias, i.e. the first term in the decomposition \eqref{eq:err_decomp_s_t}.
By a similar argument we propose to control the other key source of systematic bias, the second term of \eqref{eq:err_decomp_s_t}, by optimizing $\textnormal{MMD}_\mathcal{H}(\prob_{n,\mathcal{S}_0}^{(w)}, \prob_{n,\mathcal{T}})$.

To mitigate the influence of the third term in \eqref{eq:err_decomp_s_t}, it is vital to manage the magnitude of $w_i$. This can often be accomplished by minimizing a dispersion measure of $w_i$ as seen in covariate balancing weights literature \citep{hainmueller2012entropy,chan2016globally,wang2020minimal}. We opt to penalize a large $\sum_{i \in \mathcal{S}} w_i^2/ n_s^2 $ for several reasons. First, the conditional mean square of the third term of \eqref{eq:err_decomp_s_t} given $\{X_i, S_i, A_i\}$, $\sum_{i \in \mathcal{S}} w_i^2 \ind \{d(X_i) = A_i \} \var(\varepsilon_i \mid X_i, S_i, A_i)/ n_s^2$, is bounded by $ \sigma^2 \sum_{i \in \mathcal{S}}  w_i^2/ n_s^2$. So $\sum_{i \in \mathcal{S}} w_i^2/  n_s^2$ directly relates to the magnitude of this term. Second, for weighted data the effective sample size is usually approximated as $(\sum_i w_i)^2 / \sum_i w_i^2$ \citep{kish1965survey}, which is equal to $4 n_s^2 / \sum_{i \in \mathcal{S}} w_i^2$ in our case due to the normalizing constraints. Thus, $\sum_{i \in \mathcal{S}} w_i^2/n_s^2$ has a clear interpretation as the inverse sample size and has been advocated for diagnosing the stability of weighting approaches \citep{chattopadhyay2020balancing}.

Combining all the elements discussed, we propose to estimate weights via the optimization
\begin{equation}
    \label{eq:opt_prog_s_t}
    \begin{aligned}
    \min_w &\qquad
    \textnormal{MMD}_\mathcal{H}(\prob_{n,\mathcal{S}_1}^{(w)}, \prob_{n,\mathcal{T}})^2 + 
    \textnormal{MMD}_\mathcal{H}(\prob_{n,\mathcal{S}_0}^{(w)}, \prob_{n,\mathcal{T}})^2 +
    \frac{\lambda}{n_s^2} \sum_{i \in \mathcal{S}} w_i^2,
    \\
    \textnormal{subject to}&\qquad
    \sum_{i \in \mathcal{S}_1} w_i = \sum_{i \in \mathcal{S}_0} w_i = n_s
    ,\end{aligned}
\end{equation}
where $\lambda > 0$ is a hyperparameter controlling the level of penalty. Here we take square of the $\textnormal{MMD}_\mathcal{H}$ terms for computational convenience, which will be explained in Section S2.1 of the Supplementary Material.

\subsection{Special case with correctly specified rule class: balancing within the source population}
\label{sec:bal_trt_ctrl}
In the previous section, we discussed a strategy to find an appropriate weighting function by making $\widehat{\mathcal{V}}_n(d; w)$ a good estimator for the value function $\mathcal{V}_\mathcal{T}(d)$ without any consideration for whether the prespecified ITR class $\mathcal{D}$ provided a reasonable approximation to the true optimal ITR. The resulting optimization problem minimizes the distance between $\prob_{n,\mathcal{S}_a}^{(w)}$ and $\prob_{n,\mathcal{T}}$ subject to a smoothing penalty. However, since the objective is to find a good ITR rather than to estimate $\mathcal{V}_\mathcal{T}(d)$, such an intermediate objective is not always necessary. In this section, we consider the situation where the unrestricted true optimal ITR $d^*(x) = \ind \{\mu_1(x) \ge \mu_0(x) \}$ is contained in $\mathcal{D}$. In this case, the optimal ITR does not depend on the distribution of the covariates, obviating the need for covariate shift adjustments that make $\prob_{n,\mathcal{S}_a}^{(w)}$ similar to $\prob_{n,\mathcal{T}}$. As a consequence, we will see that the balancing requirements can be relaxed, thus allowing for use of a balancing criterion that yields less variable weights compared with the general case in Section \ref{sec:bal_to_target}.

To illustrate the key source of error in estimating the optimal ITR when $\mathcal{D}$ is well-specified, we investigate conditions under which the maximizer of $\widehat{\mathcal{V}}_n(d; w)$ will be close to $d^*$. A sufficient condition of such is to ensure that $\widehat{\mathcal{V}}_n(d, w)$ is only be greater than $\widehat{\mathcal{V}}_n(d^*, w)$ when $d$ is close to $d^*$. In other words, $\widehat{\mathcal{V}}_n(d^*; w) - \widehat{\mathcal{V}}_n(d; w)$ must be positive whenever $d$ differs substantially from $d^*$.
Let $m(x) = \{\mu_0(x) + \mu_1(x)\}/ 2$ denote the main effect function and $\tau(x) = \mu_1(x) - \mu_0(x)$ denote the conditional average treatment effect function. Let us consider the following decomposition
\begin{equation}
    \label{eq:diff_decomp_tau}
    \begin{aligned}
        \widehat{\mathcal{V}}_n(d^*; w) - \widehat{\mathcal{V}}_n(d; w)
        &= 
        \frac{1}{n_s} \sum_{i\in \mathcal{S}_1}  w_i \{d^*(X_i) - d(X_i)\} \left\{m(X_i) + \frac{\tau(X_i)}{2} + \varepsilon_i\right\} 
        \\&\quad -
        \frac{1}{n_s} \sum_{i\in \mathcal{S}_0}  w_i \{d^*(X_i) - d(X_i)\} \left\{m(X_i) - \frac{\tau(X_i)}{2} + \varepsilon_i\right\}
        \\&=
        \frac{1}{2} \int_\mathcal{X} \{d^*(x) - d(x)\} \tau(x) \textnormal{d} \big( \prob_{n,\mathcal{S}_1}^{(w)} + \prob_{n,\mathcal{S}_0}^{(w)}\big)(x)
        \\&\quad+
        \int_\mathcal{X} \{d^*(x) - d(x)\} m(x) \textnormal{d} \big( \prob_{n,\mathcal{S}_1}^{(w)} - \prob_{n,\mathcal{S}_0}^{(w)}\big) (x) 
        \\&\quad+
        \frac{1}{n_s} \sum_{i \in \mathcal{S}} (2A_i - 1) \{d^*(X_i) - d(X_i)\} w_i \varepsilon_i
    .\end{aligned}
\end{equation}

Since $d(x)$ and $d^*(x)$ only take value in $\{0, 1\}$, $d^*(x) - d(x) \in \{0, \pm1\}$. When $\mathcal{D}$ contains the unrestricted optimal ITR, $d^*(x)$ can be expressed as $\ind \{ \tau(x) \ge 0 \}$, so $\tau(x)$ is non-negative when $d^*(x) - d(x) = 1$ and negative when $d^*(x) - d(x) = -1$. Consequently,  the first term on the right-hand side of \eqref{eq:diff_decomp_tau} is always non-negative, and it will take a large positive value if $d(x)$ is substantially different from $d^*(x)$, especially when they disagree on the region in the covariate space where $\tau(x)$ is of large magnitude. On the contrary, the signs of the second and the last terms in \eqref{eq:diff_decomp_tau} are uncertain. If the magnitude of these two terms can be controlled, then $\widehat{\mathcal{V}}_n(d^*; w) - \widehat{\mathcal{V}}_n(d; w)$ can only be positive when $d$ is sufficiently close to $d^*$.

The second term arises from the imbalance between $\prob_{n,\mathcal{S}_1}^{(w)}$ and $\prob_{n,\mathcal{S}_0}^{(w)}$ on $\{d^*(x) - d(x)\} m(x)$. Using similar arguments as in the previous section, we propose to control this quantity by optimizing $\textnormal{MMD}_\mathcal{H}(\prob_{n,\mathcal{S}_1}^{(w)}, \prob_{n,\mathcal{S}_0}^{(w)})$. The last term in the decomposition always has mean zero and involves the variability of the weights; so to mitigate this noise term, we again use a penalty term $ \lambda \sum_{i \in \mathcal{S}} w_i^2/ n_s^2$. Hence, we obtain the following optimization problem

\begin{equation}
    \label{eq:opt_prog_trt_ctrl}
    \begin{aligned}
    \min_w &\qquad
    \textnormal{MMD}_\mathcal{H}(\prob_{n,\mathcal{S}_1}^{(w)}, \prob_{n,\mathcal{S}_0}^{(w)})^2 + 
    \frac{\lambda}{n_s^2} \sum_{i \in \mathcal{S}} w_i^2,
    \\
    \textnormal{subject to}&\qquad
    \sum_{i \in \mathcal{S}_1} w_i = \sum_{i \in \mathcal{S}_0} w_i = n_s
    .\end{aligned}
\end{equation}

The balancing objective in \eqref{eq:opt_prog_trt_ctrl} is to make $\prob_{n,\mathcal{S}_1}^{(w)}$ and $\prob_{n,\mathcal{S}_0}^{(w)}$ close, but does not explicitly enforce them towards any specific distribution. In Section \ref{sec:method_weight_limit} we formally establish that the resulting reweighted distribution is one that focuses on the region in the covariate space where the two treatment groups have good overlap, similar to the notion of overlap weights \citep{li2018balancing}.

Since $\textnormal{MMD}_\mathcal{H}$ is a metric of probability distributions, it follows from the triangle inequality that $\textnormal{MMD}_\mathcal{H}(\prob_{n,\mathcal{S}_1}^{(w)}, \prob_{n,\mathcal{S}_0}^{(w)}) \le \textnormal{MMD}_\mathcal{H}(\prob_{n,\mathcal{S}_1}^{(w)}, \prob_{n,\mathcal{T}}) + \textnormal{MMD}_\mathcal{H}(\prob_{n,\mathcal{S}_0}^{(w)}, \prob_{n,\mathcal{T}})$. This implies $$\textnormal{MMD}_\mathcal{H}(\prob_{n,\mathcal{S}_1}^{(w)}, \prob_{n,\mathcal{S}_0}^{(w)})^2 \le 2\textnormal{MMD}_\mathcal{H}(\prob_{n,\mathcal{S}_1}^{(w)}, \prob_{n,\mathcal{T}})^2 + 2\textnormal{MMD}_\mathcal{H}(\prob_{n,\mathcal{S}_0}^{(w)}, \prob_{n,\mathcal{T}})^2.$$ Thus, the balancing objective in \eqref{eq:opt_prog_s_t} is inherently more stringent than that in \eqref{eq:opt_prog_trt_ctrl}. As a result, \eqref{eq:opt_prog_trt_ctrl} will generally produce weights with less variability thus translating to a larger effective sample size. Balancing the weighted distribution of all confounders between the treated and control groups is indeed the minimal requirement in ITR estimation and cannot be further relaxed unless data were generated from a randomized controlled trial or no confounding is present. Without such a balance, the ITR obtained from optimizing $\mathcal{V}_{n,\mathcal{T}}(d)$ might recommend a subject to take a treatment not because it corresponds to a better potential outcome but because more people with similar characteristics received the treatment in the observed data.

\subsection{Back to the general case: a three-way balancing approach}
The objective \eqref{eq:opt_prog_s_t} works for general situations but imposes stronger balancing requirements, which may result in more extreme weights, causing the ITR learning process to focus too heavily on a few observations with large weights. On the other hand, the objective \eqref{eq:opt_prog_trt_ctrl} only requires minimal balancing requirements, but the resulting weights are only likely to yield an optimal ITR when $\mathcal{D}$ contains the unrestricted true optimal ITR. Yet, in practice, it is unclear how well $\mathcal{D}$ is specified and one may be unwilling to wager that $\mathcal{D}$ contains the unrestricted true optimal ITR and thus only balance the treated and control groups to each other. To achieve some of the benefits of each approach, we propose to combine these two balancing strategies by optimizing the following convex combination of them
\begin{equation}
    \label{eq:opt_prog_threeway}
    \begin{aligned}
    \min_w &\qquad
    \alpha \textnormal{MMD}_\mathcal{H}(\prob_{n,\mathcal{S}_1}^{(w)}, \prob_{n,\mathcal{T}})^2 + 
    \alpha \textnormal{MMD}_\mathcal{H}(\prob_{n,\mathcal{S}_0}^{(w)}, \prob_{n,\mathcal{T}})^2
    \\&\qquad
    + (1-\alpha) \textnormal{MMD}_\mathcal{H}(\prob_{n,\mathcal{S}_1}^{(w)}, \prob_{n,\mathcal{S}_0}^{(w)})^2 + 
    \frac{\lambda}{n_s^2} \sum_{i \in \mathcal{S}} w_i^2,
    \\
    \textnormal{subject to}&\qquad
    \sum_{i \in \mathcal{S}_1} w_i = \sum_{i \in \mathcal{S}_0} w_i = n_s
    .\end{aligned}
\end{equation}
Here $\alpha \in [0, 1]$ is a hyperparameter controlling the extent to which the weighted treated/control group should resemble the target sample, and $\lambda$ regularizes the smoothness of the weights.	By Lemma 6 of the Supplementary Material, we can write each of the $\textnormal{MMD}_\mathcal{H}$ terms in \eqref{eq:opt_prog_threeway} as a quadratic function of $w$. So the optimization problem is convex and its global optimum exists. We can solve this problem using standard quadratic programming tools. Furthermore, we propose a heuristic procedure to  $\alpha$ and $\lambda$ by grid search in Section S2.2 of the Supplementary Material.

\subsection{Asymptotic properties}
\label{sec:method_weight_limit}

The following theorem characterizes the convergence limit of the proposed weights. Here we impose two conditions on the RKHS: universality and that the unit ball is $\prob$-Donsker. These conditions hold for most commonly used kernels, such as the Gaussian kernel and Mat\'ern kernel \citep{sriperumbudur2016optimal}. 

\begin{theorem}
    \label{thm:weight_limit}
    Suppose $\{\hat{w_i}, i \in \mathcal{S}\}$ are the weights that solve \eqref{eq:opt_prog_threeway}. Assume the RKHS is universal, and the unit ball in it $\mathcal{H}$ is $\prob$-Donsker. Assume $\pi(x), \rho(x)$ are continuous functions on $\mathcal{X}$.
    If $0 < \alpha \le 1$, then
    \begin{equation*}
        \frac{1}{n_s} \sum_{i \in \mathcal{S}}\big \{ \hat{w}_i - w^*(A_i, X_i) \big \}^2 \overset{p}{\longrightarrow} 0
    ,\end{equation*}
    where $w^*(a, x)$ is defined in \eqref{def:w_star_a_x}.
    If $\alpha = 0$,
    \begin{equation*}
        \frac{1}{n_s} \sum_{i \in \mathcal{S}} \big \{ \hat{w}_i - w^\dagger(A_i, X_i) \big \}^2 \overset{p}{\longrightarrow} 0
    .\end{equation*}
    Here $$w^\dagger(a, x) = \frac{\pi(X) \{ 1-\pi(X) \} }{C_\pi} \left\{ \frac{ a}{\pi(x)} + \frac{1-a}{1-\pi(x)} \right\},$$ where $C_\pi = \E[\pi(X) \{ 1-\pi(X) \} \mid S = 1]$ is a normalizing constant.
\end{theorem}

The proofs of this theorem and the other theoretical results below are deferred to the Supplementary Material. The convergence results of the weights straightforwardly imply the convergence limit of $\widehat{\mathcal{V}}_n(d; \hat{w})$, the objective function for learning the optimal ITR:
\begin{corollary}
	\label{thm:vndw_limit}
	Suppose the conditions in Theorem \ref{thm:weight_limit} hold and the potential outcomes are square integrable. Then, for any treatment rule $d$, when $\alpha > 0$, 
	$\widehat{\mathcal{V}}_n(d; \hat{w}) \overset{p}{\longrightarrow} \mathcal{V}_\mathcal{T}(d)$; when $\alpha = 0$, 
	$\widehat{\mathcal{V}}_n(d; \hat{w}) \overset{p}{\longrightarrow} {C_\pi}^{-1} \E[Y(d) \pi(X) \{ 1-\pi(X) \} \mid S=1]$.
\end{corollary}

The limiting weights under both cases have the form of 
$[ {\pi(x)}^{-1} a +  \{ 1-\pi(x) \}^{-1}(1-a) ] h(x).$
They balance the weighted distributions of the covariates between the treated and control groups, and retarget the distribution to a population whose density ratio to the source population is $h(x)$. When $\alpha > 0$, $h(x) \propto \{ 1 - \rho(x) \}/\rho(x)$ is exactly the density ratio between the target population and the source population, so the corresponding weights are the classical importance weights.
For $\alpha = 0$, $h(x) \propto \pi(x) \{ 1 - \pi(x) \}$. With this choice we put more focus on the covariate region where the treated and control groups have good overlap, i.e. the region where $\pi(x)$ is close to 0.5. \citet{li2018balancing} also derived weights of this form from the perspective of minimizing the asymptotic efficient estimation variance of the weighted average treatment effect, and they referred to $w^\dagger$ as the ``overlap weights''. \citet{kallus2020more} advocated using the overlap weights for estimating the optimal ITR with observational data, especially when there is limited overlap between the treated and control groups, because the overlap weights avoid putting too much attention to a few observations and thus reduce estimation variability. In our setting of learning optimal ITR for a target population, the overlap weights are well-justified when $\mathcal{D}$ is well specified and our non-parametric balancing weights converge to the overlap weights purely by means of optimizing within-population imbalance. 

Although the weights converge to the same limit for all $\alpha > 0$, under finite samples, varying $\alpha$ from 1 to 0 provides a smooth transition from the importance weights to the overlap weights. The importance weights aim for minimizing the distributional difference between the source sample and the target population, while the overlap weights put more emphasis on improving stability and reducing any imbalance between treated and control units. Therefore, the proposed method enables us to search for a better trade-off in the middle ground, calibrating estimation towards the target population while retaining small variance and mitigating imbalance due to selection bias into the treatment.

Another advantage of our method compared to the importance weights or overlap weights is that by targeting distributional imbalance directly,  it avoids explicit modeling of the propensity score $\pi(x)$ and participation probability $\rho(x)$. As discussed before, modeling may result in the failure of weighting methods due to model misspecification or estimation error. While the proposed weights are free of model specification issues and have more stable values, Theorem \ref{thm:weight_limit} guarantees that they converge to interpretable limits. Further, since the proposed method is entirely formulated upon covariate balancing, it enjoys better finite sample balance properties, and this generally leads to more precise treatment comparison in practice \citep{chattopadhyay2020balancing}. As far as we know, our work is the first to show that overlap weights can be estimated by constructing weights that purely target imbalance of covariate distributions.


\section{Simulation studies}
\label{sec:simu}

\subsection{Setup}

In this section, we conduct simulation studies to evaluate the performance of the proposed weighting method in finite sample settings. In our simulation setup, we generate covariates $X_i$ independently from a uniform distribution on $[-2, 2]^4$. Given the covariates $X_i$, the population indicator $S_i$ is then generated from a Bernoulli distribution with success probability $\rho(X_i)$, where
$\rho(x) = G(x_2 - 1.2 x_1)$, $G(z) = 0.8 \Phi(z) + 0.1$ and $\Phi(z)$ is the standard normal cumulative probability function. 
Data with $S_i=1$ and $S_i=0$ constitute the source sample and target sample respectively. For the source sample, we further simulate the treatment assignment and outcome as follows.
To test the performance of the proposed method under a wide range of settings, we consider three propensity score models for treatment assignment:
\begin{itemize}[itemsep=0pt, parsep=0pt, topsep=4pt]
    \item[(1)] $\pi(x) = G(0.5 x_1 + 0.3 x_2 - 0.3)$; \hfill (linear assignment)
    \item[(2)] $\pi(x) = G(1.6 x_1 + 1.3 x_2 - 0.8)$; \hfill (linear assignment, bad overlap)
    \item[(3)] $\pi(x) = G(0.4 x_1^2 + 0.4 x_2^2 + 0.5 x_1 x_2 - 0.4 x_1 + 0.4 x_2 - 0.9)$. \hfill (nonlinear assignment)
\end{itemize}
In the first setting the propensity score, after a monotone transformation, linearly depends on the covariates, and thus logistic regression can provide a decent fit even though the link function is not perfectly specified. The second setting is similar, but the large coefficients induce a substantial difference between the covariate distributions of the treated and control groups. The third setting includes quadratic terms and an interaction term, and thus any model that only considers main effects may be insufficient for controlling for confounding.

The observed outcome is generated as $Y_i = m(X_i) + (A_i - 0.5) \tau(X_i) + \varepsilon_i$, where $m(x) = \Phi(-0.6 x_1 - 0.6 x_2 + 0.2 x_3 + 0.5) + 0.5$ and $\varepsilon_i \distas{i.i.d.} N(0, 0.5^2)$.
The  conditional average treatment effect function is set as $\tau(x) = \kappa \tau_\text{NL}(x) + (1 - \kappa) \tau_\text{L}(x)$, where
\begin{align*}
    \tau_\text{NL}(x) &= \Phi(1.5 x_2 + 0.8 x_1 - 0.4 (x_1 - x_2)^2 - 0.3) - 0.07 (x_1 - x_2)^2,\\
    \tau_\text{L}(x) &= \Phi(0.4 x_2 + 0.6 x_1) - 0.5
.\end{align*}
Here, $\tau(x)$ is linear combination of two parts: a nonlinear part $\tau_\text{NL}(x)$ and a linear part $\tau_\text{L}(x)$. The parameter $\kappa$ controls the degree of nonlinearity of $\tau(x)$. We consider two choices of $\kappa$: 0 and 0.4, the former indicating a linear setting and the latter a nonlinear setting. Figure \ref{fig:simu_setting_illus} visualizes $\tau(x)$ under these two choices. 
We choose $\mathcal{D}$ to be the linear rule class, i.e. ITRs of the form $d(x) = \ind(x^\mathsf{T} \beta \ge 0)$. So when $\kappa = 0$, $\mathcal{D}$ contains the unrestricted true optimal ITR, and thus the optimal ITRs for the source population and the target population are the same; otherwise, the optimal linear ITR for the target population is different from that for the source population, which necessitates adjustment for covariate shift from source to target.

We implement the proposed method using the Gaussian kernel with bandwidth chosen via the median heuristic \citep{gretton2012kernel}. The tuning parameters of proposed method ($\alpha$ and $\lambda$) are selected by our proposed resampling algorithm in Section S2.2 with grid search using default grids. To find the linear rule that maximizes $\widehat{\mathcal{V}}_n(d, w)$, we re-cast the optimization problem as a weighted classification problem. Note that $\widehat{\mathcal{V}}_n(d, w)$ can be expressed as
\begin{equation*}
	\frac{1}{n_s} \sum_{i \in \mathcal{S}} \max (w_i Y_i, 0) - \frac{1}{n_s} \sum_{i \in \mathcal{S}} |w_i Y_i| \ind\{d(X_i) \neq A_i \textnormal{sign}(w_i Y_i)\}
,\end{equation*}
so maximizing $\widehat{\mathcal{V}}_n(d, w)$ is equivalent to minimizing $\frac{1}{n_s} \sum_{i \in \mathcal{S}} |w_i Y_i| \ind\{d(X_i) \neq A_i \textnormal{sign}(w_i Y_i)\}$, which is a weighted binary classification problem that uses $A_i \textnormal{sign}(w_i Y_i)$ as the labels and $|w_i Y_i|$ as the weights. We substitute the 0-1 loss with the logistic loss and solve the optimization approximately with weighted logistic regression \citep{xu2015regularized}. The predictive modeling tasks used for value function estimation in the tuning procedure are implemented with \texttt{xgboost} \citep{chen2016xgboost}.

For comparison, we consider other weighting methods, including the overlap weights, the importance weights and inverse probability (of treatment) weighting (labeled as \texttt{Overlap}, \texttt{Importance} and \texttt{IPW}, respectively). To compute these weights, we use logistic regression to fit the propensity score and participation probability. So the propensity score model is nearly correctly specified in the first two treatment assignment scenarios but is misspecified in the last treatment assignment scenario. The oracle counterparts (plotted in gray)that use the underlying true probabilities to construct weights are also included. We further include entropy balancing \citep{hainmueller2012entropy} analogues of the importance weights and inverse probability weighting in the comparison, labeled as \texttt{ebal\_t} and \texttt{ebal\_s} respectively: \texttt{ebal\_t} finds the weights of minimum entropy such that covariate moments of the reweighted treated/control group are equal to those of the target sample, and \texttt{ebal\_s} is defined similarly but with covariate moments given by the whole source sample. 
We note that among the comparison methods, only the importance weights and \texttt{ebal\_t} use the covariate information from the target sample. After obtaining the weights from these methods, we apply the same weighted classification procedure described above to maximize $\widehat{\mathcal{V}}_n(d, w)$ and get the corresponding optimal ITRs, which are then compared to the one estimated with our weights. In the implementation, the outcome modeling tasks are also carried out using \texttt{xgboost}. In the simulation, the total sample size is set to be $n = 1600$. 

{We further evaluated the performance of our proposed weights under more flexible treatment rule classes using weighted decision trees for ITR estimation. In addition, we investigated the setting when ITRs are estimated using weighted regression models instead of outcome weighted learning. Results from these two investigations are included in the Supplementary Material.}

\subsection{Results}
\begin{figure}[!ht]
    \centering
    \includegraphics[width=.95\linewidth]{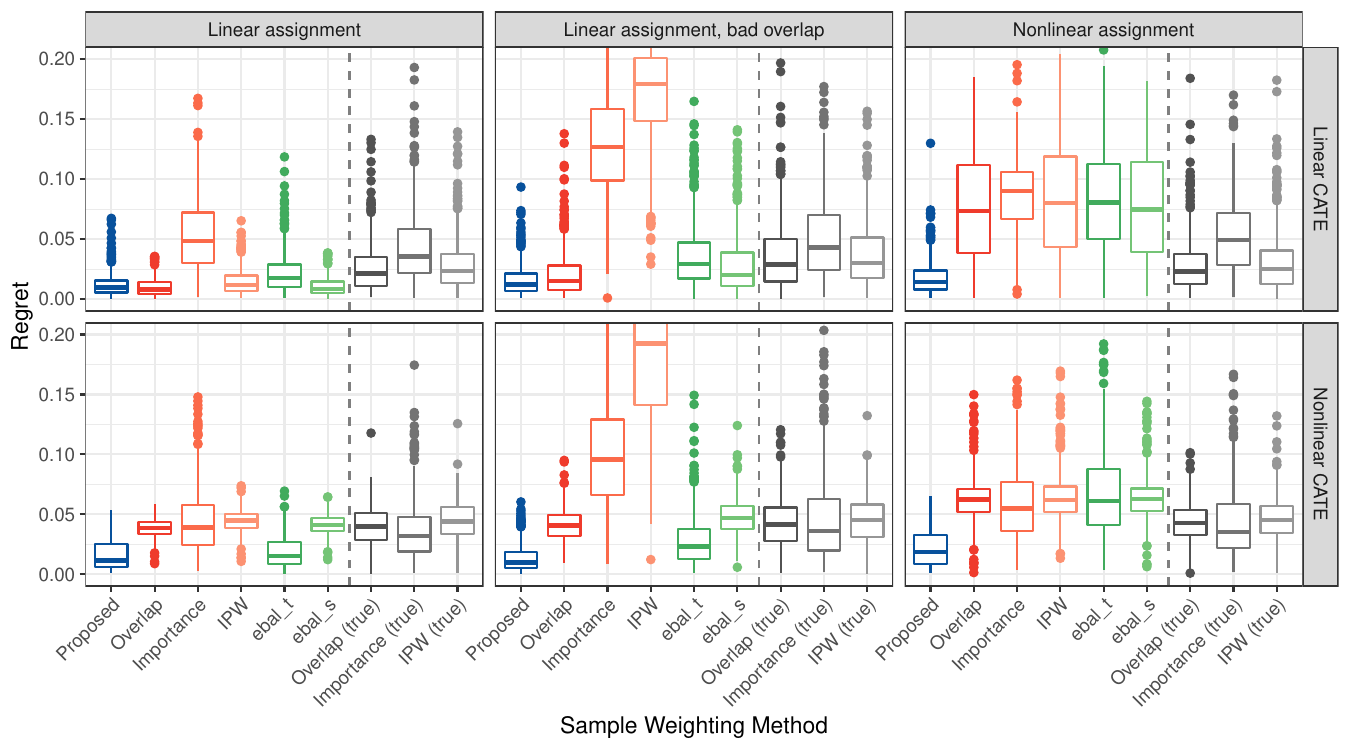}
     \caption{Simulation results. Regret compared to the optimal linear ITR using outcome weighted learning. The term ``CATE'' stands for conditional average treatment effect. Methods labeled with \texttt{(true)} compute the weights using the true values of $\pi(x)$ and $\rho(x)$ instead of the estimated ones. The y-axis is zoomed in for visual clarity.
    }
    \label{fig:simu_results}
\end{figure}

We evaluate the performance of all methods using the regret of the learned ITR in comparison to the optimal linear ITR, specifically, the difference in the value function between the learned ITR and the optimal within-class ITR. An independent test sample of size $10^5$ is generated from the target population to evaluate these quantities. We also use this test sample to search for the optimal linear ITR. Since in our settings only $X_1$ and $X_2$ are effect modifiers, the optimal linear ITR has the form of $d(x) = \ind \{\cos(\theta) x_1 + \sin(\theta) x_2 + b \ge 0 \}$, then we find the optimal $\theta$ and $b$ using the grid search procedure described in \citet{kallus2020more}. The boxplots in Figure \ref{fig:simu_results} summarize the results under different scenarios based on 500 independent simulation runs.

In the first column of Figure \ref{fig:simu_results}, the underlying assignment mechanisms and the participation probabilities are linear in the covariates after a monotonic transformation, and the treated and control groups have sufficient overlap, so these settings constitute the easiest design across our simulation settings. As we can see, even for this simple setting, whether calibrating to the target sample can yield improvement in learning the ITR for the target population does depend on the underlying outcome model. When the rule class contains the unrestricted true optimal ITR, weighting the data to the target population helps only little (if any) in achieving a larger value function. However, it will make the estimated rule more vulnerable to random noise. In this case, the overlap weights attain the best performance and our proposed method achieves a very similar result to the overlap weights, as the tuning parameter selection procedure allows it to yield weights that mimic the overlap weights. On the other hand, when the difference in covariate distributions from source to target leads to a different optimal ITR, it is necessary to tailor the weights to the target population. As a result, the overlap weights yield a suboptimal ITR, while our method adaptively chooses the calibration level $\alpha$ to incorporate covariate shift adjustments and achieves the best performance among methods. We also notice that the importance weights based on modeling $\pi(x)$ and $\rho(x)$ can occasionally achieve great performance, but this is highly unstable with occasional poor results, likely due to large weights.

The second and third columns of Figure \ref{fig:simu_results} compare the performance under more challenging scenarios. When there is a substantial difference between the covariate distributions of the treated and control groups, the performance of the overlap weights is similar to the previous cases in that it yields excellent performance but only for correctly specified rule classes; however, this benefit is attenuated in the nonlinear confounding setting when the model for the propensity score is more severely misspecified. The other modeling-based weights appear to be vulnerable to insufficient treatment-control overlap, the misspecification of the propensity score model, or the wrong choice of moments to balance. The proposed method achieves the best results among all the competing methods, especially when covariate shift needs to be adjusted for. Lastly, when the assignment mechanism is nonlinear, our method is able to retain satisfactory performance due to its non-parametric nature and ability to balance full covariate distributions not just prespecified covariate moments. The relative benefit of the proposed method is the greatest under the nonlinear assignment mechanism settings.

\begin{figure}[ht!] 
	\centering
	\includegraphics[scale = 0.5]{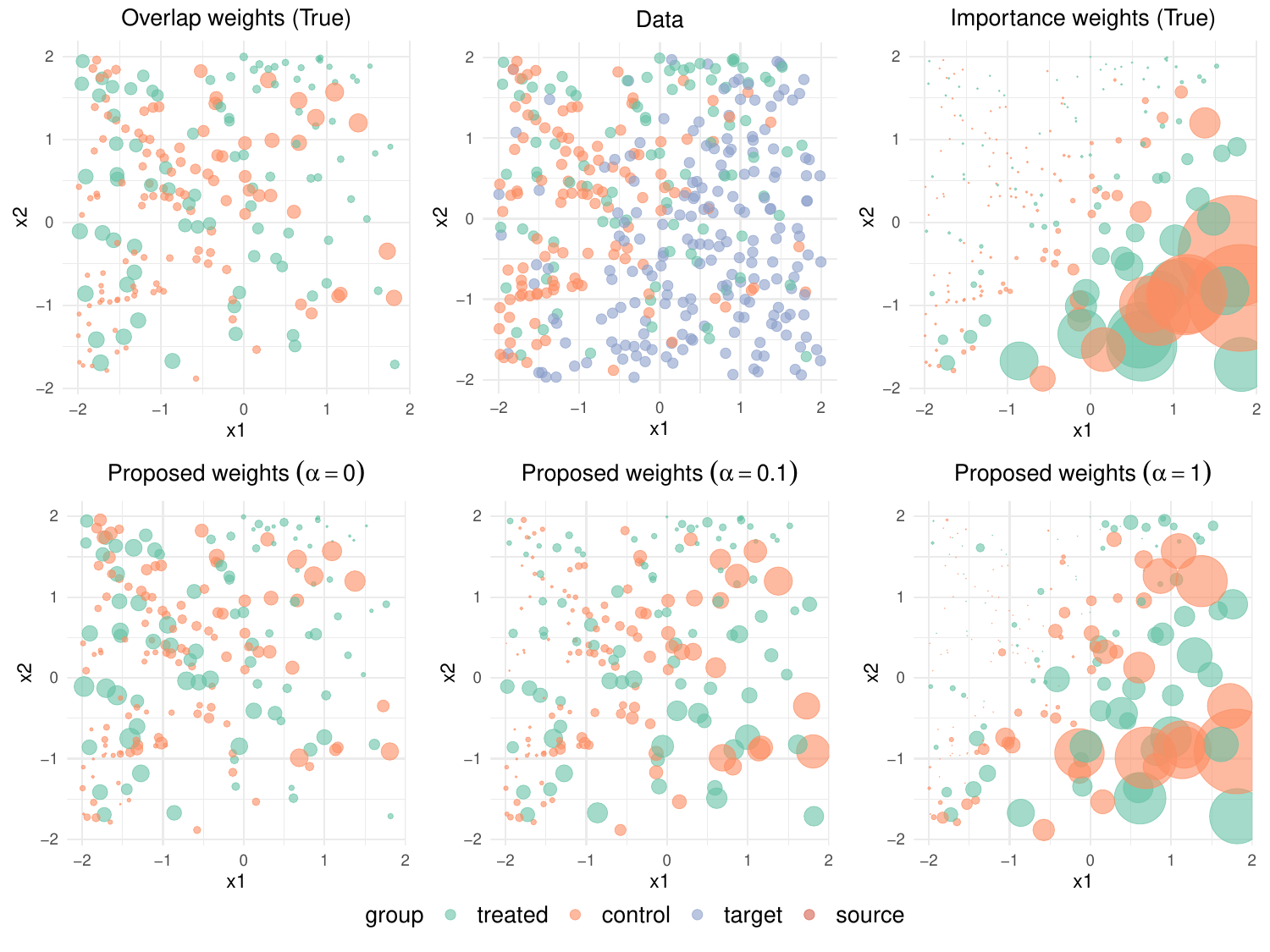}
	 \caption{Covariate distribution and weights for samples from source population under ``Linear assignment, Bad overlap''. Each point represents a sample from either the source (treated and control group) or target population. Except for the plot of the raw data, the size of the point is proportional to the weight assigned to that sample for a given weighting method.}
	\label{fig:weightsplot}
	\vspace{-0.2in}
\end{figure}

We also plot the original covariate distribution of ($X_1,X_2$) of the source and target samples under the ``linear assignment, bad overlap'' scenario as well as the corresponding weighted covariate distribution of source sample with weights generated from our methods ($\alpha =0$,$0.1$ and $1$), overlap weights and importance weights in Figure \ref{fig:weightsplot}. When $\alpha = 0$, our proposed weights are similar to the overlap weights, and when $\alpha = 1$, our proposed weights are similar to the importance weights but less extreme. When $\alpha$ is increased to $0.1$, our proposed weights appear as a smooth combination of overlap weights and importance weights. In our simulation, the chosen $\alpha$'s were all close to 0.1 in this setting.

 {In the Supplementary Material, we also demonstrate that our method can lead to better tree-based ITRs for the target population compared to other weights, indicating that our method still provides substantial benefit even when the specified rule class is flexible. Furthermore, we show that using our weights in weighted regression usually leads to better performance than other weights, including the unweighted regression approach.}


\section{Application: transthoracic echocardiography and mortality in sepsis}
\label{sec:data}

In this section, we illustrate the proposed method by studying individualized treatment recommendation of the use of transthoracic echocardiography (TTEC) for intensive care unit (ICU) patients with sepsis. \citet{feng2018transthoracic} studied the average treatment effect of TTEC on improving 28-day mortality and suggested the use of TTEC is associated with a significant improvement. However, It has been argued that TTEC use in surgical ICU  is not cost-effective due to a high failure rate \citep{cook2002transthoracic}. Hence, studying potential heterogeneous treatment effects of TTEC and identifying a subgroup of patients who can benefit from it is clinically relevant.

We use the same dataset in \citet{feng2018transthoracic}, which is derived from the MIMIC-III database \citep{johnson2016mimic}. This is an observational dataset consisting of 6361 ICU patients. Among the patients, 51.3\% had TTEC performed during or in the period less than 24 hours before their ICU admission. We use 28-day survival as the outcome. We use 17 baseline variables in the analysis: age (range 18 - 91), gender, weight; severity at admission, which is measured by the simplified acute physiology score (SAPS), the sequential organ failure assessment (SOFA) score, and the Elixhauser comorbidity score; comorbidity indicators, including congestive heart failure, atrial fibrillation, respiratory failure, malignant tumor; vital signs, including mean arterial pressure, heart rate and temperature; laboratory results, including platelet count, partial pressure of oxygen, lactate, and blood urea nitrogen.
The distributions of lab results are right skewed, so we apply log transformations on these variables. All the continuous variables are then standardized for further analysis. We impute any missing values using MissForest \citep{stekhoven2012missforest}, which is a flexible non-parametric missing value imputation approach.

In order to evaluate ITR learning results and study the influence of covariate shift and confounding, we use the following procedure to construct the training data and test data which induces different levels of covariate shift and confounding. First, we randomly sample 40\% of the data without replacement; this part will be used to construct the source sample and the remaining represents the target population. The sampling probability is proportional to a function of the covariates, $f(X_i)$. We consider two choices of $f$: (a) $f(x) = 0.5$, which means each individual is sampled with equal probability and (b) $f(x) = G(\text{age} - \text{weight} + 0.6 \times I(\text{gender} = \text{female}) - 0.4)$, so the source and target populations have different covariate distributions. Here $G(\cdot)$ is the function defined in Section \ref{sec:simu}. Then among the first part of the data, we randomly select (without replacement) 50\% of the TTEC patients and 50\% of the non-TTEC patients to form the source sample; however, the treated units are selected with probability proportional to $g(X_i)$ while the control units are with probability proportional to $0.5 - g(X_i)$. This step induces additional confounding to the source data based on $g$. Three choices of $g$ are considered: (a) $g(x) = 0$, so there is no extra confounding; (b) $g(x) = G(\text{SAPS} + 0.8 \times \text{SOFA} - 0.9\times \text{Elixhauser})$, which introduces extra confounding determined by a linear combination of the severity scores; (c) $g(x) = G(0.3 \times \text{SAPS}^2 + 0.2 \times \text{SOFA}^2 + 0.2 \times \text{Elixhauser}^2 + 0.4 \times \text{SAPS} \times \text{SOFA} - 0.3 \times \text{SAPS} + 0.4 \times \text{SOFA} + 0.2 \times \text{Elixhauser} - 0.6)$, which leads to a confounding mechanism that involves an interaction term. Among the second part of the data, 1/3 of the patients are randomly sampled with equal probability, and their covariate information is used as the target sample, and the remaining data is used as the test data. So the target sample and the test data have the same covariate distribution.

\begin{figure}[!ht]
    \centering
    \includegraphics[width=.95\linewidth]{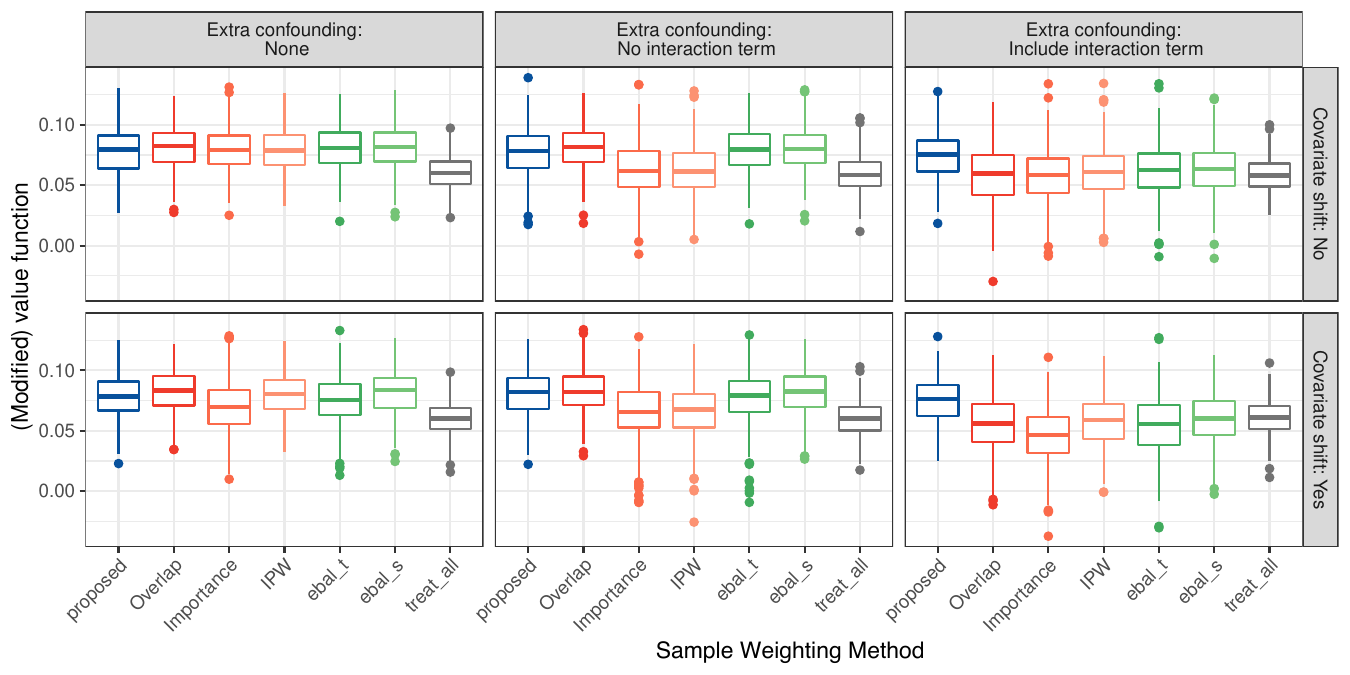}
    \caption{Modified version of value function in the real data application. \texttt{treat\_all} is a trivial ITR that assigns all patients to treatment and the result equals the average treatment effect. The other methods are the same as those in Figure \ref{fig:simu_results}.}
    \label{fig:echo_value}
\end{figure}

By the aforementioned procedure, we preserve the covariate-outcome relationship in the real data, but use various sampling designs to induce different levels of covariate shift and confounding. There are 6 scenarios; under each one, we run 400 replications of the sampling procedure. For each replication, we then apply the proposed method to estimate the optimal ITR for the target population. The competing methods in Section \ref{sec:simu} are also applied, except the ones using the true probabilities. In addition, we include a trivial treatment rule that assigns all patients to TTEC (labeled as \texttt{treat\_all}) to help calibrate the performances of the methods relative to the average effect of TTEC. The evaluation metric is a modified version of the value function: $\mathcal{V}_{\mathcal{T},2}(d) = \E[Y(d(X)) - Y(1 - d(X)) \mid S = 0]$. For comparison purpose, $\mathcal{V}_{\mathcal{T},2}(d)$ is equivalent to $\mathcal{V}_\mathcal{T}(d)$ as $\mathcal{V}_{\mathcal{T},2}(d) = 2 \mathcal{V}_\mathcal{T}(d) - \E[Y(1)+Y(0) \mid S=0]$. We use $\mathcal{V}_{\mathcal{T},2}(d)$ because it is equal to the average treatment effect when $d$ is the trivial treat-all ITR. We estimate $\mathcal{V}_{\mathcal{T},2}(d)$ on the test data using the conventional inverse probability estimator. The result is reported in Figure \ref{fig:echo_value}.

The modified value function of the trivial treat-all ITR is around 6\% (i.e. a 6\% improvement in 28-day survival rate over no TTEC), which is close to the average treatment effect estimate in \citet{feng2018transthoracic} (5\%). The value achieved by the best performing ITRs is around 8\%. This suggests the presence of heterogeneity of treatment effect,  thus personalizing decisions of whether to use TTEC  can potentially lead to an improved overall survival rate. The performance of all the methods are close when there is no covariate shift and extra confounding. 

When source population and target populations have different covariate distributions (second row of Figure \ref{fig:echo_value}), methods like \texttt{ebal\_t} and \texttt{Importance} that calibrate to the target sample perform subpar. This suggests that a linear rule is likely sufficient and that calibrating the covariates to a specific distribution may be unnecessary; in this case, the overlap weights is a desirable option. The advantage of overlap weights becomes more noticeable when the treated and control groups have less overlap (second column of Figure \ref{fig:echo_value}). In this case the proposed method also achieves good performance because it can adaptively approximate the overlap weights by selecting a small $\alpha$ value. From the third column of Figure \ref{fig:echo_value} we can see that the entropy balancing  and probability modeling methods are subject to the choice of balancing functions or model specification. In contrast, the proposed method is free of such specification issues. Further, we can see that even with covariate shift and extra confounding with an interaction, our proposed method does nearly as well as when there is no covariate shift and no extra confounding.


\section{Discussion}
\label{sec:discussion}

In this work, we have proposed a weighting framework for ITR learning in the two-population setting where the target population's covariate distribution might differ from the source. A key benefit of our method is that it directly targets covariate balance and does not require probability modeling. To assess covariate balance after weighting, several descriptive statistics can be used, including (standardized) mean differences \citep{rosenbaum1985constructing}, variance ratios \citep{austin2009balance}, and weighted Kolmogorov-Smirnov statistics \citep{austin2015moving}, all of which are implemented in the \texttt{R} package \texttt{cobalt} (see the Supplementary Material for an example). 

Our methodological development focused on the objective function \eqref{eq:weighted_estimator}, which is based on the outcome weighted learning approach \citep{zhao2012estimating}. The proposed weights, however, can be used in any weighting-based ITR learning approaches, like those in \citet{wallace2015doubly}, \citet{zhou2017residual}, and \citet{chen2017general}. By leveraging the target population's subject-level covariate information, the proposed weights improve the generalization capabilities of these ITR learning methods. 
Our framework can also be applied to learn ITRs in the single-population setting by replacing $\prob_{n,\mathcal{T}}$ in \eqref{eq:opt_prog_threeway} with $\prob_{n,\mathcal{S}}$, the empirical covariate distribution of the source sample. Similar to the two-population setting, while treated and control groups should be weighted to have similar distributions, this weighted distribution does not necessarily need to resemble the source population, especially when the ITR class is correctly specified \citep{kallus2020more}. 
Lastly, our framework can be extended to include other statistical distances by substituting $\textnormal{MMD}_\mathcal{H}$ in \eqref{eq:opt_prog_threeway} with alternatives like the energy distance \citep{szekely2013energy} or the Wasserstein distance \citep{villani2008optimal}. The resulting weights can also balance the entire covariate distributions. However, the theoretical properties of the weights under these alternative distances would require further investigation.

\section*{Acknowledgment}
We are grateful to the referees, the associate editor and the editor for their valuable comments and suggestions. Research reported in this work was partially funded through a Patient-Centered Outcomes Research Institute (PCORI) Award (ME-2018C2-13180). The views in this work are solely the responsibility of the authors and do not necessarily represent the views of the PCORI, its Board of Governors or Methodology Committee.

\section*{Supplementary material}
The Supplementary Material includes proofs for the theoretical results in Section \ref{sec:method}, computational details for the proposed method, and additional simulation and data application results.

\bibliographystyle{apalike}
\bibliography{ref_generalizability}

\newpage 

\appendix 

\renewcommand{\thesection}{S\arabic{section}}
\renewcommand{\thefigure}{S\arabic{figure}}
\renewcommand{\thetable}{S\arabic{table}}
\renewcommand{\theequation}{S\arabic{equation}}
\setcounter{figure}{0}

{\large Supplementary Material for ``Robust Sample Weighting to Facilitate Individualized Treatment Rule Learning for a Target Population''}

\section{Technical proofs}

We impose the following regularity conditions throughout the proof.
\begin{enumerate}[label = (\alph*), itemsep=0pt]
	\item $\mathcal{X}$ is a compact space. The covariates have continuous density on $\mathcal{X}$. The propensity score $\pi(x)$ and participation probability $\rho(x)$ are both continuous functions on $\mathcal{X}$. This implies that covariate distributions of the source and target populations are also continuous and they are denoted by $p_s(x)$ and $p_t(x)$ respectively.
	\item $K(x, x')$ is continuous, non-negative, and bounded.
	\item $\mathcal{H}$ is a universal RKHS. In other words, $\mathcal{H}$ is dense in $C(\mathcal{X})$ with respect to the supremum norm. 
	\item The unit ball in $\mathcal{H}$, denoted by $\mathcal{H}_1$, is $\prob$-Donsker.

\end{enumerate}

\subsection{Proof of \eqref{eq:bound_mmd_d_mu}}
\label{sec:pf_bound_mmd_d_mu}
For the inequality \eqref{eq:bound_mmd_d_mu} we assume $K(x, x')$ takes the form of $\theta^{-p} K_0(\|x-x'\| / \theta)$ and define $\gamma = \sup\{\gamma: \exists L>0, |\{w(1, x)p_1(x) - p_t(x)\} - \{w(1, x')p_s(x') - p_t(x')\}| \le L\|x-x'\|^\gamma\} > 0$. For instance, when $w(1, x)p_s$ and $p_t(x)$ are Lipschitz continuous, then this condition holds with $\gamma = 1$. We additionally assume $K_0$ satisfies $\int K_0(\|x\|) \|x\|^\gamma \textnormal{d} x < \infty$. We first show the following lemma:
\begin{lemma}
	\label{lem:mmd_compute}
	Suppose $g:\mathcal{X} \rightarrow \mathbb{R}$ is a bounded function and $\nu$ is a measure on $\mathcal{X}$. $\mathcal{H}$ is an RHKS on $\mathcal{X}$ associated with a bounded kernel function $K(x, x')$. Then,
	\begin{equation*}
		\sup_{h \in \mathcal{H}_1}
		\int h(x) g(x) \textnormal{d} \nu(x)
		=
		\left( \iint K(x, x') g(x) g(x') \textnormal{d} \nu(x) \textnormal{d} \nu(x') \right)^{\frac{1}{2}}
		.\end{equation*}
\end{lemma}
\begin{proof}
	Define a linear operator $M_g$ on $\mathcal{H}$ by 
	$M_g(h) = \int h(x) g(x) \textnormal{d} \nu(x)$ for $h \in \mathcal{H}$. Since
	\begin{equation*}
		|M_g(h)| 
		\le
		\int |\langle h, K(x, \cdot) \rangle_\mathcal{H}| |g(x)| \textnormal{d} \nu(x)
		\le
		\| h \|_\mathcal{H} \int \sqrt{K(x, x)} |g(x)| \textnormal{d} \nu(x)
		,\end{equation*}
	$M_g$ is continuous. Hence by the Riesz representer theorem, there exists a $h_g \in \mathcal{H}$ such that $M_g(h) = \langle h_g, h \rangle_\mathcal{H}$. Setting $h = K(x', \cdot)$ we can obtain
	$h_g(x') = M_g( K(x', \cdot) ) = \int K(x, x') g(x) \textnormal{d} \nu(x)$.
	So we have
	\begin{equation*}
		\langle h_g, h_g \rangle_\mathcal{H} =
		M_g(h_g) =
		\int h_g(x') g(x') \textnormal{d} \nu(x') =
		\iint K(x, x') g(x) g(x') \textnormal{d} \nu(x) \textnormal{d} \nu(x')
		.\end{equation*}
	Then the desired result follows from
	$
	\sup_{h \in \mathcal{H}_1} M_g(h) = 
	\sup_{h \in \mathcal{H}_1} \langle h, h_g \rangle_\mathcal{H} = 
	\| h_g \|_\mathcal{H}
	$.
\end{proof}
\noindent\textbf{Remark.}
This Lemma also implies Lemma \ref{lem:mmd_finite} by letting $\nu = \sum_{i \in \mathcal{I}_P \cup \mathcal{I}_Q} \delta_{x_i}$ and $g(x) = \sum_{i \in \mathcal{I}_P}p_i \ind(x = x_i) - \sum_{j \in \mathcal{I}_Q}q_j \ind(x = x_j)$. 

\begin{proof}[Proof of \eqref{eq:bound_mmd_d_mu}]
	We show the following inequality
	\begin{equation}
		\label{eq:concentration}
		\begin{aligned}
			\prob \Bigg(
			&
			\sup_{\mu_1 \in \mathcal{H}_1}
			\left\{ \int_\mathcal{X} \mu_1(x) \textnormal{d} \big[\prob_{n,\mathcal{S}_1}^{(w)} - \prob_{n,\mathcal{T}}\big](x) \right\} + 
			C_1 \left( \frac{1}{n \theta^p} + \theta^{\gamma/2} + t \right)
			\\&~<
			\sup_{\mu_1 \in \mathcal{H}_1}
			\left\{ \int_\mathcal{X} d(x) \mu_1(x) \textnormal{d} \big[\prob_{n,\mathcal{S}_1}^{(w)} - \prob_{n,\mathcal{T}}\big](x) \right\}
			\Bigg) \le \exp\{-C_2 t^2 n \theta^p\}
			,\end{aligned}
	\end{equation}
	where $C_1, C_2$ are constants not related to $d(x)$ and $t$ is any positive number.
	In the following proofs, we use $C$ to denote constants, but it may represent different values under different contexts. Without loss of generality, we assume $w(1, x)$ has been normalized, i.e., $\int_\mathcal{X} w(1, x) p_s(x) \textnormal{d} x = 1$.
	
	The proof of \eqref{eq:concentration} consists of two parts. First, we show the reweighted sample $\textnormal{MMD}_\mathcal{H}$ can be bounded by the reweighted population $\textnormal{MMD}_\mathcal{H}$ with high probability. Suppose $K_0(x, x')$ is upper bounded by a constant $\bar{K}_0$, then $K(x, x')$ is upper bounded by $\bar{K}_0 / \theta^p$. Following the same arguments in \citet{gretton2012kernel} Theorem 7, we can show that for any $t > 0$, with probability at least $1 - \exp\{-C t^2 n \theta^p\}$,
	
	{\small
		\begin{equation}
			\label{eq:mmd_du_u_sample_pop}
			\begin{aligned}
				\left| 
				\sup_{\mu_1 \in \mathcal{H}_1}
				\left\{ \int_\mathcal{X} d(x) \mu_1(x) \textnormal{d} \big[\prob_{n,\mathcal{S}_1}^{(w)} - \prob_{n,\mathcal{T}}\big](x) \right\}
				-
				\sup_{\mu_1 \in \mathcal{H}_1}
				\left\{ \int_\mathcal{X} d(x) \mu_1(x) \textnormal{d} \big[\prob_{\mathcal{S}_1}^{(w)} - \prob_{\mathcal{T}}\big](x) \right\}
				\right|
				&\le
				C\left( \frac{1}{{n \theta^p}} + t \right),
				\\
				\left| 
				\sup_{\mu_1 \in \mathcal{H}_1}
				\left\{ \int_\mathcal{X} \mu_1(x) \textnormal{d} \big[\prob_{n,\mathcal{S}_1}^{(w)} - \prob_{n,\mathcal{T}}\big](x) \right\}
				-
				\sup_{\mu_1 \in \mathcal{H}_1}
				\left\{ \int_\mathcal{X} \mu_1(x) \textnormal{d} \big[\prob_{\mathcal{S}_1}^{(w)} - \prob_{\mathcal{T}}\big](x) \right\}
				\right|
				&\le
				C\left( \frac{1}{{n \theta^p}} + t \right)
				.\end{aligned}
		\end{equation}
	}
	
	Second, we show 
	{\small
		\begin{equation}
			\label{eq:dmu_le_mu_population}
			\sup_{\mu_1 \in \mathcal{H}_1}
			\left\{ \int_\mathcal{X} d(x) \mu_1(x) \textnormal{d} \big[\prob_{\mathcal{S}_1}^{(w)} - \prob_{\mathcal{T}}\big](x) \right\}
			\le
			\sup_{\mu_1 \in \mathcal{H}_1}
			\left\{ \int_\mathcal{X} \mu_1(x) \textnormal{d} \big[\prob_{\mathcal{S}_1}^{(w)} - \prob_{\mathcal{T}}\big](x) \right\} + C \theta^{\gamma/2}
			.\end{equation}
	}%
	Note that both $\sup$ terms are non-negative due to symmetry of $\mathcal{H}_1$, so it suffices to show
	\begin{equation*}
		\sup_{\mu_1 \in \mathcal{H}_1}
		\left\{ \int_\mathcal{X} d(x) \mu_1(x) \textnormal{d} \big[\prob_{\mathcal{S}_1}^{(w)} - \prob_{\mathcal{T}}\big](x) \right\}^2
		\le
		\sup_{\mu_1 \in \mathcal{H}_1}
		\left\{ \int_\mathcal{X} \mu_1(x) \textnormal{d} \big[\prob_{\mathcal{S}_1}^{(w)} - \prob_{\mathcal{T}}\big](x) \right\}^2 + C^2 \theta^{\gamma}
		,\end{equation*}
	or equivalently,
	\begin{equation*}
		\sup_{\mu_1 \in \mathcal{H}_1}
		\left( \int_\mathcal{X} \mu_1(x) d(x) g(x) \textnormal{d} x \right)^2
		\le
		\sup_{\mu_1 \in \mathcal{H}_1}
		\left( \int_\mathcal{X} \mu_1(x) g(x) \textnormal{d} x \right)^2 + C^2 \theta^{\gamma}
		,\end{equation*}
	where let $g(x) = w(1,x)p_1(x) - p_t(x)$.
	Applying Lemma \ref{lem:mmd_compute} on the supremum terms we can see this is equivalent to showing
	\begin{equation}
		\label{eq:Kddgg_le_Kgg}
		\iint K(x, x') d(x) d(x') g(x) g(x') \textnormal{d} x \textnormal{d} x'
		\le
		\iint K(x, x') g(x) g(x') \textnormal{d} x \textnormal{d} x' + C^2 \theta^{\gamma}
		.\end{equation}
	Let $g_+(x) = \max(g(x), 0)$ and $g_-(x) = \max(-g(x), 0)$, then $g(x) = g_+(x) - g_-(x)$, $g_+(x) \ge 0$, $g_-(x) \ge 0$ and $g_+(x) g_-(x) = 0$. Then the left-hand side of \eqref{eq:Kddgg_le_Kgg} equals
	\begin{align*}
		&\iint K(x, x') d(x) d(x') g_+(x) g_+(x') \textnormal{d} x \textnormal{d} x' +
		\iint K(x, x') d(x) d(x') g_-(x) g_-(x') \textnormal{d} x \textnormal{d} x' \\
		&\qquad - 2 \iint K(x, x') d(x) d(x') g_+(x) g_-(x') \textnormal{d} x \textnormal{d} x'
		\\ \le &
		\iint K(x, x') g_+(x) g_+(x') \textnormal{d} x \textnormal{d} x' +
		\iint K(x, x') g_-(x) g_-(x') \textnormal{d} x \textnormal{d} x'
		\\ = &
		\iint K(x, x') g(x) g(x') \textnormal{d} x \textnormal{d} x' + 
		2 \iint K(x, x') g_+(x) g_-(x') \textnormal{d} x \textnormal{d} x'
		.\end{align*}
	The first inequality is because $0 \le d(x) \le 1$ and $g_+, g_-, K$ are all non-negative. Note that when $g_+(x) > 0$ and $g_-(x') > 0$, $g_+(x') = 0$, $g_+(x) \le g_+(x') + L \|x-x'\|^\gamma = L \|x-x'\|^\gamma$. So the second term of the equation above can be controlled by
	\begin{align*}
		\int_\mathcal{X} \int_\mathcal{X} K(x, x') g_+(x) g_-(x') \textnormal{d} x \textnormal{d} x'
		&\le
		\frac{L}{\theta^p} \int_\mathcal{X} g_-(x') \textnormal{d} x'
		\int_\mathcal{X} K_0(\frac{\|x-x'\|}{\theta}) \|x-x'\|^{\gamma} \textnormal{d} x
		\\&=
		L \theta^\gamma \int_\mathcal{X} g_-(x')\textnormal{d} x' \int_{\mathcal{X}/\theta - x'/\theta} K_0(\|z\|) \|z\|^\gamma \textnormal{d} z
		\\&=
		C^2 \theta^{\gamma}
		.\end{align*}
	Thus, we've proved \eqref{eq:Kddgg_le_Kgg}. Combining \eqref{eq:dmu_le_mu_population} and \eqref{eq:Kddgg_le_Kgg} yields \eqref{eq:concentration}. The desired result immediately follows by setting $t = h^{\gamma/2}$.
\end{proof}

\subsection{Proof of Theorem \ref{thm:weight_limit} and Corollary \ref{thm:vndw_limit}}

For notational simplicity, we will introduce new notation in the proof. The most frequently used notations are summarized here for easier reference ($\xi$ represents a generic random variable or a function of the observed data):
\begin{itemize}[itemsep=0pt]
	\item $w^{(1)} = \{w_i: i \in \mathcal{S}_1\}$, $w^{(0)} = \{w_i: i \in \mathcal{S}_0\}$ and $w = (w^{(1)\mathsf{T}}, w^{(0)\mathsf{T}})^\mathsf{T}$. 
	\item $\mathcal{H}_1 = \left\{ h\in \mathcal{H}: \| h \|_\mathcal{H} \le 1 \right\}$. $\mathcal{F} = \mathcal{H} \oplus \mathcal{H}$, $\mathcal{F}_1 = \mathcal{H}_1 \oplus \mathcal{H}_1$. 
	Elements in $\mathcal{F}$ are denoted as $f = (f^{(1)}, f^{(0)})$.
	\item $n_s = |\mathcal{S}|, ~ n_t = |\mathcal{T}|, ~ n_1 = |\mathcal{S}_1|, ~ n_0 = \mathcal{S}_0$.
	\item $\bar{\xi} = \prob_{n, \mathcal{S}}(\xi) = \frac{1}{n_s} \sum_{i \in \mathcal{S}} \xi_i$, 
	$\bar{\xi}^{(1)} = \prob_{n, \mathcal{S}_1}(\xi) = \frac{1}{n_1} \sum_{i \in \mathcal{S}_1} \xi_i$ and
	$\bar{\xi}^{(0)} = \prob_{n, \mathcal{S}_0}(\xi) = \frac{1}{n_0} \sum_{i \in \mathcal{S}_0} \xi_i$ denote the sample mean on the source sample, treated group and control group, respectively.
	\item For $f = (f^{(1)} , f^{(0)}) \in \mathcal{F}$, we let $\bar{f}^{(1)} = \frac{1}{n_1} \sum_{i \in \mathcal{S}_1} f^{(1)}(X_i)$ and $\bar{f}^{(0)} = \frac{1}{n_0} \sum_{i \in \mathcal{S}_0} f^{(0)}(X_i)$.
	\item $\E_\mathcal{S}(\xi) = \E(\xi \mid S=1)$.
	$\E_\mathcal{T}(\xi) = \E(\xi \mid S=0)$.
	$\E_{\mathcal{S}_1}(\xi) = \E(\xi \mid S=1, A=1)$.
	$\E_{\mathcal{S}_0}(\xi) = \E(\xi \mid S=1, A=0)$.
	$\var_{\mathcal{S}_1}(\xi) = \var(\xi \mid S=1, A=1)$.
	$\var_{\mathcal{S}_0}(\xi) = \var(\xi \mid S=1, A=0)$.
	\item $\pi = \E(\pi(X))$
\end{itemize}

\begin{proof}
	
	The analysis is broken down the following steps:
	\begin{itemize}[leftmargin=5em, itemsep=0pt, parsep=0pt, topsep=.6em]
		\item[Step 1.] Characterize the dual problem of \eqref{eq:opt_prog_threeway}.
		\item[Step 2.] Solve the limiting problem of the dual problem.
		\item[Step 3.] Show the weights converge to the limiting solution.
	\end{itemize}
	In order to keep the main ideas clear, detailed proofs of some intermediate results are deferred as lemmas after the main proof.
	
	\paragraph{Step 1.} 
	Let $\mathcal{F} = \mathcal{H} \oplus \mathcal{H}$ be the direct sum of two identical RKHS's $\mathcal{H}$; that is, $\mathcal{F}$ is a Hilbert space consisting of elements $\{f = (f^{(1)}, f^{(0)}): f^{(1)}, f^{(0)} \in \mathcal{H}\}$ and its inner product is defined by 
	$\langle f_1, f_2 \rangle_{\mathcal{F}} 
	= \langle f^{(1)}_1, f^{(1)}_2 \rangle_\mathcal{H} 
	+ \langle f^{(0)}_1, f^{(0)}_2 \rangle_\mathcal{H}$.

	Define a linear map $M: \mathbb{R}^{n_s} \rightarrow \mathcal{F}$ by
	\begin{equation*}
		M w = 
		\begin{bmatrix}
			\sum_{i \in \mathcal{S}_1} w_i K(X_i, \cdot) \\
			\sum_{i \in \mathcal{S}_0} w_i K(X_i, \cdot)
		\end{bmatrix}
		.\end{equation*}
	We also define $h_\mathcal{T} = \frac{1}{n_t} \sum_{i \in \mathcal{T}} K(X_i, \cdot) \in \mathcal{H}$ and 
	\begin{equation*}
		\Lambda(x) = \begin{cases}
			0, & \text{if } \sum_{i} x_i = n_s, \\
			+\infty, & \text{otherwise.}
		\end{cases}
	\end{equation*}
	Here we loosely define $\Lambda$ on a Euclidean space without specifying the dimension and use $x$ to denote a vector in that Euclidean space. When used below, the dimension could be $n_1 = |\mathcal{S}_1|$ or $n_0 = |\mathcal{S}_0|$.
	
	Then the optimization problem \eqref{eq:opt_prog_threeway} is equivalent to minimizing the following (extended real-valued) primal $p: \mathbb{R}^{n_s} \rightarrow \mathbb{R} \cup \{+\infty\}$ defined below:
	\begin{equation}
		\label{eq:primal_def}
		\begin{aligned}
			p(w) &= s(w) + r(M w), \quad \text{where} \\
			s(w) &= \Lambda(w^{(1)}) + \Lambda(w^{(0)}) + \frac{\lambda}{2} \sum_{i \in \mathcal{S}} w_i^2 \quad \text{and} \\
			r(f) &= \frac{\alpha}{2} \| f^{(1)} - n_s h_\mathcal{T} \|_\mathcal{H}^2 + 
			\frac{\alpha}{2} \| f^{(0)} - n_s h_\mathcal{T} \|_\mathcal{H}^2 +
			\frac{1-\alpha}{2} \| f^{(1)} - f^{(0)} \|_\mathcal{H}^2
			.\end{aligned}
	\end{equation}
	$s$ and $r$ are both convex and lower-semicontinuous functions and so is $p$. Next we characterize the Fenchel–Rockafellar dual (\citet{peypouquet2015convex}, Section 3.6) of $p$. Since the dual space of $\mathcal{F}$ is itself, the dual, denoted by $d$, is a function on $\mathcal{F}$ and has the following form:
	\begin{equation}
		\label{eq:dual_def}
		d(f) = s^*(- M^* f) + r^*(f)
		.\end{equation}
	Here $M^*: \mathcal{F} \rightarrow \mathbb{R}^{n_s}$ is the adjoint operator of $M$, and $s^*: \mathbb{R}^{n_s} \rightarrow \mathbb{R}$ and $r^*: \mathcal{F} \rightarrow \mathbb{R}$ are the Fenchel conjugates of $r$ and $s$ respectively. Specifically, $M^*$ satisfies $\langle M^* f, e_i \rangle = \langle f, M e_i \rangle_{\mathcal{F}}$
	for the standard basis vectors $e_1, ..., e_{n_s}$. By the definition of $M$,
	\begin{equation}
		\label{eq:M_star_f}
		\langle M^* f, e_i \rangle = 
		\begin{cases}
			f^{(1)}(X_i), & \text{if } i \in \mathcal{S}_1, \\
			f^{(0)}(X_i), & \text{if } i \in \mathcal{S}_0.
		\end{cases}
	\end{equation}
	We defer the detailed calculation of $s^*$ and $r^*$ to Lemma \ref{lem:dual_of_s_r} below. Using these results we obtain
	\begin{equation}
		\label{eq:dual}
		\begin{aligned}
			d(f) 
			&= 
			\frac{1}{2\lambda} \sum_{i \in \mathcal{S}_1} (f^{(1)}(X_i) - \bar{f}^{(1)})^2 +
			\frac{1}{2\lambda} \sum_{i \in \mathcal{S}_0} (f^{(0)}(X_i) - \bar{f}^{(0)})^2 
			\\&\quad-
			n_s (\bar{f}^{(1)} + \bar{f}^{(0)}) +
			n_s \prob_{n, \mathcal{T}}(f^{(1)} + f^{(0)})
			\\&\quad+
			\frac{1}{4(2-\alpha)} \| f^{(1)} - f^{(0)} \|_\mathcal{H}^2 +
			\frac{1}{4\alpha} \| f^{(1)} + f^{(0)} \|_\mathcal{H}^2
			- \frac{\lambda n_s^2 (n_1 + n_0)}{2 n_1 n_0}
			,\end{aligned}
	\end{equation}
	where $\bar{f}^{(a)} = \frac{1}{n_a} \sum_{i \in \mathcal{S}_a} f^{(a)}(X_i)$, $a \in \{0, 1\}$. When $\alpha = 0$, $d(f) = 0$ if $f^{(1)} + f^{(0)} = 0$ and $+\infty$ otherwise.
	
	Since $p$ has a minimizer $\hat{w}$ and $r$ is continuous at $M \hat{w}$, by the Fenchel–Rockafellar duality (\citet{peypouquet2015convex}, Theorem 3.51), there exists $\hat{f} \in \mathcal{F}$ such that $d(\hat{f}) = \min_{f \in \mathcal{F}} d(f) = -p(\hat{w})$, and $-M^* \hat{f} \in \partial s(\hat{w})$. Here $\partial s(w)$ denotes the set of all subgradients of $s$ at $w$; that is $\partial s(w) = \{ v \in \mathbb{R}^{n_s}: s(w') - s(w) \ge \langle v, w' - w \rangle ,~\forall w' \}$. It is easy to verify that
	\begin{equation*}
		\partial s(w) = \left\{ \begin{bmatrix} 
			\lambda w^{(1)} - \gamma_1 \mathbf{1} \\ 
			\lambda w^{(0)} - \gamma_0 \mathbf{1} 
		\end{bmatrix}: \gamma_1, \gamma_0 \in \mathbb{R}  \right\}
		.\end{equation*}
	Since $-M^* \hat{f} \in \partial s(\hat{w})$, by \eqref{eq:M_star_f} we have
	\begin{equation*}
		\begin{cases}
			-\hat{f}^{(1)}(X_i) = \lambda \hat{w}_i - \hat{\gamma}_1, &\text{if } i \in \mathcal{S}_1, \\
			-\hat{f}^{(0)}(X_i) = \lambda \hat{w}_i - \hat{\gamma}_0, &\text{if } i \in \mathcal{S}_0,
		\end{cases}
	\end{equation*}
	for some $\hat{\gamma}_1, \hat{\gamma}_0 \in \mathbb{R}$. Recall $\sum_{i \in \mathcal{S}_1} \hat{w}_i = \sum_{i \in \mathcal{S}_0} \hat{w}_i = n_s$, this implies $\hat{\gamma}_a = \bar{\hat{f}}^{(a)} + \frac{\lambda n_s}{n_a}$ for $a = 1, 0$. In other words,
	\begin{equation}
		\label{eq:w_f}
		\hat{w}_i = 
		\begin{cases}
			\dfrac{-f^{(1)}(X_i) + \bar{\hat{f}}^{(1)}}{\lambda} + \dfrac{n_s}{n_1}, &\text{if } i \in \mathcal{S}_1, 
			\vspace{4pt} \\
			\dfrac{-f^{(0)}(X_i) + \bar{\hat{f}}^{(0)}}{\lambda} + \dfrac{n_s}{n_0}, &\text{if } i \in \mathcal{S}_0.
		\end{cases}
	\end{equation}

	\paragraph{Step 2.}
	In this part, we study the limiting form of the dual problem to find out what the weight limits should be. Note that this is a heuristic procedure to identify the weight limits, and we will rigorously prove the convergence results in Step 3.
	
	Minimizing $d(f)$ over $f \in \mathcal{F}$ is equivalent to minimizing
	\begin{equation}
		\label{eq:dual_ns}
		\begin{aligned}
			& \frac{1}{n_s} d(f) + \frac{\lambda n_s^2 (n_1 + n_0)}{2 n_1 n_0}
			\\=\;&
			\frac{1}{2\lambda n_s} \sum_{i \in \mathcal{S}_1} (f^{(1)}(X_i) - \bar{f}^{(1)})^2 +
			\frac{1}{2\lambda n_s} \sum_{i \in \mathcal{S}_0} (f^{(0)}(X_i) - \bar{f}^{(0)})^2 
			\\&-
			(\bar{f}^{(1)} + \bar{f}^{(0)}) +
			\prob_{n, \mathcal{T}}(f^{(1)} + f^{(0)})
			\\&+
			\frac{1}{4(2-\alpha)n_s} \| f^{(1)} - f^{(0)} \|_\mathcal{H}^2 +
			\frac{1}{4\alpha n_s} \| f^{(1)} + f^{(0)} \|_\mathcal{H}^2
			.\end{aligned}
	\end{equation}
	We discuss the convergence limit of \eqref{eq:dual_ns} in the following cases.
	
	\noindent \textbf{(Case 1: $0 < \alpha \le 1$)} For any $f \in \mathcal{F}$, \eqref{eq:dual_ns} converges to the following almost surely as $n \rightarrow \infty$:
	\begin{equation}
		\label{eq:dual_obj_case1}
		\begin{aligned}
			&\left\{ 
			\frac{\pi}{2\lambda} \var_{\mathcal{S}_1}(f^{(1)}(X)) -
			\E_{\mathcal{S}_1}(f^{(1)}(X)) + \E_\mathcal{T}(f^{(1)}(X))
			\right\}
			\\+
			&\left\{ 
			\frac{1-\pi}{2\lambda} \var_{\mathcal{S}_0}(f^{(0)}(X)) -
			\E_{\mathcal{S}_0}(f^{(0)}(X)) + \E_\mathcal{T}(f^{(0)}(X))
			\right\}
			.\end{aligned}
	\end{equation}
	which is a separable function of $f^{(1)}$ and $f^{(0)}$. Let $f^* = (f^{*(1)}, f^{*(0)})$ be the minimizer of \eqref{eq:dual_obj_case1}, then $f^{*(1)}$ and $f^{*(0)}$ minimize the first and second part of \eqref{eq:dual_obj_case1} respectively. To find $f^{*(1)}$, we define for $h \in \mathcal{H}$,
	\begin{equation*}
		J(h) = \frac{\pi}{2\lambda} \var_{\mathcal{S}_1}(h(X)) -
		\E_{\mathcal{S}_1}(h(X)) + \E_\mathcal{T}(h(X))
		.\end{equation*}
	By the overlap and exchangability assumptions, $\E_{\mathcal{S}_1}(\xi) = \E_\mathcal{S}(\frac{\pi(X)}{\pi} \xi)$ and $\E_\mathcal{T}(\xi) = \E_\mathcal{S}(\frac{p_t(X)}{p_s(X)} \xi)$ for an r.v. $\xi$. Hence, 
	$J(h) =
	\frac{1}{2\lambda} \E_{\mathcal{S}} (\pi(X)h(X)^2) -
	\frac{\pi}{2\lambda} \left( \E_{\mathcal{S}} \left( \frac{\pi(X)}{\pi} h(X) \right) \right)^2 -
	\E_{\mathcal{S}} \left( \left( \frac{\pi(X)}{\pi} - \frac{p_t(X)}{p_s(X)} \right) h(X)\right)
	$.
	
	Next we solve for the minimizer of $J(\cdot)$, denoted by $f^{*(1)}$, ignoring the functional restriction of $\mathcal{H}$. Using a functional differentiation argument we can show that $f^{*(1)}$ must satisfy
	\begin{equation*}
		\frac{1}{\lambda} \pi(x) f^{*(1)}(x) - 
		\frac{1}{\lambda} \E_{\mathcal{S}} \left\{\frac{\pi(X)}{\pi} f^{*(1)}(X) \right\} \pi(x) +
		\left( \frac{\pi(x)}{\pi} - \frac{p_t(x)}{p_s(x)} \right)
		= 0
		.\end{equation*}
	Solving this yields
	\begin{equation*}
		f^{*(1)}(x) = -\lambda \frac{p_t(x)}{p_s(x)\pi(x)} + C_1
		.\end{equation*}
	$C_1$ here is any constant. Plugging this into \eqref{eq:w_f} and replacing $\frac{n_s}{n_1}$ with $\frac{1}{\pi}$ yields $\frac{p_t(x)}{p_s(x)\pi(x)}$, regardless the value of $C_1$. 
	By a similar argument we can obtain
	$f^{*(0)}(x) = -\lambda \frac{p_t(x)}{p_s(x)(1-\pi(x))} + C_0$ and an associated weighting function $\frac{p_t(x)}{p_s(x) (1 - \pi(x))}$. Therefore, the weight limit can be summarized as
	\begin{equation*}
		w^*(a, x) = \frac{p_t(x)}{p_s(x)} \left\{ \frac{a}{\pi(x)} + \frac{1-a}{1-\pi(x)} \right\}
		.\end{equation*}

	\bigskip
	
	\noindent \textbf{(Case 2: $\alpha = 0$)} For any $f \in \mathcal{F}$, \eqref{eq:dual_ns} converges to the following almost surely:
	\begin{equation}
		\label{eq:dual_obj_case2}
		\begin{cases}
			\dfrac{\pi}{2\lambda} \var_{\mathcal{S}_1}(h(X)) +
			\dfrac{1-\pi}{2\lambda} \var_{\mathcal{S}_0}(h(X)) +
			\E_{\mathcal{S}_0}(h(X)) -
			\E_{\mathcal{S}_1}(h(X)),
			&\text{if } f^{(1)} = -f^{(0)} = h 
			\vspace{6pt}\\
			+\infty, &\text{otherwise}.
		\end{cases}
	\end{equation}
	The finite part can be written as
	$
	J(h) =
	\dfrac{\pi}{2\lambda} \var_{\mathcal{S}_1}(h(X)) +
	\dfrac{1-\pi}{2\lambda} \var_{\mathcal{S}_0}(h(X)) +
	\E_{\mathcal{S}_0}(h(X)) -
	\E_{\mathcal{S}_1}(h(X))
	$. By a functional differentiation argument we can see the minimizer of $J(\cdot)$ must satisfy
	\begin{equation*}
		\frac{1}{\lambda} h(x) -
		\frac{1}{\lambda} \E_{\mathcal{S}} \left\{ \frac{\pi(X)}{\pi} h(X) \right\} \pi(x) -
		\frac{1}{\lambda} \E_{\mathcal{S}} \left\{ \frac{1-\pi(X)}{1-\pi} h(X) \right\} (1-\pi(x)) +
		\left\{ \frac{1-\pi(X)}{1-\pi} - \frac{\pi(X)}{\pi} \right\}
		= 0
		.\end{equation*}
	Solving this yields
	\begin{equation*}
		f^{*(1)}(x) = -f^{*(0)}(x) = \frac{\lambda \pi(x)}{\E_{\mathcal{S}}\{\pi(X)(1-\pi(X))\}} + C
		.\end{equation*}
	Regardless of the value of $C$, the associated weighting function is 
	\begin{equation*}
		w^\dagger(a, x) = \frac{a(1-\pi(x)) + (1-a)\pi(x)}{\E_{\mathcal{S}} \{\pi(X)(1-\pi(X))\}}
		.\end{equation*}
	
	\paragraph{Step 3.}
	Our analysis of this part is adapted from the proof of Theorem 1 in \citet{hirshberg2017augmented}.
	
	\noindent\textbf{(Case 1: $0 < \alpha \le 1$)}
	Let $w^*_i = w^*(A_i, X_i)$. We consider the following function on $\mathcal{F}$:
	\begin{equation}
		\label{eq:tilde_d_def}
		\begin{aligned}
			\tilde{d}(f)
			&=
			\frac{\lambda}{2} \sum_{i \in \mathcal{S}_1} \left( \frac{-f^{(1)}(X_i) + \bar{f}^{(1)}} {\lambda} + \frac{n_s}{n_1} - w^*_i \right)^2 +
			\frac{\lambda}{2} \sum_{i \in \mathcal{S}_0} \left( \frac{-f^{(0)}(X_i) + \bar{f}^{(0)}} {\lambda} + \frac{n_s}{n_0} - w^*_i \right)^2
			\\&\qquad+
			\frac{1}{4(2-\alpha)} \| f^{(1)} - f^{(0)} \|_\mathcal{H}^2 +
			\frac{1}{4\alpha} \| f^{(1)} + f^{(0)} \|_\mathcal{H}^2
		\end{aligned}
	\end{equation}
	and let $\tilde{f} = (\tilde{f}^{(1)}, \tilde{f}^{(0)})$ be its minimizer of $\tilde{d}(f)$ and let $\{\tilde{w}_i\}$ be the weights obtained by plugging $\tilde{f}$ into \eqref{eq:w_f}. For any $f \in \mathcal{F}$, write $\delta = f - \tilde{f} = (f^{(1)} - \tilde{f}^{(1)} , f^{(0)} - \tilde{f}^{(0)})$. Next we will first show $d(\tilde{f} + \delta) - d(\tilde{f})$ is negative only if $\var_{\mathcal{S}_1}(\delta^{(1)})$ and $\var_{\mathcal{S}_0}(\delta^{(0)})$ are both small, which then implies $\hat{w}$ is close to $\tilde{w}$. Then we prove $\hat{w}$ is close to $w^*$ by showing $\tilde{w}$ is.
	
	Since $\tilde{f}$ is the minimizer of $\tilde{d}$, for any $\delta$ we have
	\begin{equation}
		\label{eq:tilde_f_eq}
		\begin{aligned}
			0 &= 
			\lim_{t \rightarrow 0} \frac{\tilde{d}(\tilde{f} + t\delta) - \tilde{d}(\tilde{f})}{t} 
			\\&=
			-\sum_{i \in \mathcal{S}_1}(\delta^{(1)}(X_i) - \bar{\delta}^{(1)})(\tilde{w}_i - w_i^*)
			-\sum_{i \in \mathcal{S}_0}(\delta^{(0)}(X_i) - \bar{\delta}^{(0)})(\tilde{w}_i - w_i^*)
			\\&\quad
			+ \frac{1}{2(2-\alpha)} \langle \tilde{f}^{(1)} - \tilde{f}^{(0)}, \delta^{(1)} - \delta^{(0)} \rangle_\mathcal{H}
			+ \frac{1}{2\alpha} \langle \tilde{f}^{(1)} + \tilde{f}^{(0)}, \delta^{(1)} + \delta^{(0)} \rangle_\mathcal{H}
			.\end{aligned}
	\end{equation}
	
	Define $R(\delta) = \frac{1}{n_s} d(\tilde{f} + \delta) - \frac{1}{n_s} d(\tilde{f})$. We rearrange $R(\delta)$ as
	\begin{align*}
		R(\delta)
		&=
		\frac{1}{2\lambda n_s} \sum_{i \in \mathcal{S}_1} (\delta^{(1)}(X_i) - \bar{\delta}^{(1)})^2 +
		\frac{1}{2\lambda n_s} \sum_{i \in \mathcal{S}_0} (\delta^{(0)}(X_i) - \bar{\delta}^{(0)})^2
		\\&\quad -
		\frac{1}{n_s} \sum_{i \in \mathcal{S}_1} \tilde{w}_i (\delta^{(1)}(X_i) - \bar{\delta}^{(1)}) -
		\frac{1}{n_s} \sum_{i \in \mathcal{S}_0} \tilde{w}_i (\delta^{(0)}(X_i) - \bar{\delta}^{(0)}) + 
		\\&\quad -
		(\bar{\delta}^{(1)} + \bar{\delta}^{(0)}) + \prob_{n, \mathcal{T}} (\delta^{(1)} + \delta^{(0)})
		\\&\quad +
		\frac{1}{4(2-\alpha) n_s} \| f^{(1)} - f^{(0)} + \delta^{(1)} - \delta^{(0)} \|_\mathcal{H}^2 + \frac{1}{4\alpha n_s} \| f^{(1)} + f^{(0)} + \delta^{(1)} + \delta^{(0)} \|_\mathcal{H}^2
		\\&\quad -
		\frac{1}{4(2-\alpha) n_s} \| f^{(1)} - f^{(0)} \|_\mathcal{H}^2 + \frac{1}{4\alpha n_s} \| f^{(1)} + f^{(0)} \|_\mathcal{H}^2
		.\end{align*}
	Subtracting $\frac{1}{n_s}$ times the right-hand side of \eqref{eq:tilde_f_eq} from $R(\delta)$ we have
	\begin{equation}
		\label{eq:R_G_H}
		\begin{aligned}
			&R(\delta) = G(\delta) - H(\delta) +
			\frac{1}{4(2-\alpha) n_s} 
			\| \delta^{(1)} - \delta^{(0)} \|_\mathcal{H}^2 +
			\frac{1}{4\alpha n_s} 
			\| \delta^{(1)} + \delta^{(0)} \|_\mathcal{H}^2
			,
			\qquad \textnormal{where}
			\\&
			G(\delta) = \frac{1}{2\lambda n_s} \sum_{i \in \mathcal{S}_1} (\delta^{(1)}(X_i) - \bar{\delta}^{(1)})^2 +
			\frac{1}{2\lambda n_s} \sum_{i \in \mathcal{S}_0} (\delta^{(0)}(X_i) - \bar{\delta}^{(0)})^2, 
			\\&
			H(\delta) = 
			\frac{1}{n_s} \sum_{i \in \mathcal{S}_1} w^*_i(\delta^{(1)}(X_i) - \bar{\delta}^{(1)}) +
			\frac{1}{n_s} \sum_{i \in \mathcal{S}_0} w^*_i(\delta^{(0)}(X_i) - \bar{\delta}^{(0)}) 
			\\&\qquad\qquad+ 
			\bar{\delta}^{(1)} + \bar{\delta}^{(0)} - 
			\prob_{n, \mathcal{T}} (\delta^{(1)} + \delta^{(0)}) 
			.\end{aligned}
	\end{equation}

	By Lemma \ref{lem:G_bound} and Lemma \ref{lem:H_bound} below, for any $\epsilon > 0$, there exists $r_n = o(n^{-\frac{1}{4}})$, $n r_n^2 \rightarrow \infty$, such that on an event with probability at least $1 - C(\epsilon + \frac{1}{n r_n^2})$ (here $C$ is a constant not related to $n$), for all $\zeta \in \mathcal{F}_1$,
	\begin{align}
		|H(\zeta)| &\le r_n^2
		\hspace{2em}
		\text{ if }
		\var_{\mathcal{S}_1}(\zeta^{(1)}) \le r_n^2
		\textnormal{ and }
		\var_{\mathcal{S}_0}(\zeta^{(0)}) \le r_n^2,
		\label{eq:bound_H}
		\\
		\frac{1}{n_a} \sum_{i \in \mathcal{S}_a} &(\zeta^{(a)}(X_i) - \bar{\zeta}^{(a)})^2
		\ge \frac{1}{2} \var_{\mathcal{S}_a}(\zeta^{(a)}),
		\hspace{2em}\text{ if }
		\var_{\mathcal{S}_a}(\zeta^{(a)}) \ge r_n^2
		\label{eq:bound_G}
		.\end{align}
	We will show $\frac{1}{n_s} \sum_{i \in \mathcal{S}} (\hat{w}_i - \tilde{w}_i)^2 \le t_n r_n^2$ on this event, where
	\begin{align*}
		t_n = \max\left\{
		\frac{4\lambda n_s}{n_1},
		\frac{4\lambda n_s}{n_0},
		4 \max\{\alpha, 2-\alpha\} n_s r_n^2
		\right\}
		.\end{align*}
	
	First consider the situation where $\delta / t_n \in \mathcal{F}_1$, i.e., $\| \delta^{(a)} \|_\mathcal{H} \le t_n$ for $a = 1, 0$.
	If $\var_{\mathcal{S}_a}(\delta^{(a)}) \le (t_n r_n)^2$ for $a = 1, 0$, then $\var_{\mathcal{S}_a}(\delta^{(a)} / t_n) \le r_n^2$, so $|H(\delta)| = t_n|H(\delta / t_n)| \le t_n r_n^2$, where the inequality follows from \eqref{eq:bound_H}. So we have
	\begin{align*}
		R(\delta)
		\ge G(\delta) - |H(\delta)|
		\ge G(\delta) - t_n r_n^2
		.\end{align*}
	So $R(\delta)$ can be only negative when $G(\delta) \le t_n r_n^2$.
	If instead $\var_{\mathcal{S}_a}(\delta^{(a)}) > (t_n r_n)^2$ for $a = $ 1 or 0, let $\beta = \max\{\sqrt{\var_{\mathcal{S}_1}(\delta^{(1)})}, \sqrt{\var_{\mathcal{S}_0}(\delta^{(0)})}\}$. Obviously $\beta > t_n r_n$. 
	Let $\delta' = r_n \delta / \beta$, then 
	$\| \delta'^{(a)} \|_\mathcal{H} = \frac{\| \delta^{(a)}\|_\mathcal{H} r_n}{\beta} \le \frac{t_n r_n}{\beta} \le 1$, and $\var_{\mathcal{S}_a}(\delta^{(a)}) = \frac{\var_{\mathcal{S}_a}(\delta^{(a)}) r_n^2}{\beta^2} \le r_n^2 $, so $|H(\delta')| \le r_n^2$ by \eqref{eq:bound_H}, which implies $|H(\delta)| = \beta |H(\delta')| / r_n \le \beta r_n \le \beta^2 / t_n$. If $\beta = \sqrt{\var_{\mathcal{S}_1}(\delta^{(1)})}$, then by \eqref{eq:bound_G}
	\begin{align*}
		R(\delta)
		\ge G(\delta) - |H(\delta)|
		\ge \frac{n_1}{4\lambda n_s} \beta^2 - \frac{1}{t_n} \beta^2
		\ge 0
		.\end{align*}
	The last inequality is because $t_n \ge \frac{4\lambda n_s}{n_1}$. If $\beta = \sqrt{\var_{\mathcal{S}_0}(\delta^{(0)})}$, similarly $R(\delta) \ge 0$ as $t_n \ge \frac{4\lambda n_s}{n_0}$.

	Next, we consider the situation where $\| \delta^{(a)} \|_\mathcal{H} > t_n$ for $a = $ 1 or 0, and show that $R(\delta)$ is non-negative. Let $k = \max\{\| \delta^{(1)} \|_\mathcal{H}, \| \delta^{(0)} \|_\mathcal{H}\} / t_n > 1$ and $\delta' = \delta / k$. Then $\| \delta'^{(a)} \|_\mathcal{H} \le t_n$ for $a=1, 0$ and equality holds for at least one of them. 
	\begin{align*}
		R(\delta) - k R(\delta')
		&= (k^2 - k) G(\delta') - (k - k) H(\delta') 
		\\&\quad+
		\frac{k^2-k}{4(2-\alpha)n_s} \| \delta'^{(1)} - \delta'^{(0)} \|_\mathcal{H}^2 +
		\frac{k^2-k}{4 \alpha n_s} \| \delta'^{(1)} + \delta'^{(0)} \|_\mathcal{H}^2
		\ge 0
		.\end{align*}
	Hence, $R(\delta) \ge k R(\delta') \ge R(\delta')$. 
	
	Next we show the non-negativity of $R(\delta')$. Note that $\| \delta'^{(a)} \|_\mathcal{H} \le t_n$, so we can reuse the results from the previous situation. 
	If $\var_{\mathcal{S}_a}(\delta'^{(a)}) > (t_n r_n)^2$ for $a=$ 1 or 0, we have shown $R(\delta') \ge G(\delta') - |H(\delta')| \ge 0$. 
	If $\var_{\mathcal{S}_a}(\delta^{(a)}) \le (t_n r_n)^2$ for $a = 1, 0$, we have shown $|H(\delta')| \le t_n r_n^2$. Since $\| \delta'^{(1)} - \delta'^{(0)} \|_\mathcal{H}^2 + \| \delta'^{(1)} + \delta'^{(0)} \|_\mathcal{H}^2 = 2(\| \delta'^{(1)} \|_\mathcal{H}^2 + \| \delta'^{(0)} \|_\mathcal{H}^2) \ge 2 t_n^2$, $\| \delta'^{(1)} - \delta'^{(0)} \|_\mathcal{H}$ or $\| \delta'^{(1)} + \delta'^{(0)} \|_\mathcal{H}$ must be no less than $t_n$. Suppose $\| \delta'^{(1)} - \delta'^{(0)} \|_\mathcal{H} \ge t_n$, then we can lower bound $R(\delta')$ by
	\begin{equation*}
		R(\delta') \ge 
		G(\delta) - |H(\delta)| + 
		\frac{\| \delta^{(1)} - \delta^{(0)} \|_\mathcal{H}^2 }{4 (2-\alpha) n_s} \ge
		- t_n r_n^2 + 
		\frac{t_n^2 }{4 (2-\alpha) n_s}
		.\end{equation*}
	The right-hand side is non-negative when $t_n \ge 4(2-\alpha)n_s r_n^2$. If $\| \delta^{(1)} - \delta^{(0)} \|_\mathcal{H} < t_n$, then $\| \delta'^{(1)} + \delta'^{(0)} \|_\mathcal{H} \ge t_n$, and a similar argument can be used to show $R(\delta') \ge 0$ when $t_n \ge 4 \alpha n_s r_n^2$.
	
	Therefore, we have shown that on an event with probability $1 - C(\epsilon + \frac{1}{n r_n^2})$, $R(\delta) \le 0$ only when $G(\delta) \le t_n r_n^2$. This is the case for $\hat{\delta} = \hat{f} - \tilde{f}$ because $\tilde{f} + \hat{\delta}$ is the minimizer $d(f)$. As a result, $G(\hat{\delta}) \le t_n r_n^2$. This can be rewritten as
	$
	\frac{\lambda}{2 n_s} \sum_{i \in \mathcal{S}} (\hat{w}_i - \tilde{w}_i)^2 \le t_n r_n^2
	$ as $\hat{w}_i - \tilde{w}_i = \{ \hat{\delta}^{(a)}(X_i) - \bar{\hat{\delta}}^{(a)} \} / \lambda$ for $i \in \mathcal{S}_a$ by \eqref{eq:w_f}.
	Recall $r_n = o(n^{-\frac{1}{4}})$, then $t_n$ is also $\mathcal{O}(\sqrt{n})$, and thus $t_n r_n^2 \rightarrow 0$. Therefore, for any small positive numbers $\eta_1, \eta_2$, we set $\epsilon = \frac{\eta_2}{2C}$ and let $n$ be large enough that $t_n r_n^2 \le \eta_1$ and $n r_n^2 \ge \frac{2C}{\eta_2}$, then
	\begin{equation*}
		\prob \left( \frac{\lambda}{n_s}\sum_{i \in \mathcal{S}} (\hat{w}_i - \tilde{w}_i)^2 > \eta_1 \right) 
		\le 
		\prob \left( \frac{\lambda}{n_s}\sum_{i \in \mathcal{S}} (\hat{w}_i - \tilde{w}_i)^2 > t_n r_n^2\right) 
		\le \eta_2
		.\end{equation*}
	This implies
	$
	\frac{1}{n_s} \sum_{i \in \mathcal{S}} (\hat{w}_i - \tilde{w}_i)^2 \overset{p}{\longrightarrow} 0
	$.
	
	Now we show $\frac{1}{n_s} \sum_{i \in \mathcal{S}} (\tilde{w}_i - w^*_i)^2 \rightarrow 0$. By the universality of $\mathcal{H}$, for $f^{*(1)}(x) = -\lambda \frac{p_t(x)}{p_s(x)\pi(x)}$ and $f^{*(0)}(x) = -\lambda \frac{p_t(x)}{p_s(x)(1-\pi(x))}$ there exist two sequences of functions $\{h_j^{(1)}\}, \{h_j^{(0)}\} \subset \mathcal{H}$ such that $\left\| h_j^{(a)} - f^{*(a)}\right\|_\infty \rightarrow 0$ for $a = 1, 0$, where $\left\| \cdot \right\|_\infty$ is the supremum norm. Denote by $v_j = (v_{ji})_{i \in \mathcal{S}}$ the weights associated with $h_j = (h_j^{(1)}, h_j^{(0)})$, then $\max_{i \in \mathcal{S}} |v_{ji} - w^*_i| \rightarrow 0$. 
	We can find a subsequence of $\{h_j\}$ (possibly with repetition), $\{h_{j_n}\}$ such that $\| h_{j_n}^{(1)} - h_{j_n}^{(0)} \|_\mathcal{H} / \sqrt{n_s} \rightarrow 0$ and $\| h_{j_n}^{(1)} + h_{j_n}^{(0)} \|_\mathcal{H} / \sqrt{n_s} \rightarrow 0$. Hence,
	\begin{equation*}
		\frac{1}{n_s} \sum_{i \in \mathcal{S}} (\tilde{w}_i - w^*_i)^2
		\le
		\frac{2}{\lambda n_s}\tilde{d} (\tilde{f}) 
		\le
		\frac{2}{\lambda n_s}\tilde{d} (h_{j_n})
		\le \max_{i \in \mathcal{S}} (v_{j_n i} - w^*_i)^2 + 
		\frac{\| h_{j_n}^{(1)} - h_{j_n}^{(0)} \|_\mathcal{H}^2}{2\lambda(2-\alpha)n_s} +
		\frac{\| h_{j_n}^{(1)} + h_{j_n}^{(0)} \|_\mathcal{H}^2}{2\lambda\alpha n_s} 
	\end{equation*}
	converges 0. Therefore, $\frac{1}{n_s} \sum_{i \in \mathcal{S}} (\hat{w}_i - w^*_i)^2 \overset{p}{\longrightarrow} 0$ as the square root of it is no greater than $\sqrt{\frac{1}{n_s} \sum_{i \in \mathcal{S}} (\hat{w}_i - \tilde{w}_i)^2} + \sqrt{\frac{1}{n_s} \sum_{i \in \mathcal{S}} (\tilde{w}_i - w^*_i)^2}$ according to the triangle inequality.

	\bigskip
	
	\noindent\textbf{(Case 2: $\alpha = 0$)}
	Since $d(f)$ equals $+\infty$ when $f^{(1)} + f^{(0)} \neq 0$, we view it as a function on $h$ by restricting $f^{(1)} = - f^{(0)} = h$:
	\begin{equation*}
		d(h) = 
		\frac{1}{2\lambda} \sum_{i \in \mathcal{S}_1} (h(X_i) - \bar{h}^{(1)})^2 +
		\frac{1}{2\lambda} \sum_{i \in \mathcal{S}_0} (h(X_i) - \bar{h}^{(0)})^2 +
		\frac{1}{2} \| h \|_\mathcal{H}^2
		.\end{equation*}
	We still define $\tilde{d}$ as \eqref{eq:tilde_d_def} but restrict $f^{(1)} = - f^{(0)} = h$ and replace $w^*$ with $w^\dagger$:
	\begin{align*}
		\tilde{d}(h) 
		&=
		\frac{\lambda}{2} \sum_{i \in \mathcal{S}_1} \left( \frac{-h(X_i) + \bar{h}^{(1)}}{\lambda} + \frac{n_s}{n_1} - w^\dagger_i \right)^2 
		\\&\quad+
		\frac{\lambda}{2} \sum_{i \in \mathcal{S}_0} \left( \frac{-h(X_i) + \bar{h}^{(0)}}{\lambda} + \frac{n_s}{n_0} - w^\dagger_i \right)^2 +
		\frac{1}{2} \| h \|_\mathcal{H}^2
		.\end{align*}
	Similarly, let $\tilde{h}$ be the minimizer of $\tilde{d}(h)$ and write $\delta = h - \tilde{h}$. For any $\delta$, by the optimality of $\tilde{h}$ we have
	\begin{equation}
		\label{eq:tilde_h_eq}
		\begin{aligned}
			0 &=
			\lim_{t \rightarrow 0} \frac{\tilde{d}(\tilde{h} + t\delta) - \tilde{d}(\tilde{h})}{t}
			\\&=
			-\sum_{i \in \mathcal{S}_1}(\delta(X_i) - \bar{\delta}^{(1)})(\tilde{w}_i - w_i^\dagger)
			+\sum_{i \in \mathcal{S}_0}(\delta(X_i) - \bar{\delta}^{(0)})(\tilde{w}_i - w_i^\dagger)
			+ \langle \tilde{h}, \delta \rangle_\mathcal{H}
			.\end{aligned}
	\end{equation}
	Define $R(\delta) = \frac{1}{n_s}d(\tilde{h}+\delta) - \frac{1}{n_s}d(\tilde{h})$. We rearrange $R(\delta)$ as
	\begin{align*}
		R(\delta) &=
		\frac{1}{2\lambda n_s} \sum_{i \in \mathcal{S}_1} (\delta(X_i) - \bar{\delta}^{(1)})^2 +
		\frac{1}{2\lambda n_s} \sum_{i \in \mathcal{S}_0} (\delta(X_i) - \bar{\delta}^{(0)})^2
		\\&\quad -
		\frac{1}{n_s} \sum_{i \in \mathcal{S}_1} \tilde{w}_i (\delta(X_i) - \bar{\delta}^{(1)}) +
		\frac{1}{n_s} \sum_{i \in \mathcal{S}_0} \tilde{w}_i (\delta(X_i) - \bar{\delta}^{(0)}) 
		\\&\quad -
		(\bar{\delta}^{(1)} - \bar{\delta}^{(0)}) +
		\frac{1}{2 n_s} \| \tilde{h} + \delta \|_\mathcal{H}^2 -
		\frac{1}{2 n_s} \| \tilde{h} \|_\mathcal{H}^2
		.\end{align*}
	Subtracting $\frac{1}{n_s}$ times the right-hand side of \eqref{eq:tilde_h_eq} from $R(\delta)$ we have
	\begin{equation}
		\label{eq:R_G_H2}
		\begin{aligned}
			&R(\delta) = G(\delta) - H(\delta) +
			\frac{1}{2 n_s} 
			\| \delta \|_\mathcal{H}^2
			,
			\qquad \textnormal{where}
			\\&
			G(\delta) = 
			\frac{1}{2\lambda n_s} \sum_{i \in \mathcal{S}_1} (\delta(X_i) - \bar{\delta}^{(1)})^2 +
			\frac{1}{2\lambda n_s} \sum_{i \in \mathcal{S}_0} (\delta(X_i) - \bar{\delta}^{(0)})^2
			\\&
			H(\delta) = 
			\frac{1}{n_s} \sum_{i \in \mathcal{S}_1} w^\dagger_i(\delta(X_i) - \bar{\delta}^{(1)}) -
			\frac{1}{n_s} \sum_{i \in \mathcal{S}_0} w^\dagger_i(\delta(X_i) - \bar{\delta}^{(0)}) +
			(\bar{\delta}^{(1)} - \bar{\delta}^{(0)})
			.\end{aligned}
	\end{equation}

	By Lemma \ref{lem:G_bound} and Lemma \ref{lem:H_bound2} below, there exists $r_n = o(n^{-\frac{1}{4}})$ such that on an event with probability at least $1 - C(\epsilon + \frac{1}{n r_n^2})$, for all $\xi \in \mathcal{H}_1$,
	\begin{align}
		|H(\xi)| &\le r_n^2
		\hspace{2em}
		\text{ if }
		\var_{\mathcal{S}_1}(\xi) \le r_n^2
		\textnormal{ and }
		\var_{\mathcal{S}_0}(\xi) \le r_n^2,
		\label{eq:bound_H2}
		\\
		\frac{1}{n_a} \sum_{i \in \mathcal{S}_a} &(\xi(X_i) - \bar{\xi}^{(a)})^2
		\ge \frac{1}{2} \var_{\mathcal{S}_a}(\xi),
		\hspace{2em}\text{ if }
		\var_{\mathcal{S}_a}(\xi) \ge r_n^2
		\label{eq:bound_G2}
		.\end{align}
	
	Similar as the previous case, we first show $\frac{1}{n_s} \sum_{i \in \mathcal{S}} (\hat{w}_i - \tilde{w}_i)^2 \le t_n r_n^2$, where
	\begin{equation*}
		t_n = \max\left\{ \frac{4\lambda n_s}{n_1}, \frac{4\lambda n_s}{n_0}, 2 n_s r_n^2 \right\}
		.\end{equation*}
	
	Consider the situation where $\| \delta \|_\mathcal{H} \le t_n$. Following the same argument in Case 1 (except that $\beta$ should be changed to $\max\{\sqrt{\var_{\mathcal{S}_1}(\delta)}, \sqrt{\var_{\mathcal{S}_0}(\delta)}\}$) we can show $R(\delta)$ is negative only when $G(\delta) \le t_n r_n^2$ given $t_n \ge \max\{\frac{4\lambda n_s}{n_1}, \frac{4\lambda n_s}{n_0}\}$. 
	
	Next, we consider the situation where $\| \delta \|_\mathcal{H} > t_n$ and show $R(\delta)$ is non-negative. Let $k = \| \delta \|_\mathcal{H} / t_n > 1$ and $\delta' = \delta / k$. Again we have $R(\delta) \ge k R(\delta') \ge R(\delta')$ because
	\begin{align*}
		R(\delta) - k R(\delta')
		= (k^2 - k) G(\delta') - (k - k) H(\delta') + \frac{(k^2 - k) \| \delta \|_\mathcal{H}^2}{2 n_s}
		\ge 0
		.\end{align*}
	Next we show the non-negativity of $R(\delta')$, which then implies $R(\delta) \ge 0$. Since $\| \delta' \|_\mathcal{H} \le t_n$, we can reuse the results from the previous situation. 
	If $\var_{\mathcal{S}_a}(\delta') > (t_n r_n)^2$ for $a=$ 1 or 0, we have shown $R(\delta') \ge G(\delta') - |H(\delta')| \ge 0$. 
	If $\var_{\mathcal{S}_a}(\delta) \le (t_n r_n)^2$ for $a = 1, 0$, we have $|H(\delta')| \le t_n r_n^2$. Then we can lower bound $R(\delta')$ by
	\begin{equation*}
		R(\delta') \ge 
		G(\delta) - |H(\delta)| + 
		\frac{\| \delta \|_\mathcal{H}^2 }{2 n_s} \ge
		- t_n r_n^2 + 
		\frac{t_n^2 }{2 n_s}
		.\end{equation*}
	The right-hand side is non-negative when $t_n \ge 2 n_s r_n^2$. 
	
	Therefore, $R(\delta)$ can only be negative when $G(\delta) \le t_n r_n^2$. Then using the same argument as in Case 1, we conclude that $\frac{1}{n_s} (\hat{w}_i - \tilde{w}_i)^2 \overset{p}{\longrightarrow} 0$. For $h^\dagger(x) = \frac{\lambda \pi(x)}{\E_{\mathcal{S}}\{\pi(X)(1-\pi(X))\}}$, there exists a sequence $\{h_j\} \subset \mathcal{H}$ such that $\left\| h_j - h \right\|_\infty \rightarrow 0$, and we can find a subsequence with $\| h_{j_n} \|_\mathcal{H}/\sqrt{n_s} \rightarrow 0$. Then $\frac{1}{n_s} \sum_{i \in \mathcal{S}} (\tilde{w}_i - w^\dagger_i)^2 \le \frac{2}{\lambda n_s} \tilde{d}(\tilde{h}) \le \frac{2}{\lambda n_s} \tilde{d}(h_{j_n}) \rightarrow 0$. By the triangle inequality we conclude $\frac{1}{n_s} \sum_{i \in \mathcal{S}} (\hat{w}_i - w^\dagger_i)^2 \overset{p}{\longrightarrow} 0 $.
	
\end{proof}

\vspace{1in}

\begin{lemma}
	\label{lem:dual_of_s_r}
	The functions $s$ and $r$ defined in \eqref{eq:primal_def} have the following Fenchel conjugates:
	\begin{align}
		s^*(v) &= \frac{1}{2\lambda} \sum_{i \in \mathcal{S}_1} (v_i - \bar{v}^{(1)})^2
		+ \frac{1}{2\lambda} \sum_{i \in \mathcal{S}_0} (v_i - \bar{v}^{(0)})^2
		+ n_s (\bar{v}^{(1)} + \bar{v}^{(0)}) - \frac{\lambda n_s^2 (n_1 + n_0)}{2 n_1 n_0}
		\label{eq:s_star}\\
		r^*(g) &= 
		\frac{1}{4(2-\alpha)} \| f^{(1)} - f^{(0)} \|_\mathcal{H}^2 +
		\frac{1}{4\alpha} \| f^{(1)} + f^{(0)} \|_\mathcal{H}^2 +
		n_s \prob_{n, \mathcal{T}}(f^{(1)} + f^{(0)})
		\label{eq:r_star}
		.\end{align}
\end{lemma}
\begin{proof}
	\underline{Calculation of $s^*$:}
	We write $s$ as $s(w) = s_0(w^{(1)}) + s_0(w^{(0)})$, where $s_0(x) = \Lambda(x) + \frac{\lambda}{2} \sum_i x_i^2$. Similar to $\Lambda$, here we define $s_0$ on a Euclidean space without specifying the dimension. The conjugate of $s_0$ is defined on the same Euclidean space:
	\begin{equation*}
		s_0^*(y) =
		\sup_{x} \left\{ \langle y, x \rangle - s_0(x) \right\} =
		\sup_{x: \mathbf{1}^\mathsf{T} x = n_s} \left\{ y^\mathsf{T}x - \tfrac{\lambda}{2} \left\| x \right\|_2^2 \right\} 
		.\end{equation*}
	By the KKT conditions of this constrained optimization problem, there exists $\gamma \in \mathbb{R}$ such that the solution $x^*$ satisfies $y - \lambda x^* + \gamma \mathbf{1} = 0$ and $\mathbf{1}^\mathsf{T}x^* = n_s$. Solving them yields $\gamma = \frac{\lambda n_s}{n_x} - \bar{y}$ and $x^* = \frac{y - \bar{y}\mathbf{1}}{\lambda} + \frac{n_s}{n_x}\mathbf{1}$, where $n_x$ is the dimension of $x$ and $y$. So $s_0^*(y) = y^\mathsf{T} x^* - \frac{\lambda}{2} \left\| x^* \right\|_2^2 = \frac{1}{2\lambda} \sum_i (y_i - \bar{y})^2 + n_s \bar{y} - \frac{n_s^2\lambda}{2n_x}$.
	Then we can calculate $s^*$ as below: for $v = (v^{(1)}, v^{(0)}) \in \mathbb{R}^{n_1} \oplus \mathbb{R}^{n_0}$,
	\begin{align*}
		s^*(v)
		&= \sup_{w^{(1)}} \left\{ \langle v^{(1)}, w^{(1)} \rangle - s_0(w^{(1)}) \right\} 
		+ \sup_{w^{(0)}} \left\{ \langle v^{(0)}, w^{(0)} \rangle - s_0(w^{(0)}) \right\} \\
		&= s_0^*(v^{(1)}) + s_0^*(v^{(0)}) \\
		&= \frac{1}{2\lambda} \sum_{i \in \mathcal{S}_1} (v_i - \bar{v}^{(1)})^2
		+ \frac{1}{2\lambda} \sum_{i \in \mathcal{S}_0} (v_i - \bar{v}^{(0)})^2
		+ n_s (\bar{v}^{(1)} + \bar{v}^{(0)}) - \frac{\lambda n_s^2 (n_1 + n_0)}{2 n_1 n_0}
		.\end{align*}
	
	\noindent\underline{Calculation of $r^*$:}
	For $f = \in \mathcal{F}$, by definition
	$
	r^*(f) = \sup_{g \in \mathcal{F}}
	\{ \langle f, g \rangle_\mathcal{F} -
	\frac{\alpha}{2} \| g^{(1)} - n_s h_\mathcal{T} \|_\mathcal{H}^2 + 
	\frac{\alpha}{2} \| g^{(0)} - n_s h_\mathcal{T} \|_\mathcal{H}^2 +
	\frac{1-\alpha}{2} \| g^{(1)} - g^{(0)} \|_\mathcal{H}^2\}
	$.
	Using functional differentiation we can see that the solution of the optimization problem satisfies
	\begin{equation*}
		\begin{cases}
			g^{*(1)} + (\alpha - 1)  g^{*(0)} = \alpha n_s h_\mathcal{T} - f^{(1)}, \\
			(\alpha - 1) g^{*(1)} +  g^{*(0)} = \alpha n_s h_\mathcal{T} - f^{(0)}.
		\end{cases}
		.\end{equation*}
	Then we can solve for $g^*$ and plug it into $r^*(f) = \langle f, g^* \rangle_\mathcal{F} - r(g^*)$, which gives
	\begin{equation*}
		r^*(g) = 
		\frac{1}{4(2-\alpha)} \| f^{(1)} - f^{(0)} \|_\mathcal{H}^2 +
		\frac{1}{4\alpha} \| f^{(1)} + f^{(0)} \|_\mathcal{H}^2 +
		n_s \langle f^{(1)} + f^{(0)}, h_\mathcal{T} \rangle_{\mathcal{H}}
		.\end{equation*}
	By the reproducing property of $\mathcal{H}$, $\langle f^{(1)} + f^{(0)}, h_\mathcal{T} \rangle_{\mathcal{H}} = \prob_{n, \mathcal{T}}(f^{(1)} + f^{(0)})$ and \eqref{eq:r_star} immediately follows.
\end{proof}

\begin{lemma}
	\label{lem:G_bound}
	There exists $r_n = o(n^{-\frac{1}{4}})$ such that for all $r \ge r_n$ the following holds with probability at least 
	$1 - 2\exp\{-c_\mathcal{H} n r^2\} - \frac{4}{n}$:
	\begin{equation*}
		\prob_n (\xi - \prob_n \xi)^2 \ge 
		\frac{1}{2} \var(\xi)
	\end{equation*}
	as long as $\xi \in \mathcal{H}_1$ with $\var(\xi) \ge r^2$. Here $c_\mathcal{H}$ is a constant solely depending on $\mathcal{H}_1$.
\end{lemma}
\begin{proof}
	Since $\mathcal{H}_1$ is $\prob$-Donsker, \citet{hirshberg2017augmented} showed in their proof of Theorem 1: there exists $r_n = o(n^{-\frac{1}{4}})$ such that with probability at least $1 - 2 \exp\{-c_\mathcal{H} n r_n^2\}$, 
	\begin{equation*}
		\prob_n h^2 \ge \frac{3}{4} \E h^2
		.\end{equation*}
	as long as $h \in \mathcal{H}_1$ and $\prob h^2 \ge r_n^2$ (their $\eta_Q$ is set as $\frac{3}{4}$ here). Here $c_\mathcal{H}$ is a constant solely depending on $\mathcal{H}_1$. The result also holds if we replace $r_n$ with any $r \ge r_n$.
	We write 
	\begin{equation*}
		\prob_n (\xi - \bar{\xi})^2
		=
		\prob_n (\xi - \E \xi)^2 - (\E \xi - \bar{\xi})^2
		.\end{equation*}
	When $\var(\xi) = \E (\xi - \E\xi)^2 \ge r_n^2$, the first term is lower bounded by $\frac{3}{4}\var(\xi)$. As for the second term, we apply the Chebyshev's inequality,
	\begin{equation*}
		\prob\left( |\E \xi - \bar{\xi}| > \frac{\sqrt{\var(\xi)}}{2} \right)
		\le \frac{\var(\bar{\xi})}{\var(\xi) / 4} = \frac{4}{n}
		.\end{equation*}
	So the desired inequality holds with probability at least $1 - 2\exp\{-c_\mathcal{H} n r_n^2\} - \frac{4}{n}$.
	
\end{proof}

\begin{lemma}
	\label{lem:H_bound}
	Define $H: \mathcal{F} \rightarrow \mathbb{R}$ as in \eqref{eq:R_G_H}. For any positive $\epsilon$, there exist $r_n = o(n^{-\frac{1}{4}})$, $n r_n^2 \rightarrow \infty$, such that for all $r \ge r_n$ the following holds with probability at least $1 - C(\epsilon + \frac{1}{n r_n^2})$, where $C$ is a constant related to $\pi(X)$:
	\begin{equation*}
		|H(\zeta)| \le r_n^2
	\end{equation*}
	as long as $\zeta \in \mathcal{F}_1$ and $\var_{\mathcal{S}_a}(\zeta^{(a)}) \le r_n^2$ for both $a = 1, 0$.
\end{lemma}
\begin{proof}
	For any $\zeta \in \mathcal{F}_1$, we can write $H(\zeta) = H_1(\zeta^{(1)}) + H_0(\zeta^{(0)})$, where 
	\begin{align*}
		H_1(h) &= \frac{1}{n_s} \sum_{i \in \mathcal{S}_1} w^*_i (h(X_i) - \bar{h}^{(1)}) + \bar{h}^{(1)} - \prob_{n, \mathcal{T}} h \\
		H_0(h) &= \frac{1}{n_s} \sum_{i \in \mathcal{S}_0} w^*_i (h(X_i) - \bar{h}^{(0)}) + \bar{h}^{(0)} - \prob_{n, \mathcal{T}} h
		.\end{align*}
	Recall that for $i \in \mathcal{S}_1$, $w^*_i = w^*(1, X_i)$ and note that $\E_\mathcal{S}[A w^*(1, X) h(X)] = \E_\mathcal{T}[h(X)]$ for all integrable $h(X)$. We can rearrange $H_1(h)$ as
	\begin{align}
		H_1(h) 
		&= \frac{1}{n_s} \sum_{i \in \mathcal{S}} A_i w^*_i (h(X_i) - \bar{h}^{(1)}) - \prob_{n, \mathcal{T}} (h - \bar{h}^{(1)})
		\nonumber
		\\&=
		\frac{1}{n_s} \sum_{i \in \mathcal{S}} A_i w^*_i (h(X_i) - \E_{\mathcal{S}_1} h) - \prob_{n, \mathcal{T}} (h - \E_{\mathcal{S}_1} h) + (\frac{1}{n_s} \sum_{i \in \mathcal{S}} A_i w^*_i - 1)(\E_{\mathcal{S}_1} h - \bar{h}^{(1)})
		\nonumber
		\\&=
		(\prob_{n, \mathcal{S}} - \prob_{\mathcal{S}}) [A w^*(1, X)(h(X) - \E_{\mathcal{S}_1} h)]
		\label{eq:H_bound_term1}
		\\&\quad-
		\left\{ 
		(\prob_{n, \mathcal{T}} - \prob_\mathcal{T}) (h(X) - \E_{\mathcal{S}_1} h)
		\label{eq:H_bound_term2}
		\right\}
		\\&\quad+
		(\frac{1}{n_s} \sum_{i \in \mathcal{S}} A_i w^*(1, X_i) - 1)(\E_{\mathcal{S}_1} h - \bar{h}^{(1)})
		\label{eq:H_bound_term3}
		.\end{align}
	By the overlap assumption $Aw^*(1, x) \le \frac{p_t(x)}{\pi(x)p_s(x)}$ is upper bounded. Then the function class consisting of only $a w^*(1, x)$ is obviously uniformly bounded and $\prob$-Donsker. Since $\mathcal{H}_1$ is also uniformly bounded as $|h(x)| = |\langle h, K(x, \cdot) \rangle|_\mathcal{H} \le \| h \|_\mathcal{H} \sqrt{K(x, x)} \le \sqrt{K(x, x)} $ for all $h \in \mathcal{H}_1$, then by \citet{kosorok2007introduction}, Corollary 9.31 (iii), $\{A w^*(1, X) h: h \in \mathcal{H}_1\}$ is also a $\prob$-Donsker class. Suppose $\frac{p_t(x)}{\pi(x)p_s(x)} \le M$. 
	\citet{hirshberg2017augmented} showed in their proof of Theorem 1: on an event with probability at least $1 - \epsilon$, where $\epsilon$ can be arbitrarily small, there exists $r_{1,1,n}' = o(n^{-\frac{1}{4}})$ such that for all $r \ge r_{1,1,n}'$ the following holds:
	\begin{equation*}
		(\prob_{n, \mathcal{S}} - \E_\mathcal{S}) (A w^*(1, X) g(X)) \le \frac{r_{1,1,n}'^2 }{6\pi M^2} 
		,\end{equation*}
	when $\E_\mathcal{S} (A w^*(1, X) g(X))^2 \le r_{1,1,n}'^2$ (their $\eta_M$ is set as $\frac{1}{6 \pi M^2}$ here). The result still holds when we replace $r_{1,1,n}'$ with any $r \ge r_{1,1,n}'$.
	
	Let $r_{1,1,n} = r_{1,1,n}'/(\sqrt{\pi}M)$. $\{r_{1,1,n}\}$ is also $o(n^{-\frac{1}{4}})$. When $\var_{\mathcal{S}_1}(h) \le r_{1,1,n}^2$, $\E_\mathcal{S} (Aw^*(1,X)[h(X) - \E_{\mathcal{S}_1}h(X)])^2 \le r_{1,1,n}'^2$, and thus we can upper bound \eqref{eq:H_bound_term1} by $\frac{r_{1,1,n}'^2}{6 \pi C^2}\le \frac{r_{1,1,n}^2}{6}$. Since $\mathcal{H}_1$ is $\prob$-Donsker, using the same argument we can find $r_{1,2,n} = o(n^{-\frac{1}{4}})$ and bound \eqref{eq:H_bound_term2} by $\frac{r_{1,2,n}^2}{6}$ with probability at least $1 - \epsilon$. 
	
	Let $r_{1,n} = \max\{r_{1,1,n}, r_{1,2,n}, n^{-\frac{1}{3}}\}$, then on an event with probability at least $1 - 2\epsilon$, \eqref{eq:H_bound_term1} and \eqref{eq:H_bound_term2} are both bounded by $\frac{r_{1,n}^2}{6}$ when $\var(h) \le r_{1,n}^2$. By  Chebyshev's inequality,
	\begin{align*}
		\prob(|\bar{h}^{(1)} - \E h| > \sqrt{\var_{\mathcal{S}_1}(h)}) \le \frac{1}{n_1} \\
		\prob(|\frac{1}{n_s} \sum_{i \in \mathcal{S}}A_i w^*(1, X_i) - 1| > \frac{r_{1,n}}{6}) \le \frac{\sigma^2}{n r_{1,n}^2}
		,\end{align*}
	where $\sigma^2 = \var_{\mathcal{S}}(A w(1, X))$. So with probability at least $1 - \frac{1}{n_1} - \frac{\sigma^2}{n r_{1,n}^2}$, the magnitude of \eqref{eq:H_bound_term3} is also bounded by $\sqrt{\var_{\mathcal{S}_1}(h)} \frac{r_{1,n}}{6} \le \frac{r_{1,n}^2}{6}$. Therefore, with probability at least $1 - \frac{1}{n_1} - \frac{1}{n_1 r_{1,n}^2} - 2\epsilon$, $H_1(h) \le \frac{r_{1,n}^2}{2}$. The inequality also holds when $r_{1,n}$ is substituted with any $r \ge r_{1,n}$.
	
	For the same reason there exists $r_{0,n} = o(n^{\frac{1}{4}})$ such that with probability at least $1 - \frac{1}{n_0} - \frac{1}{n_0 r_{0,n}^2} - 2\epsilon$, $H_0(h) \le r^2$ when $\var_{\mathcal{S}_0}(h) \le r^2$ for all $r\ge r_{0,n}$. Finally we let $r_n = \max\{r_{1,n}, r_{0,n}\}$, then the desired result follows.
	
\end{proof}

\begin{lemma}
	\label{lem:H_bound2}
	Define $H: \mathcal{H} \rightarrow \mathbb{R}$ as in \eqref{eq:R_G_H2}. For any positive $\epsilon$, there exist $r_n = o(n^{-\frac{1}{4}})$, $n r_n^2 \rightarrow \infty$, such that for all $r \ge r_n$ the following holds with probability at least $1 - C(\epsilon + \frac{1}{n r^2})$, where $C$ is a constant related to $\pi(X)$:
	\begin{equation*}
		|H(\zeta)| \le r^2
	\end{equation*}
	as long as $\zeta \in \mathcal{H}_1$ and $\var_{\mathcal{S}_a}(\zeta) \le r^2$ for both $a = 1, 0$.
\end{lemma}
\begin{proof}
	Recall $w_i = w(A_i, X_i)$ and note that $\E_\mathcal{S}[A w(1, X) f(X)] = \E_\mathcal{S}[(1-A) w(0, X) f(X)]$. We can rearrange $H(f)$ as
	\begin{equation}
		\label{eq:H_bound2_terms}
		\begin{aligned}
			H(\zeta) &=
			(\prob_{n, \mathcal{S}} - \prob_{\mathcal{S}})\{Aw(1, X)(\zeta(X) - \E_{\mathcal{S}_1} \zeta(X))\}
			\\&\quad-
			(\prob_{n, \mathcal{S}} - \prob_{\mathcal{S}})\{(1-A)w(0, X)(\zeta(X) - \E_{\mathcal{S}_0} \zeta(X))\}
			\\&\quad-
			(\frac{1}{n_s} A_i w_i - 1) (\bar{\zeta}^{(1)} - \E_{\mathcal{S}_1}\zeta) +
			(\frac{1}{n_s} (1-A_i) w_i - 1) (\bar{\zeta}^{(0)} - \E_{\mathcal{S}_0}\zeta)
			.\end{aligned}
	\end{equation}
	Following the same arguments as in the proof of Lemma \ref{lem:H_bound}, there exists $r_n = o(n^{-\frac{1}{4}})$, $n r_n^2 \rightarrow \infty$, such that with probability at least $1 - C(\epsilon + \frac{1}{n r_n^2})$, each of terms above can be bounded by $\frac{r^2}{4}$ as long as $r \ge r_n$, $\zeta \in \mathcal{H}_1$ and $\var_{\mathcal{S}_a}(\zeta) \le r^2$ for $a = 1, 0$. Then the desired result immediately follows.
\end{proof}

\paragraph{Proof of Corollary \ref{thm:vndw_limit}.} 
We rewrite $\widehat{\mathcal{V}}_n(d; w) = \frac{1}{n_s} \sum_{i \in \mathcal{S}} w_i \ind(A_i = d(X_i)) Y_i$
When $\alpha > 0$,
\begin{align*}
	|\widehat{\mathcal{V}}_n(d; w) - \widehat{\mathcal{V}}_n(d; w^*)|
	&= |\frac{1}{n_s} \sum_{i \in \mathcal{S}} (\hat{w}_i - w^*_i) \ind(A_i = d(X_i)) Y_i| \\
	&\le \sqrt{ \frac{1}{n_s} \sum_{i \in \mathcal{S}} (\hat{w}_i - w^*_i)^2 }
	\sqrt{\frac{1}{n_s} \sum_{i \in \mathcal{S}} \ind(A_i = d(X_i)) Y_i^2 } \\
	&\le \sqrt{ \frac{1}{n_s} \sum_{i \in \mathcal{S}} (\hat{w}_i - w^*_i)^2 }
	\sqrt{\frac{1}{n_s} \sum_{i \in \mathcal{S}} Y_i^2 }
	.\end{align*}
The first inequality follows from the Cauchy–Schwarz inequality, and the last one is because $\ind(A_i = d(X_i)) \le 1$. Since the potential outcomes are square integrable, $\frac{1}{n_s} \sum_{i \in \mathcal{S}} Y_i^2$ is of order $\mathcal{O}_p(1)$.  Then $|\widehat{\mathcal{V}}_n(d; w) - \widehat{\mathcal{V}}_n(d; w^*)| \overset{p}{\longrightarrow} 0$ as $\frac{1}{n_s} \sum_{i \in \mathcal{S}} (\hat{w}_i - w^*_i)^2 \overset{p}{\longrightarrow} 0$ by Theorem \ref{thm:weight_limit}. By the law of large numbers $\widehat{\mathcal{V}}_n(d; w^*) \overset{p}{\longrightarrow} \mathcal{V}_\mathcal{T}(d)$, and $\widehat{\mathcal{V}}_n(d; \hat{w})$ converges to the same limit in probability. The result for $\alpha = 0$ follows from a similar argument.

\section{Computation}
\label{sec:computaton}
In this section, we discuss the computational details of our proposed methods. In particular, we show that the optimization problem of finding the weights is a quadratic program and we also propose an algorithm for the tuning parameters ($\alpha$ and $\lambda$) selection. 
\subsection{Finite sample representation}
\label{sec:finite_sample_representation}

Since $\mathcal{H}$ is an infinite-dimensional function space, direct evaluation and optimization of \eqref{eq:opt_prog_threeway} may appear intractable. The following lemma is based on Lemma 6 of \citet{gretton2012kernel} and provides a finite sample representation for the $\textnormal{MMD}_\mathcal{H}$ between two (weighted) empirical distributions.
\begin{lemma}
	\label{lem:mmd_finite}
	Let $P = \sum_{i \in \mathcal{I}_P} p_i \delta_{x_i}$ and $Q = \sum_{j \in \mathcal{I}_Q} q_j \delta_{x_j}$ be two discrete probability distributions on $\mathcal{X}$. Here $\mathcal{I}_P$ and $\mathcal{I}_Q$ are two finite index sets, and $\{p_i, i \in \mathcal{I}_P\}$ and $\{q_j, j \in \mathcal{I}_Q\}$ are point mass that satisfy $\sum_{i \in \mathcal{I}_P} p_i = \sum_{i \in \mathcal{I}_Q} q_i = 1$. Then
	\begin{align*}
		\textnormal{MMD}_\mathcal{H}(P, Q)^2 = 
		\sum_{i \in \mathcal{I}_P} \sum_{i' \in \mathcal{I}_P} p_i p_{i'} K(x_i, x_{i'}) + 
		\sum_{j \in \mathcal{I}_Q} \sum_{j' \in \mathcal{I}_Q} q_j q_{j'} K(x_j, x_{j'}) -
		2\sum_{i \in \mathcal{I}_P} \sum_{j \in \mathcal{I}_Q} p_i q_j K(x_i, x_j)
		.\end{align*}
\end{lemma}

We can apply Lemma \ref{lem:mmd_finite} to evaluate the $\textnormal{MMD}_\mathcal{H}$ terms in \eqref{eq:opt_prog_threeway}. Specifically, denote $K_{\mathcal{S}_a, \mathcal{S}_{a'}} = \left( K(x_i, x_j) \right)_{i \in \mathcal{S}_a, j \in \mathcal{S}_{a'}}$ for $a, a' \in \{0, 1\}$, and $K_{\mathcal{S}_a, \mathcal{T}} = \left( K(x_i, x_j) \right)_{i \in \mathcal{S}_a, j \in \mathcal{T}}$. Let $w$ be a vector of length $n_s$ by stacking $(w_i)_{i \in \mathcal{S}_1}$ and $(w_i)_{i \in \mathcal{S}_0}$. By Lemma \ref{lem:mmd_finite}, we can write each of the $\textnormal{MMD}_\mathcal{H}$ terms in \eqref{eq:opt_prog_threeway} as a quadratic function of $w$. Then the optimization objective of \eqref{eq:opt_prog_threeway} can be rearranged as
\begin{equation*}
	w^\mathsf{T} \Sigma w - 2 b^\mathsf{T} w + c
\end{equation*}
with
\begin{equation*}
	\Sigma = \frac{1}{n_s^2} \begin{bmatrix}
		K_{\mathcal{S}_1, \mathcal{S}_1} + \lambda I & (\alpha - 1) K_{\mathcal{S}_1, \mathcal{S}_0} \\
		(\alpha - 1) K_{\mathcal{S}_0, \mathcal{S}_1} & K_{\mathcal{S}_0, \mathcal{S}_0} + \lambda I
	\end{bmatrix},
	~
	b = \frac{\alpha}{n_s n_t} \begin{bmatrix}
		K_{\mathcal{S}_1, \mathcal{T}} \mathbf{1} \\
		K_{\mathcal{S}_0, \mathcal{T}} \mathbf{1}
	\end{bmatrix},
	~
	c = \frac{2\alpha}{n_t^2} \sum_{i \in \mathcal{T}} \sum_{j \in \mathcal{T}} K(x_i, x_j) 
	.\end{equation*}
Here $\mathbf{1}$ is a vector of 1's. The matrix $\Sigma$ is positive definite, and the constraints in \eqref{eq:opt_prog_threeway} are linear. So the optimization problem is convex and its global optima exists. We can solve this problem using standard quadratic programming tools.

\subsection{Hyperparameter selection}
\label{sec:tuning_parameters}

There are two hyperparameters controlling the result of \eqref{eq:opt_prog_threeway}. The hyperparameter $\alpha$ controls the extent to which the weighted source sample should resemble the target sample, and $\lambda$ regularizes the smoothness of the weights. In this section we propose a heuristic procedure to select the hyperparameters by grid search. Before discussing the details and motivation of the procedure, we first outline the procedure:

\bigskip

\begin{algorithm}[H]
	\SetAlgoNoLine
	\KwIn{A grid of values for $\alpha$: $\{\alpha_j\}_{j=1}^J$ where $0=\alpha_1<\alpha_2< \cdots < \alpha_J=1$ and a grid for $\lambda$: $\{\lambda_l\}_{l=1}^L$ where $0<\lambda_1<\lambda_2< \cdots < \lambda_L$}
	\For{each $\alpha_j\in\{\alpha_j\}_{j=1}^J$}{
		\For{each $\lambda_l\in\{\lambda_l\}_{l=1}^L$}{
			Optimize \eqref{eq:opt_prog_threeway} to obtain weights $\hat{w}$ \;
			Record the resulting treatment-control imbalance using the MMD criterion on a subsample
		}
		Given $\alpha_j$, choose the $\lambda_l$ (denoted by $\lambda_{\alpha_j}$) that results in the minimum treatment-control imbalance and evaluate the resulting ITR's value function on the target sample;
	}
	Output the pair $(\alpha_j,\lambda_{\alpha_j})$ that results in an ITR with the largest value 
	\caption{Tuning parameter selection procedure}
	\label{alg:alg1}
\end{algorithm}

\bigskip

For each $\alpha$ in a grid over $[0, 1]$, we first select $\lambda$ based on treatment-control balance. As discussed in Section \ref{sec:bal_trt_ctrl}, if the distribution of confounders in the treated and control groups is not well balanced, then the ITR may be biased towards the treatment assignment in the observed data. For each $\lambda$, we calculate the weights under $(\alpha, \lambda)$, and evaluate the $\textnormal{MMD}_\mathcal{H}$ between the weighted treated group and the weighted control group. To obtain an ``honest'' evaluation, we compute this quantity on subsamples of the source sample instead of the entire source sample, and use the average result computed from many subsamples as the selection criterion. With a right penalization level, the weights are robust to sampling variation and lead to small imbalance on the subsamples. In this way, the degree of weight penalization $\lambda$ and treatment-control balance are not necessarily competing directly. In our implementation, the size of subsample is set as 80\% of the source sample, and the computation is done on 50 subsamples. Then the $\lambda$ that yields the smallest treatment-control $\textnormal{MMD}_\mathcal{H}$ will be selected for the given $\alpha$. We denote this selected value as $\lambda_\alpha$.

The selection of $\alpha$ is primarily based on an assessment of the performance of the corresponding ITR as measured by the value function. Let $\hat{w}_\alpha$ be the optimal weights under $(\alpha, \lambda_\alpha)$. We first solve for $\hat{d}_\alpha$ by maximizing $\widehat{\mathcal{V}}_n(d; \hat{w}_\alpha)$, and then evaluate the value function of $\hat{d}_\alpha$. When $\alpha$ is small, the weighted treated and control groups do not resemble the target population, so $\widehat{\mathcal{V}}_n(d; \hat{w}_\alpha)$ is not necessarily a good estimator for $\mathcal{V}_\mathcal{T}(d)$. Hence, we use a regression-based approach to evaluate the value function: the source sample is used to fit the potential outcome models and impute the potential outcomes for the target, then a plug-in estimate of $\mathcal{V}_\mathcal{T}(\hat{d}_\alpha)$ based on these imputations is computed. Then we select the $\alpha$ that yields the largest $\mathcal{V}_\mathcal{T}(\hat{d}_\alpha)$.

The default grid for $\alpha$ is $(0,0.1^4,(0.1+0.09 \times 1)^4,\ldots, (0.1+0.09 \times 9)^4,1) $ and the default grid for $\lambda$ is $(0.001, 0.002,0.005,0.012,0.028,0.066,0.152,0.351,0.811,1.874,4.329, 10)$.  Even though any kernels could be used in our proposal algorithm, we recommend to use universal kernels such as Gaussian kernels or Laplacian kernels rather than the nonuniversal kernels such as linear kernels due to universal kernels's ability of approximating complex functions well \citep{simon2018kernel}. Moreover, the theoretical properties proved in the Section~\ref{sec:method_weight_limit} only hold when universal kernels are used.

This hyperparameter tuning procedure involves solving the quadratic programming \eqref{eq:opt_prog_threeway} under various choices of $(\alpha, \lambda)$. To speed up computation, when solving \eqref{eq:opt_prog_threeway} under a new pair of hyperparameters, we use the solution under the previous hyperparameters as a warm start. This is possible when using the \texttt{OSQP} package \citep{osqp}. The empirical results in the Simulation Studies section suggest that this tuning procedure works quite well in practice and is able to select suitable hyperparameters.

\section{Additional simulation results}
\subsection{Computation time of the proposed method}
\begin{figure}[!ht]
	\centering
	\includegraphics[width=.59\linewidth]{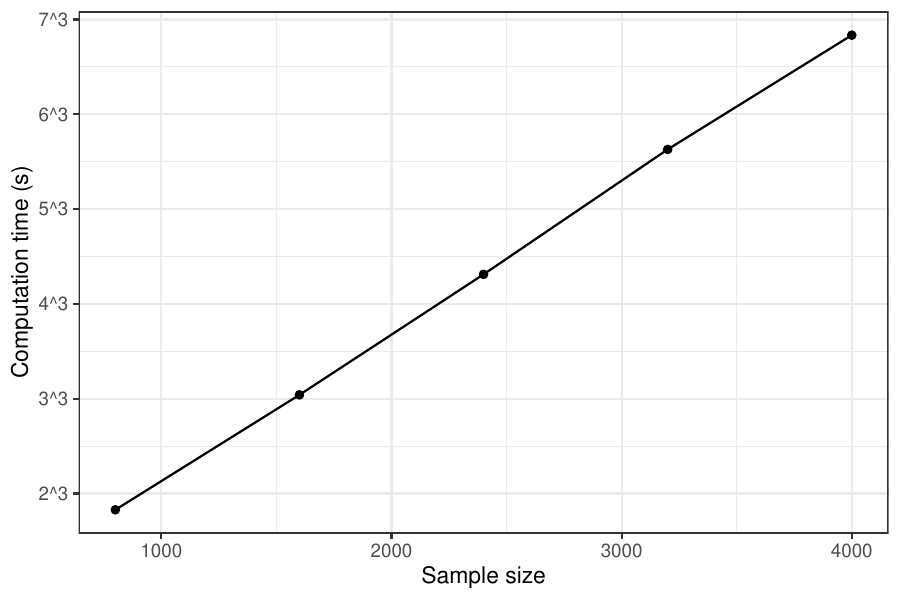}
	\caption{Computation versus $n$.}
	\label{fig:computation_time_vs_n}
\end{figure}

Here we report the computation time of the proposed method under various sample sizes, using the first setting in Section \ref{sec:simu} for illustration. Figure \ref{fig:computation_time_vs_n} shows that our method has cubic time complexity with respect to sample sizes. The experiment was run on a laptop under a macOS system with 2.3 GHz Quad-Core Intel Core i5 CPU and 16GB 2133 MHz LPDDR3 memory. 

\subsection{Non-linear ITR class}
\begin{figure}[!ht]
	\centering
	\includegraphics[width=.99\linewidth]{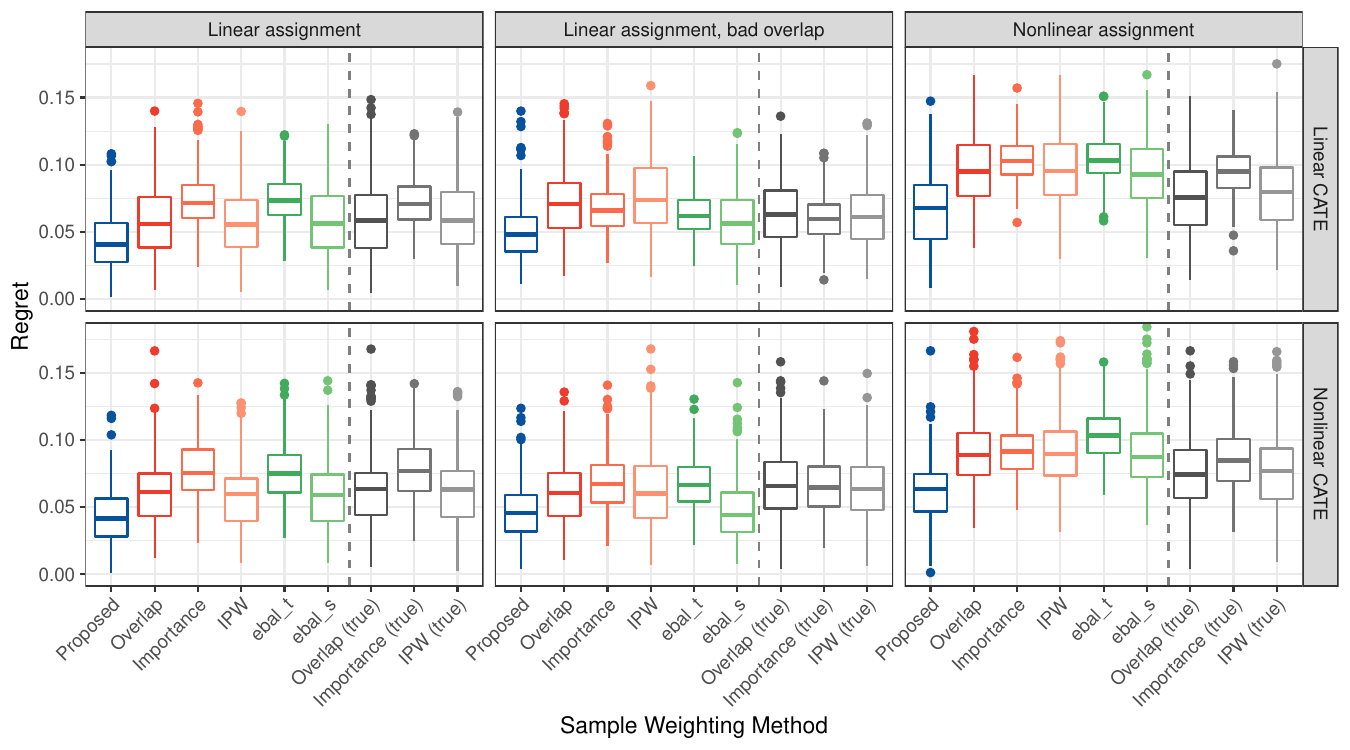}
	\caption{Regret compared to the optimal tree-based ITR using outcome weighted learning.}
	\label{fig:simu_value_tree}
\end{figure}

Although the proposed method is mainly designed for situations where a limited ITR class is employed, we explore its compatibility with more flexible rule classes. In this section, we repeat the simulation studies of Section \ref{sec:simu}, but after obtaining the sample weights with different methods, we apply them to decision tree-based ITR learning. Note that the grid search procedure for identifying the ground truth of optimal linear ITR for the target population mentioned in Section  \ref{sec:simu} is not applicable for identify the optimal tree ITR. Instead, we constructed a classification tree to approximate the sign of the conditional average treatment effect (CATE) values using samples from the target population and used it as the ground truth to calculate the regret. The results are presented in Figure \ref{fig:simu_value_tree}, where we note that our proposed weights result in ITRs with smaller regret than competing weights in all settings. For the setting with the linear assignment, bad overlap and a nonlinear CATE, our weights result in similar but slightly better performance than \texttt{ebal\_s}.

\subsection{Linear ITR by regression}

Throughout the main text and the previous simulation studies we applied sample weights to obtain ITR by solving a weighted classification problem as detailed in Section \ref{sec:simu}, which is derived from maximizing the value function. In this section, we couple the weights with another type of ITR learning approach that regresses the outcome on the covariates and the interaction between the treatment and covariates \cite{qian2011performance, chen2017general}. Specifically, we first solve
\begin{equation*}
	\min_{\gamma_0, \gamma, \beta_0, \beta} \frac{1}{n_s} \sum_{i \in \mathcal{S}} w_i \left( Y_i - \gamma_0 - \gamma^\mathsf{T} X_i - (A_i - 0.5) (\beta_0 + \beta^\mathsf{T} X_i) \right)^2
\end{equation*} to obtain $( \hat{\gamma_0}, \hat{\gamma}, \hat{\beta_0}, \hat{\beta} )$; then the ITR is defined as
$\hat{d}(x) = \textnormal{sign} \left( \hat{\beta_0} + \hat{\beta}^\mathsf{T} x \right)$. The weights $w_i$ are used for ITR generalization. We also compare with ITR from no weights or equivalently $w_i\equiv 1$.

\begin{figure}[!ht]
	\centering
	\includegraphics[width=.99\linewidth]{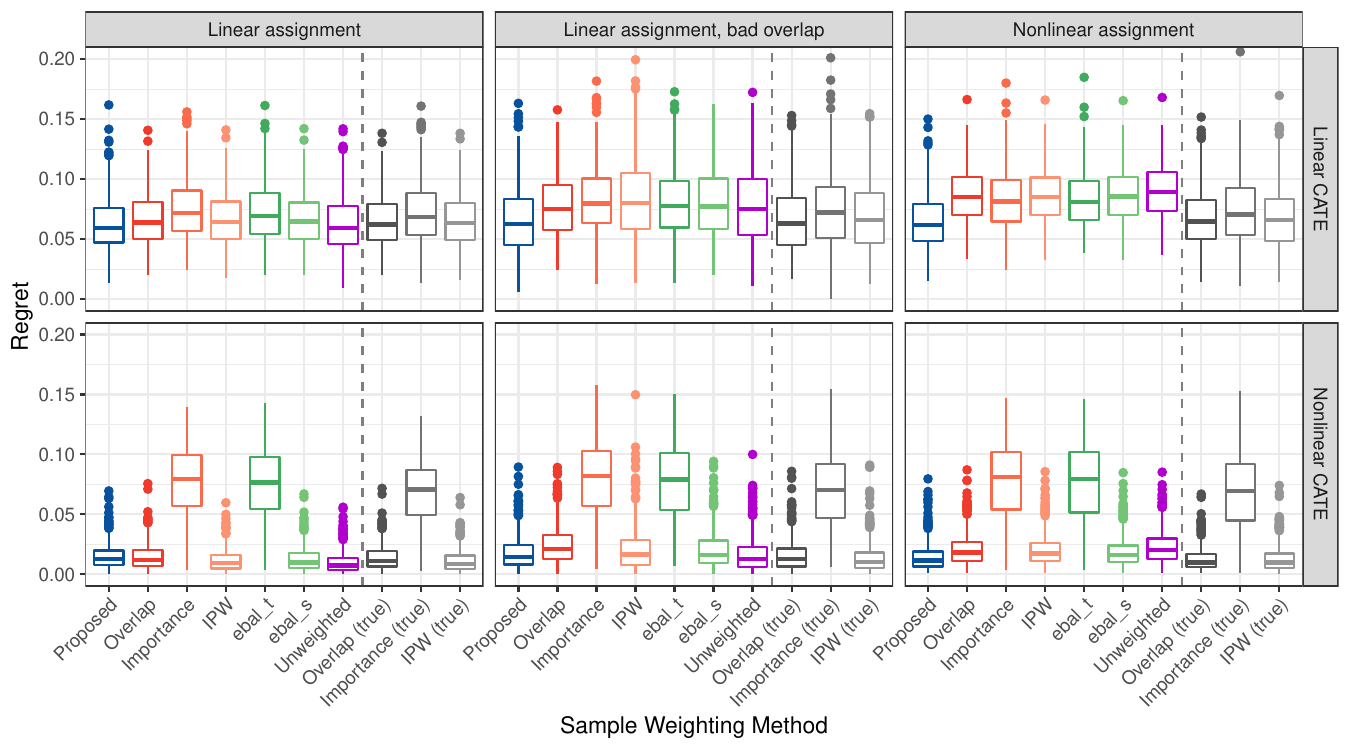}
	\caption{Regret compared to the optimal linear ITR using a regression based approach.}
	\label{fig:simu_results_lm}
\end{figure}

The regression-based method in essence works by predicting $Y$ using both $A$ and $X$. The rationale of weighted regression can be intuitively understood as the following: an estimated model that minimizes the training error (the source population) may not minimize the testing error (the target population) under covariate shift, especially when the model is misspecified. The importance weights aim to calibrate the training error to the testing error using the covariate information of the source and target populations. However, as discussed in the main text, a drawback of importance weights is that it could reduce the effective sample size. Hence, it is unnecessary to address the covariate shift issue if the model is correctly specified. With the regression-based ITR learning method, we repeat the evaluation process in Section \ref{sec:simu}, comparing the same set of sample weighting methods. As shown in Figure \ref{fig:simu_results_lm}, our proposed weighting method consistently yields the best performance; the other methods that have comparable performance are those relying on true probability functions, which are of course not available in practice.

\section{Additional simulation results for a new setting}

We additionally conduct simulations with a substantially more complex data generating mechanism. This data generating mechanism has more complex nuisance parameter functions (the participation probability, the propensity score, the main effect function, and the CATE) with more number of covariates. Specifically, we generate 10-dimensional covariates $X_i$ independently from a uniform distribution on $[-2, 2]^{10}$; the first 6 of these covariates are active in at least one element of the data generation mechanism and the last 4 are noise covariates. As in the simulation setup in the main text, given the covariates $X_i$, the population indicator $S_i$ is then generated from a Bernoulli distribution with probability $\rho(X_i)$ of being in the source population, where 
$$
\rho(x) = G((x_1 + x_2)^2 + x_3 + x_4 + (x_3 + x_4 )^2 + 2\ind(|x_5| > 1) - 0.5x_5x_6),
$$
where $G(z)$ is the same as in the simulation study in the main text. The treatment assignment probability is set as
$$
\pi(x) = G(-.8(x_1 - x_2)^2 + .4 (x_3 - x_4)^2 + \ind(x_5 > x_6) - .5).
$$

The observed outcome is generated as $Y_i = m(X_i) + (A_i - 0.5) \tau(X_i) + \varepsilon_i$, where 
$$
m(x) = -x_1 - x_2 + x_3 + x_4 + x_1 x_2 + x_1 ^ 2 + x_2^2 + 
(1 + x_3 x_4) ^ 2 + 2 x_5 x_6 + 2 \cos(x_1 + x_2) + .5
$$ 
and $\varepsilon_i \distas{i.i.d.} N(0, 0.5^2)$. Similar to the setup of the main text, the CATE function is set as 
$\tau(x) = \kappa \tau_\text{NL}(x) + (1 - \kappa) \tau_\text{L}(x)$, where
\begin{align*}
	\tau_\text{NL}(x) &= (2x_1 - x_2)^2 + (2x_1 + x_2)^2 - 2x_3 + x_4 + 2(x_5 - x_6)^2\text{sign}(x_3) - 1,\\
	\tau_\text{L}(x) &= 0.4 x_2 + 0.6 x_1 - 0.5.
\end{align*}
For these simulation studies, we use $\kappa = 0.4$.

\begin{figure}[ht!]
	\centering
	\includegraphics[width=.49\linewidth]{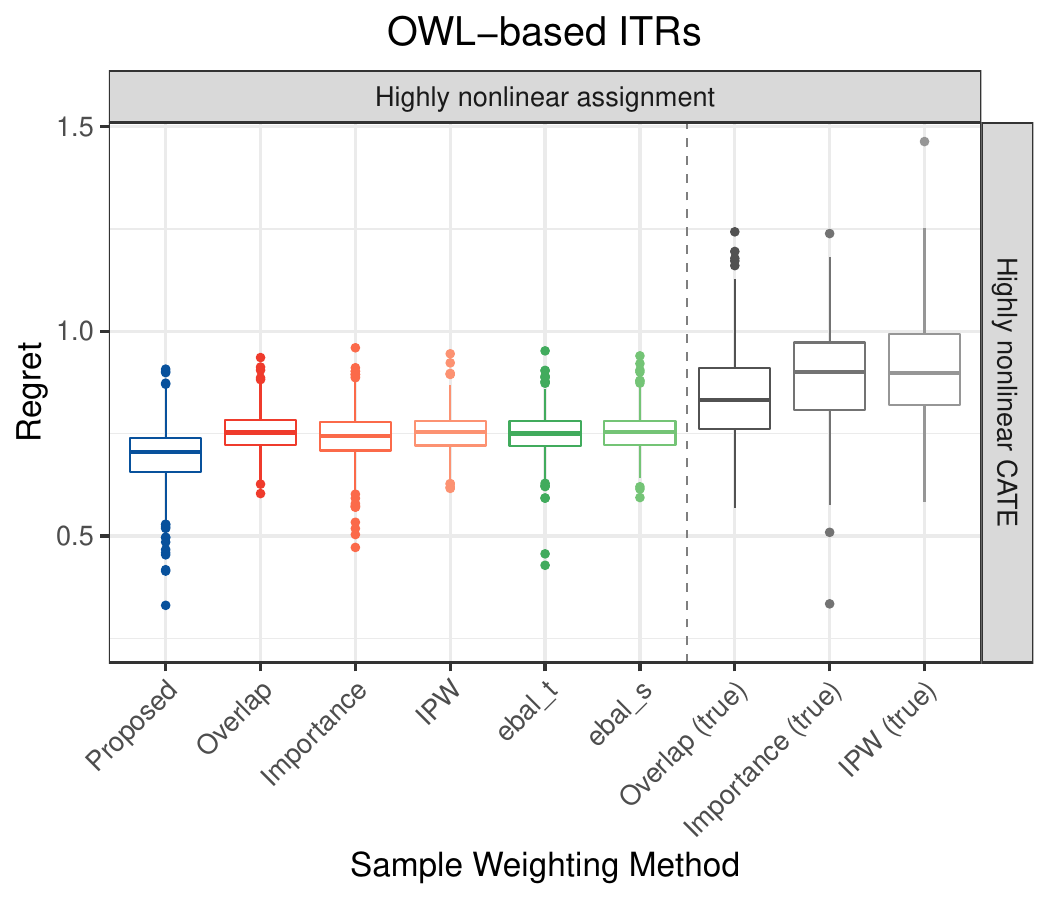}
	\includegraphics[width=.49\linewidth]{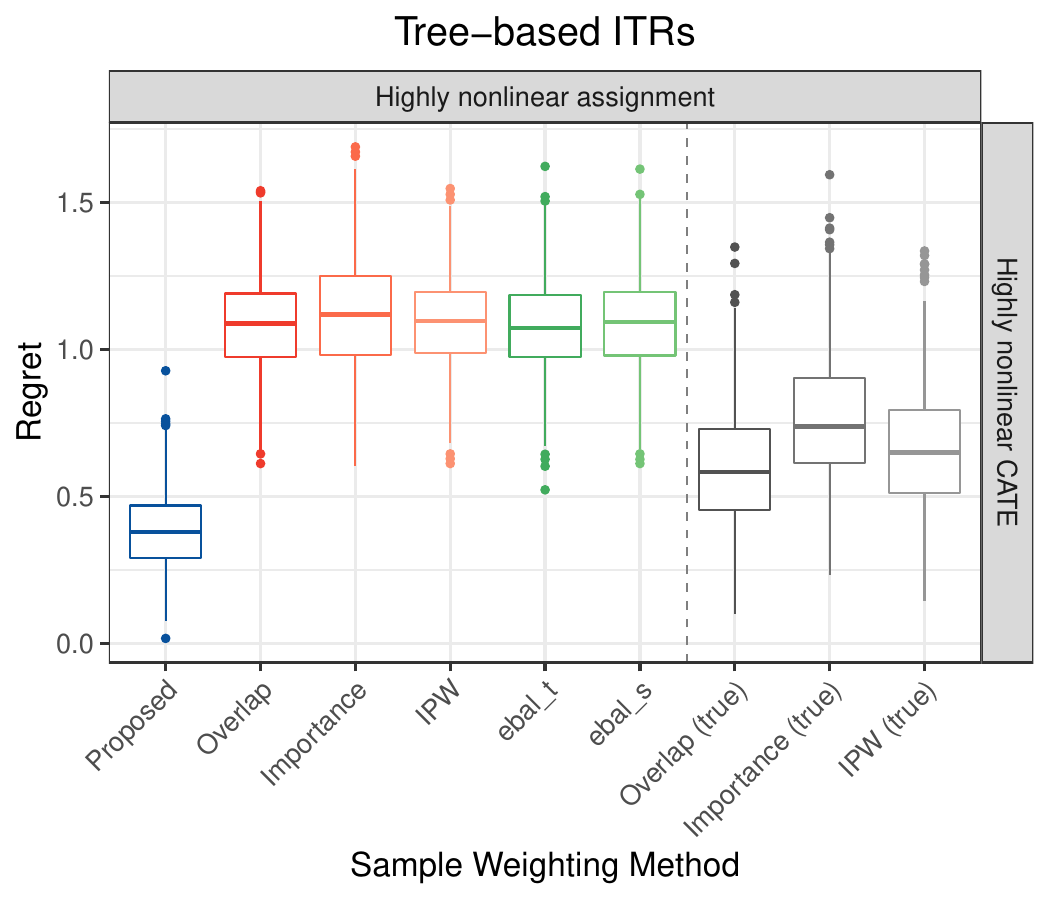}
	\includegraphics[width=.49\linewidth]{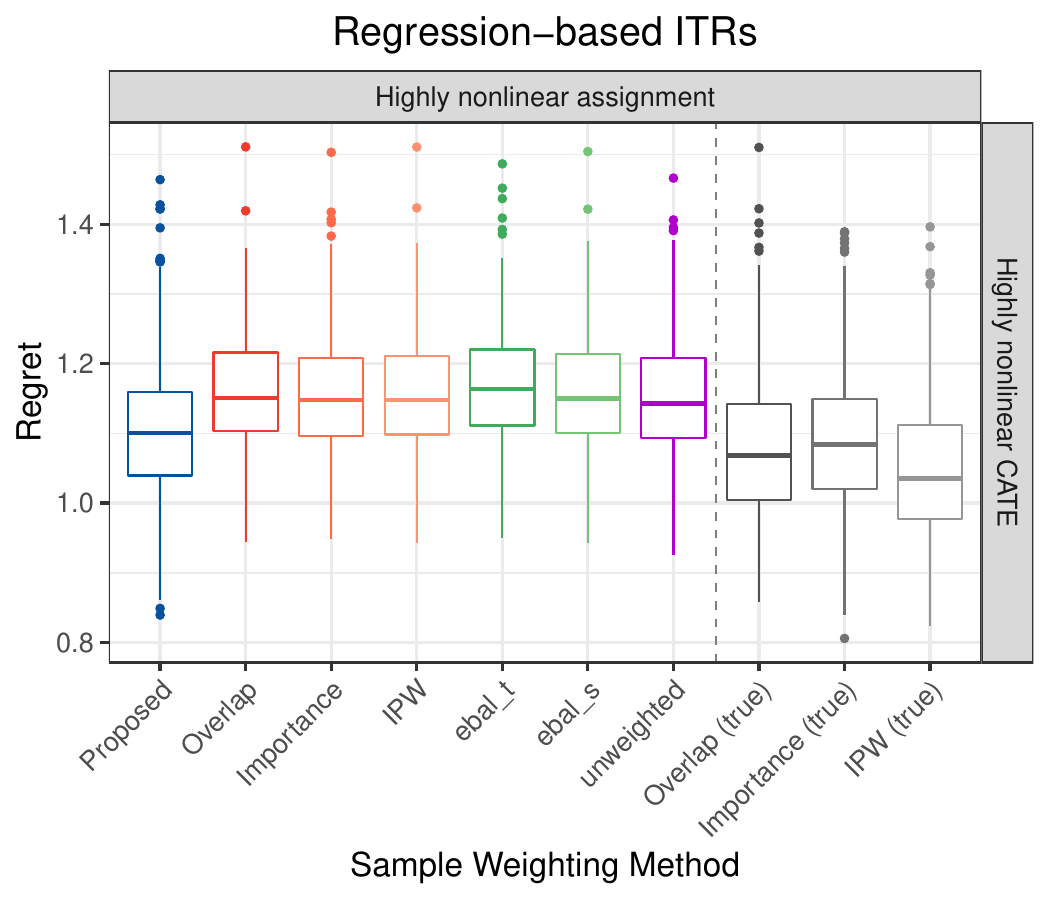}
	\caption{Regret for more complex simulation setting}
	\label{fig:simuregretlmnewdata212-24-22}
\end{figure}

We apply all weights used in the main simulation study to three different estimation approaches for the ITR: 1) outcome weighted learning (OWL)-based estimators as used in the main text, 2) weighted tree-based ITRs as described and used in Section S3.2, and 3) weighted regression-based ITR estimators as described and investigated in Section S3.3.
The results for each of these three ITR estimation settings are displayed in Figure \ref{fig:simuregretlmnewdata212-24-22}. For all three settings , our approach consistently works better than all other weights. Occasionally, some other weighting methods would have comparable performance to ours, but those are the one relying on true probability functions, which are not available in practice.

\clearpage

\section{Additional real data results regarding covariate balancing diagnosis}
\begin{figure}[!ht]
	\centering
	\includegraphics[height= .55\linewidth, width=.9\linewidth]{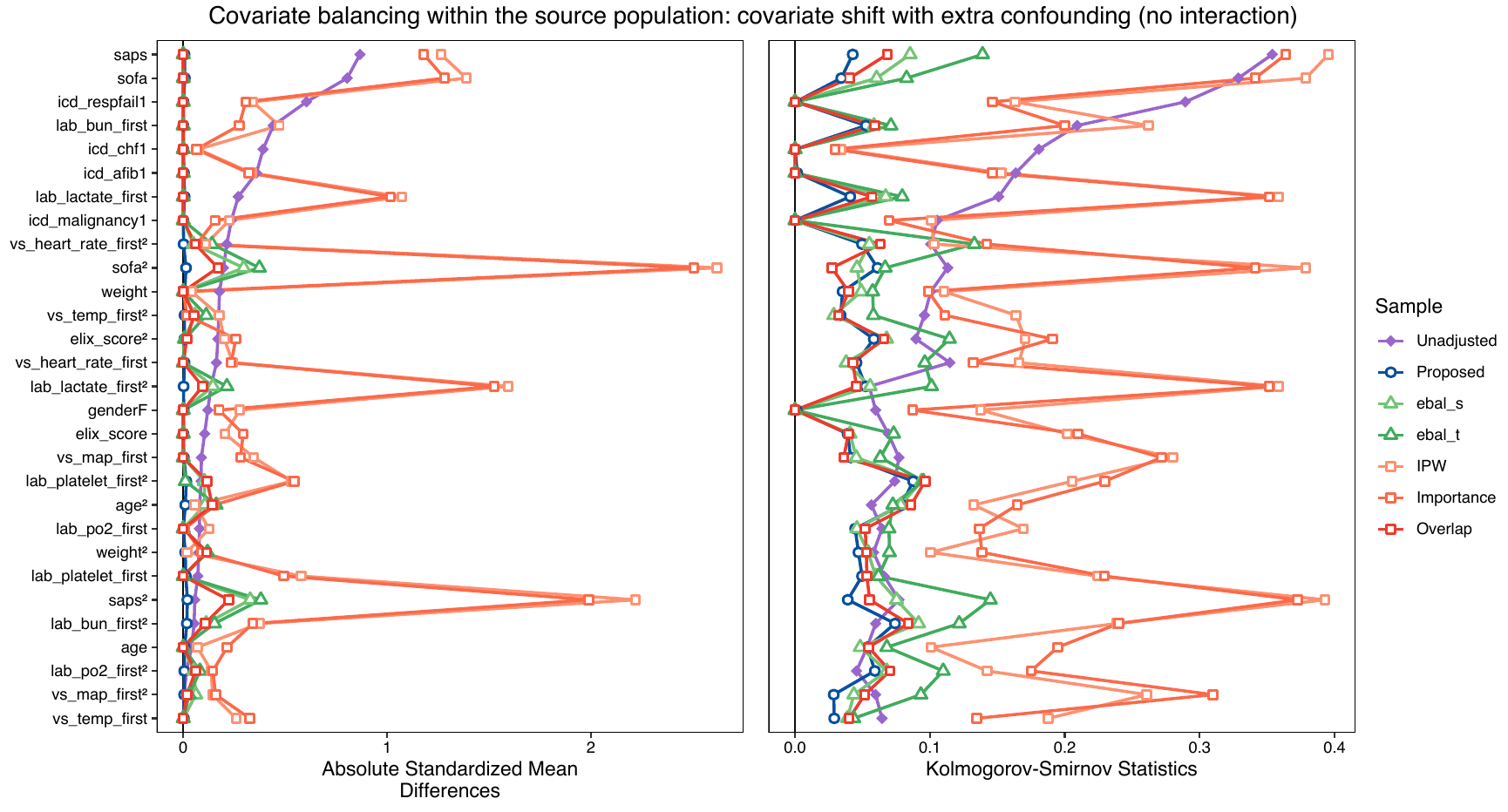}
	\caption{``Love" Plots for Assessing Covariate Balancing between the Treatment and Control Groups in the Source Sample.}
	\label{figure:love_plot1}
\end{figure}

\begin{figure}[!ht]
	\centering
	\includegraphics[height= .55\linewidth, width=.9\linewidth]{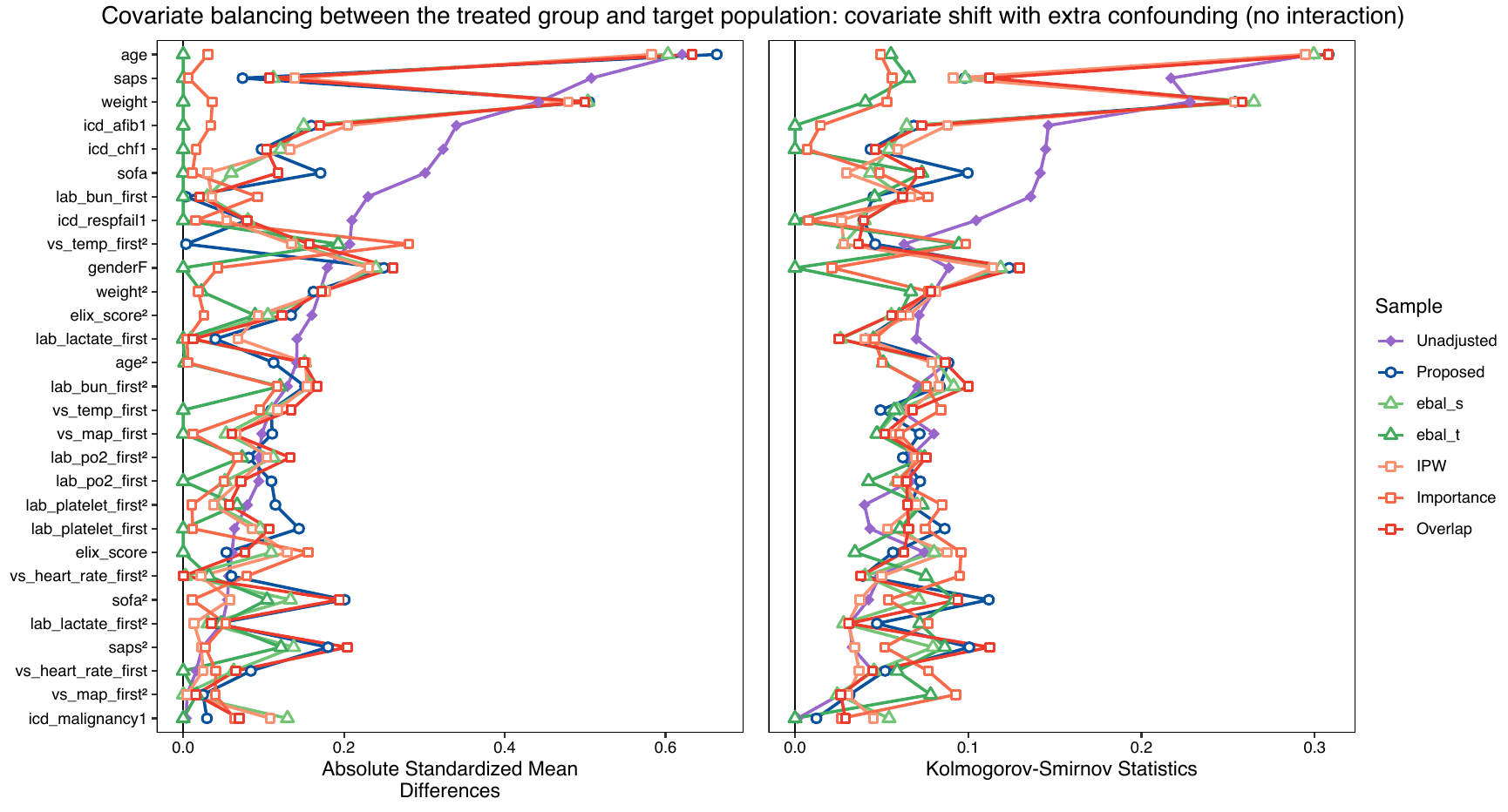}
	\caption{``Love" Plots for Assessing Covariate Balancing between the Treatment Group in the Source Sample and the Target Sample.}
	\label{figure:love_plot2}
\end{figure}

\begin{figure}[!ht]
	\centering
	\includegraphics[height= .55\linewidth, width=.9\linewidth]{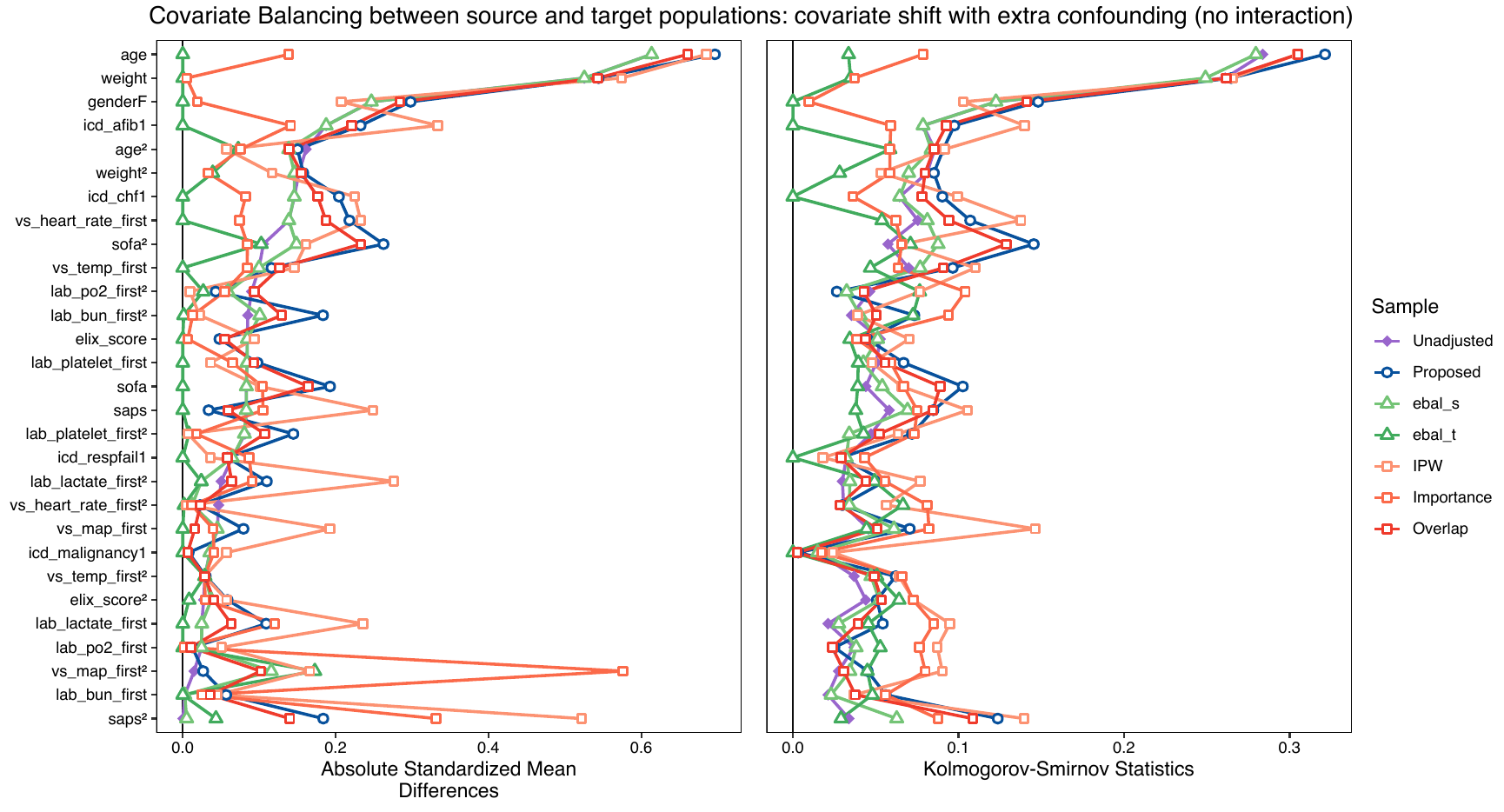}
	\caption{``Love" Plots for Assessing Covariate Balancing between the Control Group in the Source Sample and the Target Sample.}
	\label{figure:love_plot3}
\end{figure}

Covariate balance between the treatment and control groups in the source sample, and between the source and the target samples, can be assessed by the following descriptive statistics after weighting: standardized mean differences (SMDs) \citep{rosenbaum1985constructing}, variance ratios \citep{austin2009balance}, and weighted Kolmogorov-Smirnov (KS) statistic \citep{austin2015moving}.  In particular, a useful overview of how to assess covariate balance and compare balancing methods using \texttt{R} package \texttt{cobalt} is available in the following vignette: \begin{verbatim} https://ngreifer.github.io/cobalt/articles/cobalt.html \end{verbatim}

Here, we provide ``love'' plots \citep{ahmed2006heart} in Figures \ref{figure:love_plot1}, \ref{figure:love_plot2}, and \ref{figure:love_plot3} to visualize these descriptive statistics under one simulation run of the data analysis presented in Section \ref{sec:data} under the setting of having both covariate shift and extra confounding (with no covariate interactions).

For both SMDs and KS, closer to $0$ indicates better balance.
As seen in these plots, our proposed method balances the treated and control group in both first and second moments of the covariates (according to the SMDs presented on the left), while other approaches only balance the first moments; additionally, according to the weighted KS statistics, our proposed methods additionally balance the marginal \textit{distributions} of the covariates well. However, we note that our weights do not balance the source sample well to the target population because our method in this case selected $\alpha$ close to $0$, which emphasizes balance between the treatment groups; such an $\alpha$ was chosen by our tuning parameter selection procedure because the ITR class was evidently well-specified, making source-target balance less critical for this dataset.

\end{document}